\documentclass{article}

\usepackage{jfrExamplee}
\usepackage[letterpaper,margin=1in]{geometry}
\usepackage{amsmath,amsthm,amsfonts,amscd} 
\usepackage{amssymb}
\usepackage{citesort}
\usepackage{url}
\usepackage{bm}
\usepackage{graphicx}
\usepackage{cancel}
\usepackage{subfigure} 
\usepackage{setspace}
\usepackage{caption}
\usepackage{natbib}

\newcommand{\ve}[1]{\ensuremath{{\bf #1}}}
\newcommand{\vet}[1]{\ensuremath{{\bf #1}(t)}}

\newcommand{\vedash}[1]{\ensuremath{{\bf #1}'}}
\newcommand{\norm}[1]{\ensuremath{{\left|\left|{#1}\right|\right|}}}
\newcommand{\tny}[1] {\ensuremath{{\scriptscriptstyle #1}}}

\renewcommand{\norm}[1]{\left| \left| #1 \right| \right|}
\newcommand{\thder}[1]{\dddot{#1}}
\newcommand{\sth}{\sin{\theta}}
\newcommand{\cth}{\cos{\theta}}
\newcommand{\Eq}[1]{Equation~(\ref{#1})}
\newcommand{\Sec}[1]{Section~\ref{#1}}
\newcommand{\Fig}[1]{Figure~\ref{#1}}

\author{
Shilpa Gulati\thanks{{\bf Corresponding author}. This work was performed as part of Shilpa's Ph.D.~\citep{Gulati_2011} in the Mechanical Engineering Department at the University of Texas, Austin, TX 78712, USA. Shilpa now works at Bosch Research and Technology Center, 4009 Miranda Ave Suite 150, Palo Alto, CA 94304, USA. {\bf Email:} shilpa.gulati@gmail.com}
\And
Chetan Jhurani\thanks{Tech-X Corporation,	5621 Arapahoe Ave, Boulder, CO 80303, USA. {\bf Email:} chetan.jhurani@gmail.com} \\
\And
Benjamin Kuipers\thanks{Electrical Engineering and Computer Science Department, University of Michigan, Ann Arbor, MI 48109 USA. {\bf Email:} kuipers@umich.edu} \\
}

\title{A Nonlinear Constrained Optimization Framework for Comfortable and Customizable Motion Planning of
Nonholonomic Mobile Robots -- Part I}

\begin{document}

\singlespace
\maketitle

\begin{abstract}
{
In this series of papers, we present a motion planning framework for planning
comfortable and customizable motion of nonholonomic mobile robots
such as intelligent wheelchairs and autonomous cars.  In this first one we present the mathematical foundation of our framework.

The motion of a mobile robot that transports
a human should be comfortable and customizable. We identify several properties that a trajectory must have
for comfort. We model 
motion discomfort as a weighted
cost functional and define comfortable motion planning as a nonlinear constrained optimization problem of 
computing trajectories that minimize this discomfort given the appropriate boundary conditions and constraints.
The optimization problem is infinite-dimensional
and we discretize it using conforming finite elements.
We also outline a method by which different users may customize the motion to achieve personal comfort.

There exists significant past work in kinodynamic motion 
planning, to the best of our knowledge, our work
is the first comprehensive 
formulation of kinodynamic motion planning
for a nonholonomic mobile robot as a nonlinear optimization
problem that includes all of the following -- a careful
analysis of boundary conditions, continuity requirements on trajectory,
dynamic constraints, obstacle avoidance constraints, and a robust numerical
implementation.

In this paper, we present the mathematical foundation of the
motion planning framework and formulate the full nonlinear
constrained optimization problem. We describe, in brief, the
discretization method using finite elements and the
process of computing initial guesses for the optimization
problem. Details of the above two are presented in Part II~\citep{Gulati_2013b} of the
series.

}
\end{abstract}

\setcounter{secnumdepth}{2}

\section{Introduction}
\label{sec:introduction}
Autonomous mobile robots such as intelligent wheelchairs and autonomous cars have the potential to 
 improve the quality of life of many demographic groups. Recent surveys have concluded 
that many users with mobility impairments find it difficult or impossible to operate existing power
wheelchairs because they lack the necessary motor skills or cognitive 
abilities~\citep{Fehr_2000,Simpson_2008}. Assistive mobile robots such as smart wheelchairs and 
scooters that can navigate autonomously benefit such users by increasing their mobility~\citep{Fehr_2000}. Autonomous cars have the potential to increase
the mobility of a significant proportion of the elderly whose driving ability is reduced due to
age-related problems~\citep{kpmg_2012}.

The motion of an autonomous mobile robot should be comfortable to be acceptable to human users. 
Moreover, since the feeling of comfort is subjective, different users
should be able to customize the motion according to their comfort. 
Motion planning is a challenging problem and has received significant
attention. See ~\citep{Latombe_1991,Hwang_1992,Choset_2005,LaValle_2006,LaValle_2011a,LaValle_2011b}.
However, most 
of the existing motion planning methods have been developed
for robots that do not transport a human user and issues
such as comfort and customization have not been explicitly addressed.

In this paper we focus on planning comfortable motion for
nonholonomic mobile robots such that the motion can
be customized by different users.
Our key contributions are as follows:
\begin{itemize}
\item We model user discomfort as a weighted cost functional. This is informed by studies
of human comfort in road and railway vehicle literature that
indicate
that human discomfort increases with the magnitude of acceleration
and jerk and that comfortable levels of these quantities have
different magnitudes in the direction of motion and perpendicular to
the direction of motion~\citep{Suzuki_1998}. Thus, our cost functional
is a weighted sum of
the following three physical quantities: total travel time, tangential jerk,
and normal jerk.  

Minimum jerk cost functionals have previously been used in literature~\citep{Zefran_1996,Arechavaleta_2008} for
optimal motion planning.
What is new here is the separation of tangential and normal components, 
and computing the weights using the technique of dimensional analysis~\citep{Langhaar_1951}
that allows us to develop a straightforward procedure for varying the weights for 
customization.

\item We develop a 
framework for planning
comfortable and customizable motion.
Here, we present a precise mathematical formulation of kinodynamic
motion planning of a nonholonomic mobile robot moving on a plane as a nonlinear constrained optimization
problem. This includes an in-depth analysis of conditions under which the cost-functional is
mathematically meaningful, analysis of boundary conditions, 
and precise formulation of constraints necessary for motion comfort
and for obstacle avoidance. 
To the best of our knowledge, such a formulation is absent from the literature. 

The idea of computing optimal trajectories that
minimize a cost functional is not new and has been used
for planning optimal trajectories for 
wheeled robots~\citep{Dubins_1957,Reeds_1990,Balkcom_2002,Bianco_2005}
and manipulators~\citep{Fernandes_1991,Shiller_1994,Zefran_1996,Arechavaleta_2008}.
All of these formulations make several limiting assumptions, such as known travel 
time, or known path, or boundary conditions on configuration but not its derivatives. 
None of these approaches consider obstacles. The closest existing work to ours in 
terms of problem formulation
and numerical solution method is~\citep{Zefran_1996}, but obstacle avoidance constraints are
not part of this formulation.

The trajectories planned by our framework
have several useful properties -- they exactly satisfy boundary conditions on position, orientation,
curvature, speed and tangential acceleration, satisfy kinematic and dynamic constraints, and avoid obstacles
while minimizing discomfort. Further, our framework is capable of planning a family of trajectories
between a given pair of boundary conditions and can be customized by different users to obtain a trajectory that
satisfies their comfort requirements.

\item We represent obstacles 
as star-shaped domains with piecewise $C^2$ boundary. 
This choice allows treatment of non-convex obstacles without subdividing
them into a union of convex shapes. This reduces the number of 
constraints imposed due to obstacles and leads to a faster 
optimization process. Such a representation of obstacles
is not very common in robotics where most
collision-detection algorithms
assume polygonal obstacles, and detect collisions
between non-convex polygons by subdividing them into convex 
polygons~\citep{Quinlan_1994,Mirtich_1998,Lin_2004}.

\item We use the Finite Element Method to discretize the above 
infinite-dimensional problem into a finite dimensional problem.
The finite element method is not unknown in trajectory planning
but it is not very common. 
However, it is a natural choice for problems like ours and we
strongly believe that using it provides us insight, flexibility,
and reliability that is not easily obtained by choosing other discretization methods.

\item Our method can be used independently for motion planning of nonholomic
mobile robots. It can also
be a used as local planner in sampling-based methods~\citep{LaValle_2006} 
since the trajectories
computed by our method exactly satisfy boundary conditions, kinodynamic constraints,
continuity requirements, and avoid obstacles.

\end{itemize}

\section{Background and related work}
\label{sec:background_and_related_work}
In this section, we characterize motion comfort by analyzing studies
in ground vehicles, elevator design, and robotics. 
We then review existing motion planning methods and identify their
strengths and limitations in planning comfortable motion.

\subsection{Comfort}
\label{sec:comfort}
\emph {Comfort - What is it? Comfort has both psychological and 
physiological components, but it involves a sense of subjective 
well-being and the absence of discomfort, stress or pain}
~\citep{Richards_1980}.

Studies to characterize comfort
in ground vehicles such as automobiles and trains have shown that the feeling of
comfort in a vehicle is affected by various characteristics of the 
vehicle environment including dynamic factors (such as acceleration 
and jerk), ambient factors (such as temperature and air quality), 
and spatial factors (such as seat quality and leg room)~\citep{Richards_1980}.
In this work we focus on comfort due to dynamic factors alone. 

Passenger discomfort increases as the 
magnitude of acceleration increases~\citep{Suzuki_1998,Jacobson_1980,Pepler_1980,Forstberg_2000,Chakroborty_2004}.
This is because an increase in magnitude of acceleration implies
increase in magnitude of force experienced by a passenger.
Two separate components
of acceleration effect discomfort -- tangential component
along the direction of motion and normal component perpendicular
to the direction of motion~\citep{Jacobson_1980,Pepler_1980,Forstberg_2000}. 
The normal component is zero in a straight line motion but
becomes important when traversing curves.
The actual values of comfortable
bounds of the two components may be different~\citep{Suzuki_1998}, may vary
across people, may depend on the mode of transportation, and
may depend on the passenger's position~\citep{Pepler_1980,Forstberg_2000}. Hence, guidelines
for ground transportation design prescribe maximum values of accelerations~\citep{Suzuki_1998,Chakroborty_2004,Iwnicki_2006}, or maximum values 
of comfort indices 
that are functions of accelerations~\citep{ISO_1997,CEN_1999}.

Discomfort also increases as the magnitude of jerk increases~\citep{Pepler_1980,Forstberg_2000}. 
This is because a high rate of change of jerk implies a high rate of change of magnitude or direction or both of
the forces acting on the passenger.
Upper bounds on
jerk for comfort have been proposed for road~\citep{Chakroborty_2004}
and railway vehicles~\citep{Suzuki_1998}. 
In elevator design, motion profiles are
designed for user comfort by choosing profiles with smooth
accelerations and low jerk~\citep{Hall_1970,Krapek_1993,Spielbauer_1995}.  

From a geometric standpoint, it has been known for more than a
century that sharp changes in
curvature of roads and railway tracks can be dangerous and can cause passenger discomfort~\citep{Laundhart_1887,Glover_1900,Lamm_1999}. 
For a point mass moving on a path, the normal acceleration at a point is given
by $\kappa v^2$ where $\kappa$ is the curvature of the path and $v$ is
the speed at that point. If curvature is not continuous, then normal
acceleration cannot be continuous unless the speed goes to zero at
the point of discontinuity. This is clearly undesirable
for comfort. In robotics, the desire to 
drive a robot with non-zero speed from start to goal
has led to the development of methods for planning continuous curvature paths
~\citep{Lamiraux_2001,Fraichard_2004,Bianco_2004,Bianco_2005,Piazzi_2007}.

To summarize, in a motion planning context, a trajectory should have the following
properties for comfort. First, the acceleration should be continuous and bounded.
Second, jerk should be bounded. Third, the geometric path should have 
curvature continuity so that is is possible to travel from start to end without
stopping.  Fourth, a trajectory should exactly satisfy appropriate end point boundary conditions
boundary conditions on position, orientation, curvature, speed, and acceleration
since many tasks require precise these (for example, positioning
at a desk for an intelligent wheelchair, parking in a tight parking space
for a car). Fifth, it should be possible to join multiple trajectories such that 
the combined trajectory has the above properties. This means that a trajectory should 
satisfy the above described boundary conditions on both ends.

\subsection{Motion planning} 
\label{sec:motion_planning}
There exists a large body of work on motion planning. Before
reviewing this work, we define some terms. The space of all possible 
positions and orientations
of a robot is called {\em configuration space}.
The space of all possible configurations
and their first derivatives is called {\em state space}. A {\em trajectory} is
a time-parameterized function of configuration. A {\em control trajectory}
is a time-parameterized function of control inputs.

Motion planning is the problem 
of finding either a trajectory, or a control trajectory, or both, given the 
initial and final configuration, and possibly their  first and higher derivatives, 
such that the geometric path does not intersect any obstacles, and the trajectory 
satisfies {\em kinematic} and {\em dynamic constraints}. Kinematic
constraints refer to constraints on configuration and dynamic constraints refer
to constraints on velocity and its higher 
derivatives. These constraints arise from physics, engineering limitations,
or comfort requirements.

A variety of methods have been used to solve various aspects of the
motion planning problem. {\em Path Planning} methods 
focus on the purely geometric problem of finding a collision-free path.
Another set of
methods, stemming from differential geometric control theory,
focus on computing control inputs that 
steer a robot to a specified position and orientation or that 
make a robot follow a specified path. {\em Kinodynamic motion planning} methods, 
consider both dynamics and obstacles and focus on computing collision-free
trajectories that satisfy kinematic and dynamic constraints. 
See ~\citep{Hwang_1992,Latombe_1991,Choset_2005,LaValle_2006} for excellent 
presentation of all three kinds of methods,~\citep{Laumond_1998} for 
differential geometric control methods, 
and~\citep{Donald_1993,Fraichard_1993,LaValle_2001_2,Hsu_2002} for 
kinodynamic planning. In this work, we use motion planning in
the sense of~\citep{Donald_1993}, that is, we speak of 
kinodynamic motion planning, consistent with the informal definition
presented above.

\subsubsection{Sampling-based methods}
\label{sec:sampling_based_methods}
Sampling-based methods have found widespread acceptance and practical
use for motion planning. These methods are used
for both path planning~\citep{LaValle_1998,Kavraki_1996,LaValle_2001}
and for motion 
planning~\citep{Canny_1988,Barraquand_1989,Donald_1993,Fraichard_1993,LaValle_2001,LaValle_2001_2,Hsu_2002}.
See~\citep{LaValle_2006} for an in-depth discussion.
The main idea in all sampling-based methods is to sample the
state space~\citep{Donald_1993,LaValle_2001_2} or state-time
space~\citep{Erdmann_1983,Barraquand_1990,Fraichard_1993,Hsu_2002} to construct a directed
graph called a {\em roadmap} from the start state to the goal region. The vertices of this
graph are points in the obstacle free region of the appropriate 
space (state space or state-time space) and the edges 
are trajectory segments that satisfy kinodynamic constraints.
The sequence of control inputs associated with 
the edges of the roadmap is the control trajectory.
Among the most computationally efficient methods here
are the ones that add vertices to the graph by randomized sampling.

Randomized sampling-based algorithms follow two paradigms -- multiple-query and
single-query. In the multiple-query paradigm, a roadmap is constructed
once and used to answer multiple path planning queries. These algorithms
are particularly computationally efficient in an unchanging environment since
a single roadmap can be used to answer multiple queries. Some of the
most well-known algorithms that follow this paradigm are Probabilistic
Roadmaps (PRMs) and its variants~\citep{Kavraki_1996}.

In the single-query paradigm, a roadmap is constructed for each query.
Some of the most well-known algorithms that follow this
paradigm are Randomly Exploring Dense Trees (RDT)~\citep{LaValle_2006} and
its variants~\citep{Hsu_2002,Karaman_2011}. These methods start with a
roadmap rooted at the start state and iteratively add vertices
by randomized sampling of the appropriate space. Different variants
differ in the way they add a new vertex to the roadmap.
We describe RDT in some detail here. A new vertex is added as follows
(i) a sample point $q_{new}$ is chosen from a randomized sequence (ii) a vertex
$q_{curr}$ in the graph that is closest to the sample point, according to a distance
metric, is selected (iii) all controls from a set of discretized controls are applied
to $q_{curr}$ and the system is allowed to evolve for a fixed time $\Delta t$ (iv) out of all the
new points that can be reached via collision-free trajectories satisfying differential
constraints, the point nearest $q_{curr}$ is chosen and added to the graph.
This process is continued till a vertex in the goal region is added
to the graph.

The closeness of the end point of the trajectory to the goal state
increases as the resolution increases, but in general, it is not possible to 
find a trajectory that exactly reaches the goal state. 
If it
is desired to reach a goal state exactly, then a boundary value
problem has to be solved between the end state of
the solution trajectory and the goal state. This is a non-trivial problem since
the solution must avoid obstacles and satisfy kinodynamic constraints.
Some sampling-based methods are bidirectional, that is, they simultaneously 
grow roadmaps from the start state as well as the goal state. In this
case, a solution trajectory exactly satisfies the boundary conditions. However, like before,
a boundary value problem has to be solved to connect the two roadmaps.

Since a fixed value of control input is applied for a finite
length of time at each step, the planned path lacks curvature continuity and has
to be smoothed in a post-processing step. Curvature continuity can be attained at
the cost of increasing the dimensionality of the state space, and 
has been demonstrated only for a path planning problem~\citep{Scheuer_1998}.
Similarly, for achieving acceleration continuity the dimensionality of
the state space has to be increased resulting in increased computational complexity.

Recently, sampling-based algorithms described above have been shown
to almost always converge to solution that has non-optimal cost~\citep{Karaman_2011}
and a new algorithm, RRT* was proposed for planning asymptotically optimal paths.
Results in a two dimensional configuration space showed that algorithm is computationally
efficient. While promising, these results are very recent, and extending
this work to kinodynamic motion planning is yet to be carried out.

Another set of sampling-based methods can compute optimal trajectories by constructing
a grid over the state space or state-time space and searching
this discrete grid using graph-search algorithms such as A*~\citep{Canny_1988,Barraquand_1989,Fraichard_1993}. This grid is 
called the state-lattice. Each pair of neighboring vertices of the grid are 
connected to each other by a trajectory that satisfies kinodynamic constraints. 
Three key choices effect the solution quality. First,
the choice
of discretization determines the closeness of the solution to the true optimum
and the speed of computing the solution. Second, the choice of a neighborhood (e.g. k-nearest) for
a vertex determines the connectivity of the space.
Third, the choice of a method for computing 
trajectory segments between vertices determines the quality of the solution trajectory. Computing trajectory segments between adjacent states involve solving
a non-trivial boundary value problem.
For continuity of curvature, velocity and acceleration
between connected trajectory segments, the
state space should include curvature, and the first and second derivative of configuration. 
This results in increase in dimensionality of the search space and hence increase in computational time.
For this reason, lattice-based methods have been shown to plan trajectories, with
some but not all of the properties necessary for comfort (\Sec{sec:comfort})
in autonomous driving applications. Continuous curvature trajectories are demonstrated in~\citep{Pivtoraiko_2009},
continuous velocity but not continuous curvature trajectories are demonstrated in~\citep{Likhachev_2009}. Trajectories
with continuous curvature, speed, and acceleration are demonstrated in~\citep{McNaughton_2011}
Here
the problem is tractable because the sampling can be restricted to the road on which 
the vehicle drives.
Efficiently planning trajectories that satisfy all properties of comfort as described
in \Sec{sec:comfort} in less structured environments very much remains an open problem.

\subsubsection{Optimal-control based methods}
\label{sec:optimal_motion_planning}
The problem of planning trajectories that are optimal with respect to
some performance measure and also avoid obstacles 
has been shown to very hard~\citep{Canny_1987}, even in relatively simple cases.
However, for many applications, we do require that a solution trajectory
be optimal with respect to some performance measure such as 
time, path length, energy etc.

Optimal control methods~\citep{Bryson_1975,Troutman_1995} have traditionally been used for computing
optimal
trajectories for systems subject to dynamic 
constraints in the absence
of obstacles
and have been widely applied in
aerospace engineering and control-systems engineering. 
The formulation consists of constructing a
cost functional representing the cumulative cost over the duration of motion
and minimizing the cost functional to find a desired state trajectory or 
control trajectory or both. A {\em functional} is an operator 
that maps a function to a real or complex number. 

Sufficient conditions for a solution of the minimization problem are given by
the Hamilton-Jacobi-Bellman (HJB) equation. HJB is a second-order partial
differential equation with end-point boundary conditions. Analytic
solutions of the HJB equation for linear systems with
quadratic cost have long been known~\citep{Bryson_1975}. 
For general nonlinear systems, the HJB equation has to
be solved numerically.

Necessary conditions for optimality are derived using Pontryagin's
principle and consist of a set of first-order ordinary
differential equations. These differential equations convert the
optimization problem into a two-point boundary value problem.
The system of differential equations can either be solved 
analytically (where possible) or numerically using methods such
as the shooting method or finite-difference methods. 

Analytical solution to the problem of finding minimum length
paths for Dubins~\citep{Dubins_1957} car and 
Reeds and Shepp~\citep{Reeds_1990} car (see~\citep{Soueres_1998}) was 
found using such an approach. Dubins car
is only allowed to move forward while Reeds and Shepp car is also allowed
to move backward. These paths are
comprised of straight line and arc segments and minimize the distance
traveled by the mid-point of the rear axle. Each path segment is
traversed at a fixed speed, so the trajectories corresponding to these paths are
also time-optimal for a given speed. More recently, shortest paths for a 
differential drive wheeled robot were developed by including a rotation
cost in the cost functional~\citep{Balkcom_2002} (since a differential drive robot
can turn in place). Such minimum-time paths lack curvature continuity and 
require frequent stopping and reorienting of wheels.

More complex problems generally require a numerical
solution. One frequently used numerical method is the shooting
method where the two point boundary value problem is converted 
into an initial value problem. Shooting methods have been used for trajectory 
planning for nonholonomic mobile robots~\citep{Howard_2007,Ferguson_2008}. 
However, in shooting methods, it is challenging to specify a 
good initial guess of the unknown parameters that produces a final 
state reasonably close to the specified state. In general, the trajectories
computed do not exactly satisfy end point boundary conditions.

Instead of solving the differential equations
representing necessary conditions, approximation methods that discretize
the infinite-dimensional problem into a finite-dimensional one
and optimize the cost functional directly in this finite-dimensional space
can be used. Such methods have been used for planning
optimal trajectories of robots. In~\citep{Fernandes_1991}, control
inputs that minimize total control energy to travel between a given pair
of boundary states are computed. Here Fourier basis functions are used
for discretization.
In~\citep{Zefran_1996}, trajectories 
that minimize the integral of square of $L^2$ norm of end-effector jerk and 
the square of $L^2$ norm of time derivatives of joint torque
vector, subject to torque constraints, are computed. Here a 
finite-element discretization is used. Other discretizations are
also possible, such as B-spline~\citep{Bobrow_2001}
and spectral~\citep{Strizzi_2002} discretization.  

Very few of the existing optimal control approaches include obstacle-avoidance.
Not only do obstacle avoidance constraints make the optimal control problem
highly nonlinear, but also each obstacle divides the set of feasible solutions
into disjoint regions.
One of the earliest methods that included dynamic constraints and obstacle-avoidance
for motion planning of autonomous vehicles used a two step approach --
first an obstacle free path was found and then an optimal speed 
on this path was computed~\citep{Shiller_1991_1,Shiller_1991_2}. Because of 
path-velocity decomposition, the resulting trajectory is, in general, not
optimal. Obstacles were included as hard constraints for a
two-dimensional translating robot in~\citep{Tominaga_1990}.

\subsubsection{Learning methods}
\label{sec:learning_methods}
Optimal control methods require an accurate model of the kinematics and
dynamics of the robot as well as models of the robot's interactions with
the world. Such models are not always available. Further, it is not straightforward
to develop an appropriate cost functional for a given task. Even if such models
and cost functionals are available, searching through the high dimensional
configuration space of the robot (e.g. in the case of humanoid robots) for an 
optimal trajectory can be computationally expensive. One set of learning-based methods
use the key observation that, in practice, robot trajectories are restricted 
to a manifold by the task and by the kinodynamic constraints. The dimension of
this manifold is, in general, lower than the dimension of the configuration space.
These methods aim to learn the structure of this manifold from observed data
of the robot's movement~\citep{Ramamoorthy_2008}. Another set of methods aim to 
learn motion primitives for a specific task using observed data from 
human movements~\citep{Schaal_2003}. 
A detailed discussion of these methods is beyond the
scope of this work and the interested reader is referred to the following 
works for more details:~\citep{Full_1999,Schaal_2003,Calinon_2009,Ramamoorthy_2008,Havoutis_2012}. 

\subsubsection{Summary}
\label{sec:motion_planning_summary}
Trajectories computed by sampling-based methods, in general, lack continuity
of curvature and acceleration. While these problems can be solved by increasing
the dimensionality of state space at the cost of increased computational complexity, 
the problems of lack of optimality
and not satisfying the goal boundary conditions exactly still remain.

Optimal control methods have primarily been demonstrated for trajectory
planning in the absence of obstacles. Further, a comprehensive
formulation of kinodynamic motion planning problem for nonholonomic mobile
robots that includes obstacle avoidance is absent. 
Thus, none of the existing methods can be directly applied to planning comfortable
and trajectories.  To this end, we develop a motion planning framework to compute
trajectories that result in comfortable motion.

\section{Overview of the approach}
\label{sec:overview_of_approach}
At the root of our framework is the assumption that user discomfort
can be quantified as a cost functional, and that trajectories
that minimize this discomfort and avoid obstacles will result in user-acceptable
motion. We outline the main steps of our approach below.

\begin{itemize} 
\item \emph {Formulate user discomfort as a mathematically meaningful cost functional}.
Based on existing literature, and making the assumption that 
a user would like to travel as fast as is
consistent with comfort, we define a measure of discomfort as a
weighted sum of the following three terms: total travel time, time
integrals of squared tangential jerk and squared normal jerk.

Each weight used in the discomfort measure to add different quantities
is the product of two factors. The first factor has physical units so that
the physical quantities with different dimensions can be added together.
It is a fixed function of known length and velocity scales. The second factor is a
dimensionless parameter that can be varied according to user
preferences.  The dimensional part is derived using the standard
technique of dimensional analysis~\citep{Langhaar_1951}.

\item \emph{Define the problem}.
We formulate our motion planning problem as follows: ``Given the
appropriate boundary conditions, kinodynamic constraints, the weights in the cost 
functional, and a representation of obstacles, find a
trajectory that minimizes the cost functional, satisfies boundary conditions,  
respects constraints, and avoids obstacles''. This description is 
transformed into a precise mathematical
problem statement using a general nonlinear constrained optimization
approach.

\item \emph{Choose a parameterization of the trajectory}.
Mathematically, one can use different functions to fully describe
a trajectory.  We express the trajectory by an orientation and
a velocity as functions of a scaled arc-length parameter where 
the scaling factor is an additional scalar unknown to be solved for. 
This leads to a relatively simple expression
for discomfort.
We use a scaled arc-length parameterization  Thus, we do not assume that the
path length is known until the problem is solved.

\item \emph{Analyze the boundary conditions}.
A complete analysis of boundary conditions shows that for the
optimization problem to be well-posed, we need to impose
boundary conditions on position, orientation, curvature, speed, and
tangential acceleration on each end. Further, we find that
three different types of boundary conditions on speed
and tangential acceleration on each end describe all types 
of motion tasks of interest such as starting/ending at rest or not. 

\item \emph{Choose a representation of obstacles}.
To incorporate obstacle avoidance, we make the assumption that each
obstacle can be modeled as a star-shaped domain with a boundary
that is a piecewise smooth curve with continuous second order derivative.
If an obstacle is not
star-shaped, our framework can still handle it if
it can be expressed as a finite union of piecewise smooth star-shaped domains.
It is assumed that a representation of each obstacle is known in
polar coordinates where the origin lies in the interior of the kernel
of the star-shaped domain.  Since each obstacle is assumed star-shaped,
the constraint that the trajectory stay outside obstacles can be
easily cast as an inequality.   

To efficiently incorporate obstacle avoidance constraints, we
have to introduce position on the path as an additional unknown.
This leads to a sparse Hessian of constraint inequalities, which
otherwise would be dense.  The position as an unknown is redundant
in that it can be computed from the two primary unknowns
(orientation and speed).  Hence that relation is included as an extra
equality constraint.

\item \emph{Discretize the problem}.
We use finite elements to convert the infinite-dimensional
minimization problem to a finite dimensional one.  For discomfort
to be mathematically meaningful and bounded, both speed
and orientation must have square-integrable second derivatives.
We use a uniform mesh and cubic Hermite polynomial shape functions
on each element for speed and orientation.  Starting or
stopping with zero speed is a special case
that requires that speed have an infinite derivative
(with respect to scaled arc-length) with a known strength on the
corresponding boundary point.  In this case we use singular
shape functions for speed only on elements adjacent to the
corresponding boundary.

In the non-discretized version of the optimization problem
the obstacle avoidance constraint can be expressed as the
condition that each point on the trajectory should be outside
each obstacle.  We discretize this into a finite dimensional
set of inequalities by requiring that some fixed number of
points on the trajectory be outside each obstacle.

\item \emph{Compute an appropriate initial guess}.
A good initial guess is necessary for efficiently solving any 
nonlinear optimization problem. In general, there exist infinitely many
trajectories between any given pair of boundary conditions. 
Based on our analysis of this non-uniqueness,
we compute a set of four good quality initial guesses
by solving another, simpler, optimization problem.
These initial guesses do not incorporate obstacle-avoidance
constraints. Four discomfort minimization problems, corresponding
to these four initial guesses, are solved to find four
trajectories. The lowest cost
trajectory can be chosen as the final solution.

\item \emph {Implement and solve}.
We use Ipopt, a robust large-scale nonlinear constrained
optimization library~\citep{Wachter_2006} to solve the
discretized problem. 

\end{itemize}

\section{Organization of this paper}
\label{sec:organization_of_paper}
This paper is organized as follows. \Sec{sec:preliminary_material} presents some preliminary 
material on the motion of nonholonomic mobile robot on a plane and on parametric curves. \Sec{sec:motion_planning_as_constrained_optimization} lays out the mathematical foundation of our framework, and is followed
by the numerical solution method in \Sec{sec:numerical_solution} and computing an initial guess in \Sec{sec:initial_guess}. Evaluation of the framework and results are presented in \Sec{sec:evaluation_and_results}, followed
by concluding remarks and direction for future work in \Sec{sec:concluding_remarks}.

\section{Preliminary material}
\label{sec:preliminary_material}
In this section, we present the notation and some preliminary material that is relevant
to our formulation.
We begin by an analysis of motion of a nonholonomic mobile robot
moving on a plane. We then provide a brief introduction to parametric
curves and arc-length parameterization of curves. 

\subsection{Motion of a nonholonomic mobile robot moving on a plane}
\label{sec:motion_of_nonholonomic_robot_on_plane}
The configuration of a rigid body moving on a plane at any time $t$ can be completely specified
by specifying the position vector $\vet{r} = \left\{x(t),y(t) \right\}$ and orientation $\theta(t)$ 
of a body-fixed frame with
respect to a fixed reference frame. Suppose the rigid body starts from an initial configuration
at time $t = 0$ and reaches a final configuration at time $t = \tau$. To fully specify the 
motion of the body it is necessary to specify
the functions $x(t),y(t)$ and $\theta(t)$ on $I = [0,\tau]$. If this
body is a physical system, it cannot change its position instantaneously.
Further, since forces of infinite magnitude cannot be applied in the
real world, the acceleration of the body must be finite. Hence $x(t),y(t)$, 
and $\theta(t)$ must be at least $C^1$ on $I$.

If this rigid body has directional wheels, its motion should obey the following nonholonomic
constraint
\begin{equation}
\label{eq:wheeled_robot_constraint}
\dot{x} \sin\theta - \dot{y}\cos\theta = 0.
\end{equation}
Here dot, $(\dot{\,})$, represents derivative with respect to $t$. 
For motion planning, it is common to model a mobile robot as a wheeled rigid body
subject to above nonholonomic constraint,
and we will do the same. A motion of such a body can be
specified by specifying a travel time $\tau$ and a trajectory $\vet{r}$ for $t \in [0,\tau]$.
The orientation $\theta(t)$ can be computed from \Eq{eq:wheeled_robot_constraint}.
Essentially, $\theta(t) = \mbox{arctan2}(\dot{\ve{r}}(t))$.
If $\dot{\ve{r}}(t)$ is zero, which means the velocity is zero, then this equation cannot be used.
If the instantaneous velocity is zero at $t = t_0$, and non-zero in a neighborhood
of $t_0$, then $\theta(t_0)$ can be defined as a $\lim_{t \to t_0} \mbox{arctan2}(\dot{\ve{r}}(t))$.

\subsection{Parametric curves and the arc-length parameterization}
\label{sec:parametric_curves}
We present a brief introduction to parametric curves and the arc-length
parameterization. The reader can refer
to any book on differential geometry of curves for more details.

Let $q_a < q_b$ and $I = [q_a,q_b] \subset \mathbb{R}$.
A planar parametric curve is a mapping $\ve{r}:I \mapsto \mathbb{R}^2$.
If components of $\ve{r}$ are of class $C^1$, the vector space of functions with
continuous first derivatives, the tangent vector at $\ve{r}(q)$ for $q \in [q_a,q_b]$
is $\vedash{r}(q)$. In this section, we denote derivatives with respect to the
parameter $q$ by a prime $({\,}')$.

Let the length of a curve be denoted by $\lambda$, where
\begin{equation}
\label{eq:def_length}
\lambda = \int_{q_a}^{q_b} \norm{\vedash{r}(q)}dq.
\end{equation}
%
Define a function $s = s(q)$, which is the length of the curve between $[q_a, q]$. Then,
\begin{equation}
\label{eq:def_arc_length}
s(q) = \int_{q_a}^{q} \norm{\vedash{r}(q)}dq.
\end{equation}
%
Note that the integrand $\norm{\vedash{r}(q)}$ is non-negative
throughout $I$.  We make an assumption that it is zero
only at a finite number of $q$'s in $I$.  If $q$ represented
time, the physical interpretation is that the velocity is equal
to zero only at a finite number of discrete instants in time.
This assumption implies that $s$ is an increasing function of $q$. That is,
if $q_2 > q_1$, then $s(q_2) > s(q_1)$.  This, in turn, means
that for any given $s \in [0, \lambda]$, a unique $q = q(s)$ can be
found that corresponds to that $s$. If components of $\ve{r}$ are of class $C^1$,
then $\norm{\vedash{r}(q)}$ is continuous, and thus $s = s(q)$
is also in $C^1$.  Thus, $\frac{d s}{d q}$ is defined and is a continuous
function.  Obviously, $\frac{d s}{d q} = \norm{\vedash{r}(q)}$.

With the assumption above that $\norm{\vedash{r}(q)}$ can be zero
only at a finite number of $q$'s, it is possible to introduce the
arc-length parameterization.  For $s \in [0, \lambda]$ define
\begin{equation}
\widehat{\ve{r}}(s) = \ve{r}(q) \; \mbox{where}\; s = s(q).
\end{equation}
The function $\widehat{\ve{r}}$ is well-defined because for each
$s \in [0, \lambda]$ a unique $q$ can be found.
Using the chain-rule for differentiation,
$$
\frac{ d \widehat{\ve{r}}}{d s} = \frac{d \ve{r}}{d q} \frac{d q}{d s}.
$$
Now $\frac{d \ve{r}}{d q}$ exists and is continuous and $\frac{d q}{d s} =
\frac{1}{\frac{d s}{d q}} = \frac{1}{\norm{\vedash{r}(q)}}$
also exists (and is continuous) if $\norm{\vedash{r}(q)}$ is not zero.
Thus, at points where $\norm{\vedash{r}(q)} > 0$,
$$
\norm{\frac{ d \widehat{\ve{r}}}{d s}} = \norm{\vedash{r}(q)} / \norm{\vedash{r}(q)} = 1.
$$
On points where $\norm{\vedash{r}(q)} = 0$, $\norm{\frac{ d \widehat{\ve{r}}}{d s}}$
cannot be computed by the expression above.  However, the
choice that makes it continuous for all $s$ is 1.  This is analogous to
computing the limiting value of the orientation when velocity is zero
as shown earlier in this section.

\begin{figure}
\begin{center}
\includegraphics[scale=1,trim = 10 0 0 0]{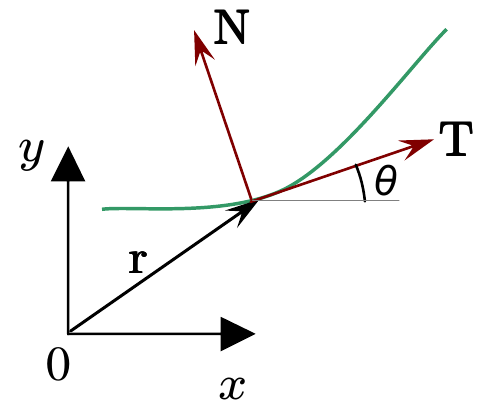}
\end{center}
\caption{Tangent and Normal to a curve}
\label{fig:tangent_and_normal}
\end{figure}

Symbolically, the curve has been parameterized by the arc-length.
Since $\norm{\frac{ d \widehat{\ve{r}}}{d s}} = 1$,
the tangent vector computed in the new parameterization is a unit vector.
The tangent vector is $\ve{T}(s)$ and the unit normal vector is $\ve{N}(s)$, where
\begin{equation}
\label{eq:def_tangent_normal}
\begin{split}
\ve{T}(s) &= \frac{ d \widehat{\ve{r}}}{d s}\\
\ve{N}(s) &= \frac{ \frac{d \ve{T}}{d s} }{\norm{ \frac{d \ve{T}}{d s} }}
\end{split}
\end{equation}
See Figure~\ref{fig:tangent_and_normal}.
The signed curvature $\kappa(s)$ is defined as
\begin{equation}
\label{eq:def_signed curvature}
\kappa(s) = \frac{ d \theta}{d s}
\end{equation}
where $\theta(s)$ is the tangent angle.
%

\section{Formulating motion planning as a constrained optimization problem}
\label{sec:motion_planning_as_constrained_optimization}
This section presents the mathematical formulation of our framework for planning
comfortable and customizable motion of a planar nonholonomic mobile robot.

The steps involved are: 
(1) formulating a discomfort cost functional (\Sec{sec:discomfort_cost_functional}) (2) dimensional analysis of cost functional (\Sec{sec:dimensional_analysis}) (3) formulating an informal problem statement (\Sec{sec:problem_statement}) (4) choosing an appropriate 
parameterization of the trajectory (\Sec{sec:parameterization_of_trajectory}), (5) choosing the function space to
which the trajectory should belong for the cost functional to be well-defined 
(\Sec{sec:function_spaces_for_finite_discomfort}), (6) analysis of boundary conditions to determine the boundary conditions
that should be imposed for the problem to be well-posed (\Sec{sec:analysis_of_boundary_conditions}), (7) choosing
a representation of obstacles and imposing constraints for obstacle avoidance (\Sec{sec:obstacle_avoidance}),
and finally, (8) formulating the full infinite-dimensional constrained optimization problem 
(\Sec{sec:full_nonlinear_constrained_optimization_problem}).

\subsection{The discomfort cost functional}
\label{sec:discomfort_cost_functional}
In \Sec{sec:comfort}, we
saw that for motion comfort, it is
necessary to have  
continuous
and bounded acceleration along the tangential and normal directions.
It is possible that the actual values of the bounds
on the tangential and normal components are different.
It is also desirable to keep jerk small and bounded.
We model user discomfort as
a weighted sum of the following three terms: total travel time, time
integral of squared tangential jerk and time integral of
squared normal jerk. Travel time is included because we make the 
justifiable assumption
that a user would prefer to reach a goal as fast as is consistent
with comfort. Thus, longer travel time implies greater discomfort.
We will see later in \Sec{sec:function_spaces_for_finite_discomfort} that
this cost functional is mathematically meaningful only when
both tangential and normal acceleration are continuous. Thus, 
we get continuous accelerations by construction. To keep 
accelerations within comfortable bounds, we impose explicit constraints on the maximum
and minimum values.

We construct a cost functional $J$ as follows:
\begin{equation}
\label{eq:cost_functional}
J = \tau ~+~ w_{\tny T} \int_{0}^\tau (\dddot{\bf r}\cdot{\bf T})^2 ~dt ~+~ w_{\tny N} \int_{0}^\tau (\dddot{\bf r}\cdot{\bf N})^2 ~dt.
\end{equation}
Here $\tau$ is the total travel time and ${\bf r}$ is the position of robot at time $t \in [0, \tau]$.
$\dddot{\bf r}$ represents the jerk. $\dddot{\bf r}\cdot{\bf T}$ and $\dddot{\bf r}\cdot{\bf N}$ are
the tangential and normal components of jerk respectively. We assume that $\vet{r}$ is smooth enough for the cost functional to be
well-defined. This means (at least) that the acceleration vector is continuous
and normal and tangential components of jerk are square integrable.

The term $\tau$ is necessary. If it is not included in the functional, the optimal
solution is to reach the destination at $\tau = \infty$ traveling at essentially zero speed in the limit
(except perhaps at the end-points where the speed is already specified). Thus,
minimizing just the integral terms will not lead to a good solution.

The weights ($w_{\tny T}$ and $w_{\tny N})$ are
non-negative known real numbers.  We separate tangential and normal jerk to
allow a choice of different weights ($w_{\tny T}$ and $w_{\tny N}$).

The weights serve two purposes. First, they act as scaling factors for
dimensionally different terms. Second, they determine the relative
importance of the terms and provide a way 
to adjust the robot's performance according to user preferences. For
example, for a wheelchair, some users may not tolerate high jerk and
prefer traveling slowly while others could tolerate relatively high jerks if they
reach their destination quickly. The typical values of weights will be chosen
using dimensional analysis.

\subsection{Dimensional analysis of cost functional and determination of characteristic weights}
\label{sec:dimensional_analysis}
Choosing the weights in an {\em ad hoc} manner does not provide weights that lead to similar comfort levels independent of the input (the boundary conditions). Moreover, since the different components of the total discomfort are
different physical quantities, choice of weights should reflect this. In other words, for the total discomfort to make physical
sense, the weights cannot be dimensionless numbers but should have physical units.  We determine
the weights using dimensional analysis~\citep{Langhaar_1951}. If the weights are chosen without
the dimensional analysis step, the optimal trajectory will be different just
by specifying the input in different physical units. In addition, using the same numerical weights
for different tasks will not lead to similar quantitative discomfort level.

All the physical quantities in the cost functional (time, jerk) depend on only two units $-$ length $L$ and time $T$.  From \Eq{eq:cost_functional} we see that $J$ has dimensions $L^0T^1$ due to the first term ($\tau$).  Thus $w_{\tny T}$ should have dimensions $T^6/L^{2}$.  Similarly, the dimensions of
$w_{\tny N}$ is $T^6/L^{2}$. Alternatively, since $T =  L/V$, $w_{\tny T}$ and $w_{\tny N}$ has dimensions
$L^4/V^{6}$.

We now determine the base values of weights analytically. 
The main idea behind determining the base values is that the
correct base values should keep the maximum speed below the maximum allowable
speed.  A user can then customize the weights by multiplying the base values
by a dimensionless constant that indicates user preference.

\subsubsection{Weight for tangential and normal jerk}
\label{sec:weight_tangential_and_normal_jerk}
We first determine $w_{\tny T}$.  Consider a one dimensional motion with
a trajectory that starts from origin and travels a distance $L > 0$ in an
unknown time $\tau > 0$.  The starting and ending speeds and accelerations are zero.
We choose the exact form of the trajectory to be a quintic polynomial in
time $t \in [0,\tau]$.  This choice uniquely determines the trajectory.
The reason we have chosen a quintic is that it minimizes integral of
squared jerk (a third derivative), just like a cubic spline minimizes integral
of squared second derivative.  Additionally, we choose the quintic to
satisfy the boundary conditions.

Let $s(t)$ be the distance traveled in time $t$.  It is easily seen that the
quintic
$$
s(t) = \frac{L t^3}{\tau^5} \left( 6 t^2 - 15 t \tau + 10 \tau^2 \right)
$$
satisfies all the boundary conditions.  For such a trajectory, the
discomfort functional is
$$
J = \tau + w_{\tny T} \int_{0}^{\tau} \dddot{s}(t)^2 dt =
\tau + \frac{720 L^2 w_{\tny T}}{\tau^5}.
$$

We do not know $\tau$ and $w_{\tny T}$ yet.  We first choose a $\tau$
that minimizes $J$ for all $w_{\tny T}$. This means
$$
\tau = \left( 3600 L^2 w_{\tny T} \right)^{1/6}.
$$
Obviously, choosing a large value of $w_{\tny T}$ will increase $\tau$,
which is natural because doing so penalizes jerk and would slow down the
motion.  We now choose a $w_{\tny T}$ so that the maximum
speed during the motion is $V$, a dimensional velocity scale.
It can be seen that the maximum speed occurs at $t = \tau/2$ and it
is
$$
\left( \frac{225}{2048} \right)^{1/3} \left( \frac{L^4}{w_{\tny T}} \right)^{1/6}.
$$
Hence we choose
\begin{equation}
\label{eq:wjt_base}
w_{\tny T} = \left( \frac{225}{2048} \right)^2 \frac{L^4}{V^6}. 
\end{equation}

The base value for the weight corresponding to the normal jerk
($w_{\tny N}$) is chosen to be the same.
We emphasize that both $w_{\tny T}$ and $w_{\tny N}$ will be present
in a real problem and the maximum speed constraint is imposed explicitly
rather than relying on weights.  The analysis done here is to get
dimensional dependencies of the base weight and reasonable proportionality
constants using a simple problem that can be treated analytically. If a different
problem is chosen, these base values will change.

\subsubsection{Factoring the weights for customization}
\label{sec:factoring_weights_for_customization}
In the preceding discussion, we determined the base values of weights
using simple analytical problems. We will
refer to these base values as $\widehat{w}_{\tny T}$ and $\widehat{w}_{\tny N}$.
Let $R_*$ be the minimum turning radius of the robot. For any given input, we
determine the characteristic length $L_*$ as max$(\Delta L, \pi R_*)$ where 
$\Delta L$ is the straight line distance between the start and
end points. The characteristic speed $V_*$ is the maximum allowable speed
of the robot. The base values of weights are then computed as
\begin{equation}
\label{eq:base_weights}
\widehat{w}_{\tny T} = \widehat{w}_{\tny N} = \left( \frac{225}{2048} \right)^2 \frac{L_*^4}{V_*^6}.
\end{equation}
The weights for the actual problem are chosen as a multiple of these
base weights where the multiplying factors $f_{\tny T}$
and $f_{\tny N}$ are chosen by a user.
\begin{equation}
\begin{split}
\label{eq:factor_weights}
w_{\tny T} &= f_{\tny T} \widehat{w}_{\tny T},\\ 
w_{\tny N} &= f_{\tny N} \widehat{w}_{\tny N}. 
\end{split}
\end{equation}

\subsection{Problem statement}
\label{sec:problem_statement}
We formulate motion planning as
a constrained optimization problem as follows:  
Given the appropriate
boundary conditions on position, orientation, and derivatives of position and orientation, bounds
on curvature, speed, tangential and normal accelerations, the weight factors $f_{\tny T}$ and $f_{\tny N}$ (\Eq{eq:base_weights}), and a representation of obstacles, find a trajectory that minimizes the cost functional representing user
discomfort \Eq{eq:cost_functional} such that the trajectory satisfies boundary conditions, respects bounds, and avoids obstacles

\begin{figure}[t!]
\begin{center}
	\subfigure[]{\label{fig:bc_illustrate}
	    \includegraphics[scale=0.6]{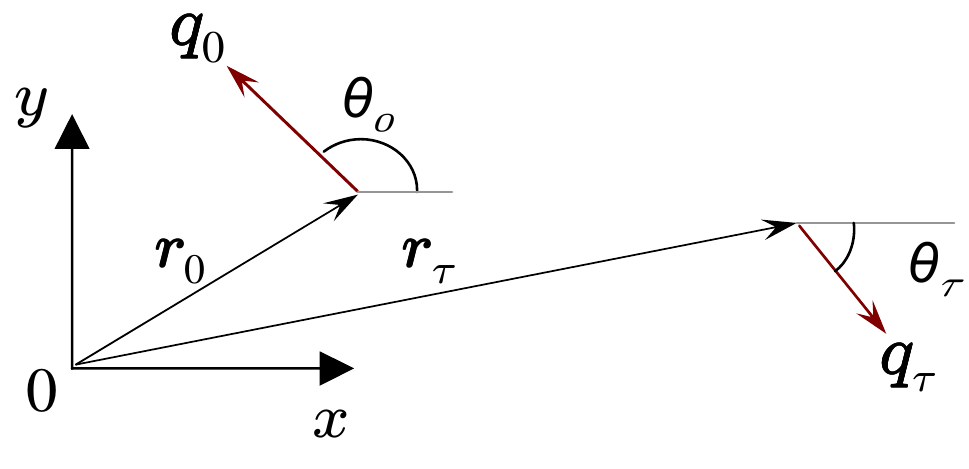}
	}
	\subfigure[]{\label{fig:multiple_paths}   
	    \includegraphics[scale=0.6,angle=0]{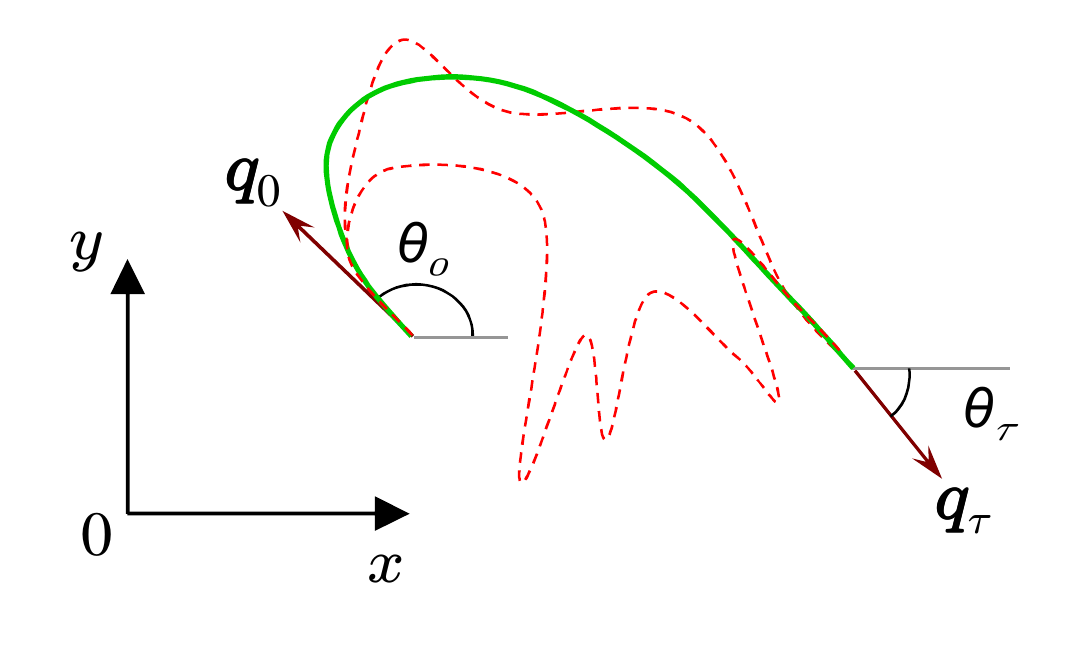}
	}
\end{center}
\caption
{Illustration of the optimization problem.}{\small{(a) The initial configuration of the robot at time $t = 0$ is given by
the position $\ve{r}_0$ and orientation $\theta_0$. The final configuration 
at time $t = \tau $ is given by the position $\ve{r}_{\tau}$ and orientation $\theta_{\tau}$. The speed at an end point, when non-zero, is necessarily along the vector $\ve{q}$. (b) There exist infinitely many
trajectories that satisfy boundary conditions and respect constraints, illustrated by the solid and dotted curves. Infinitely many of such trajectories will not result in comfortable motion, illustrated by the dotted curves. Our objective is to find a trajectory $\vet{r}$ that additionally minimizes the cost functional of \Eq{eq:cost_functional} and results in comfortable motion. Such a trajectory is illustrated by the solid curve. }}
\label{fig:problem_def}
\end{figure}

We model the robot as a rigid body moving on a plane,
subject to the nonholonomic constraint of \Eq{eq:wheeled_robot_constraint},
and assume that the robot moves with non-zero speed except at a finite
number of points.
Let the robot start from $\ve{r}_{0}$ at $t = 0$ and reach $\ve{r}_{\tau}$ 
in time $\tau$ (Figure~\ref{fig:problem_def}). 
From the discussion in \Sec{sec:motion_of_nonholonomic_robot_on_plane}, we see that to
fully specify the motion of the robot, we 
need only to specify a curve $\vet{r}$ on $t \in [0, \tau$]
such that the curve is at least $C^1$ continuous.
Henceforth, in this chapter, we will use {\em trajectory} to refer to
a function of robot position with respect to time.

We now transform the above problem description into a precise mathematical
problem statement using a general nonlinear constrained optimization
approach.

\subsection{Parameterization of the trajectory}
\label{sec:parameterization_of_trajectory}

Mathematically, one can use different primary variables to describe a
trajectory. For example, assuming the trajectory starts at zero
time, one way to describe a trajectory is to provide the final time
and the position vector as a function of time in between. Another way is
to provide the final time and specify the orientation and velocity as 
functions of time. Another way is to represent the
geometric path separately, using either position vector or orientation as
a function of arc-length. The velocity at each point on the path is provided
separately in this case. 

We have found that making the assumption that speed be non-zero except at
boundaries and expressing the trajectory solely in terms of {\em speed} and
{\em orientation} as functions of a {\em scaled} arc-length parameter leads to
relatively simple expressions for all the remaining physical quantities
(such as accelerations and jerks). We shall see below, that with this
parameterization, the primary variables (speed and orientation) and 
their derivatives enter the cost functional polynomially.
This would not have been the case if everything were expressed in terms 
of $\ve{r}$ as a function of time as we did in our previous work~\citep{Gulati_2009}.

In the following discussion, we implicitly assume that all the quantities 
have sufficient smoothness for expressions to be mathematically
meaningful.  In some cases, the derivatives appear not as point-wise
values but inside an integral sign.  In such a case we will assume
that the integrands belong to an appropriate space of functions so
that the integrals are well-defined. We explicitly state the
requirements on the regularity when posing the optimization problem later
in \Sec{sec:function_spaces_for_finite_discomfort}.

\subsubsection{Scaled arc-length parameterization}
\label{sec:scaled_arc_length_parameterization}
Let $u \in [0,1]$.  The trajectory is parameterized by $u$. The starting
point is given by $u = 0$ and the ending point is given by $u = 1$.
Let $\ve{r} = \ve{r}(u)$ denote the position vector of the robot
in the plane.  Let $v = v(u)$ be the speed. Both $\ve{r}$ and $v$ are 
functions of $u$. Let $\lambda$ denote the length of the trajectory. 
Since only the start and end positions are known, $\lambda$ cannot
be specified in advance. It has to be an unknown that will
be found by the optimization process. 

Let $s \in [0, \lambda]$ be the arc-length parameter.
We choose $u$ to be a scaled arc-length parameter where 
$u = \frac{s}{\lambda}$ so that the unknown constant $\lambda$ is not 
used in defining an unknown sized interval (as would be the case
if $u$ was chosen as the arc-length parameter).

In the following discussion we will see that the trajectory, 
$\vet{r}, t \in [0, \tau]$ is completely specified by the 
{\em trajectory length} $\lambda$, the {\em speed} $v = v(u)$, 
and the {\em orientation} or the tangent angle $\theta = \theta(u)$ 
to the curve. $\lambda$ is a scalar while speed
and orientation are functions of $u$. These are the three 
unknowns, two functions and one scalar, that will be determined by the optimization process.

Since speed is the rate of change of arc length, we have
\begin{equation}
\label{eq:ds_dt}
v(u) = \frac{ds}{dt}.
\end{equation}
Using $u = \frac{s}{\lambda}$ in the above equation, we get
\begin{equation}
\label{eq:du_dt}
\frac{du}{dt} =  \frac{v(u)}{\lambda}.
\end{equation}
This gives,
\begin{equation}
t = t(u) = \int_0^u \frac{\lambda}{v(u)} du.
\label{eq:t}
\end{equation}
If $v(u)$ is zero only at a finite number of points in $[0,1]$, then
$t(u)$ is well defined for all $u \in [0,1]$. 

\Eq{eq:t} is a key relation and gives us the means to 
convert between the time domain and scaled arc-length domain. 
We now introduce the third unknown -- the orientation or
the tangent angle to the curve $\theta = \theta(u)$.
Using the results of \Sec{sec:parametric_curves}, we can show that
\begin{equation}
\label{eq:dr_du}
\norm{\ve{r}'(u)} = \lambda.
\end{equation}
The tangent vector $\ve{r}'(u)$ to the curve $\ve{r}(u)$
is given by
\begin{equation}
\label{eq:drdu_vec_2}
\ve{r}'(u) = \norm{\ve{r}'(u)} \ve{T}(u) = \lambda \ve{T}(u)
\end{equation}
where $\ve{T}(u)$ is the tangent function.
\begin{equation}
\label{eq:T}
\ve{T}(u) = \left\{ \cos(\theta(u)), \sin(\theta(u)) \right\}.
\end{equation}
The braces $\left\{\right\}$ enclose the components of a 2D vector.

Thus, $\ve{r}(u)$ can be computed via the following integrals.
\begin{equation}
\label{eq:r_by_integral}
\ve{r}(u) = \ve{r}(0) + \lambda \left\{ \int_0^u \cth(u)\, d u, \int_0^u \sth(u)\, d u \right\}.
\end{equation}

Now, if $\theta(u)$ is known, $\ve{r}(u)$ can be computed from \Eq{eq:r_by_integral}.
If $v(u)$  and $\lambda$ are known, $t(u)$ can be computed from \Eq{eq:t}. 
Using these two, we can determine the function $\vet{r}, t \in [0, \tau]$.

We now have all the basic relations to use chain-rule to derive
expressions for all the physical quantities needed to pose the constrained
optimization problem.
We drop explicit references to $u$ as a function parameter
to keep the expression concise.

We compute first, second, and third derivatives of $\ve{r}$ with respect
to time.  These expressions are easily derived in one or two steps of
chain rule differentiation and so we do not present the intermediate steps in detail.
\begin{eqnarray}
\label{eq:r_t_derivs}
\dot{\ve{r}} &=& v \left\{ \cth, \sth \right\}\\
\ddot{\ve{r}} &=& \frac{v}{\lambda}( v' \left\{ \cth, \sth \right\} 
+ v \theta' \left\{ - \sth, \cth \right\})\\
\thder{\ve{r}} &=&
\frac{v}{\lambda^2} \left(
  (v'^2 + v v'' - v^2 \theta'^2)  \left\{ \cth, \sth \right\} \right) \nonumber \\  
 &+& \frac{v}{\lambda^2} \left((3 v v' \theta' + v^2 \theta'') \left\{ - \sth, \cth \right\}
\right)
\end{eqnarray}
From the equations above, the expressions for tangential acceleration $a_{\tny T}$ and
normal acceleration $a_{\tny N}$ are
\begin{equation}
\label{eq:aT}
a_{\tny T} = \ddot{\ve{r}} \cdot \ve{T} = \frac{v v'}{\lambda}
\end{equation}
\begin{equation}
\label{eq:aN}
a_{\tny N} = \ddot{\ve{r}} \cdot \ve{N} = \frac{v^2 \theta'}{\lambda}.
\end{equation}
The tangential jerk $j_{\tny T}$ is
\begin{equation}
\label{eq:jT}
j_{\tny T} = \thder{\ve{r}} \cdot \ve{T} = \frac{v}{\lambda^2} (v'^2 + v v'' - v^2 \theta'^2)
\end{equation}
and the normal jerk $j_{\tny N}$ is
\begin{equation}
\label{eq:jN}
j_{\tny N} = \thder{\ve{r}} \cdot \ve{N} = \frac{v^2}{\lambda^2} (3 v' \theta' + v \theta'').
\end{equation}
Here $\ve{N}$ is the direction normal to the tangent (rotated
$\frac{\pi}{2}$ anti-clockwise).
The signed curvature is given by
\begin{equation}
\label{eq:kappa}
\kappa(u) = \frac {\theta'}{\lambda}.
\end{equation}
The angular speed $\omega$ is given by
\begin{equation}
\label{eq:omega}
\omega(u) = \frac{\theta'v}{\lambda}.
\end{equation}
We can use the Equations~\ref{eq:jT} and \ref{eq:jN}. 
to express the total discomfort
\begin{equation}
\label{eq:J_total_t}
J(\ve{r}, \tau) =
\int_0^{\tau} d t
+ w_{\tny T} \int_0^{\tau} (\thder{\ve{r}} \cdot \ve{T})^2 d t
+ w_{\tny N} \int_0^{\tau} (\thder{\ve{r}} \cdot \ve{N})^2 d t
\end{equation}
in terms of $v$, $\theta$, and $\lambda$. First, we express the travel time $\tau$ 
in terms of the primary unknowns. 
\begin{equation}
\label{eq:tau}
\tau = \int dt = \int_0^1 \frac{d t}{d u} d u = \int_0^1 \frac{\lambda}{v} d u.
\end{equation}

Using a similar change of variables in the integration ($t \to u$), the total discomfort can be written as
\begin{equation}
\label{eq:J_total}
J(v, \theta, \lambda) =
\int_0^1 \frac{\lambda}{v} d u
+ w_{\tny T} \int_0^1 \frac{v}{\lambda^3} (v'^2 + v v'' - v^2 \theta'^2)^2 d u
+ w_{\tny N} \int_0^1 \frac{v^3}{\lambda^3} (3 v' \theta' + v \theta'')^2 d u.
\end{equation}
The first integral ($J_\tau$) is the total time, the second integral ($J_{\tny T}$) is
total squared tangential jerk, and the third integral ($J_{\tny N}$) is
total squared normal jerk.

Note that except for the term due to total travel time, the primary
variables $v$ and $\theta$ and their derivatives enter the total discomfort
expression polynomially.  

The discomfort $J$ is now a function of the primary unknown functions $v$, $\theta$,
and a scalar $\lambda$, the trajectory length.  All references to time $t$ have
disappeared. However, once the unknowns are found via optimization, we must
compute a mapping between $t$ and $u$.  This can be done using \Eq{eq:t}.

\subsection{Function spaces for $v$ and $\theta$ for a finite discomfort}
\label{sec:function_spaces_for_finite_discomfort}

Now that we have a concrete expression for the discomfort $J$ in \Eq{eq:J_total}, it can be
used to define the function spaces to which $v$ and $\theta$ can belong
so that the discomfort is well-defined (finite). This will, in turn, lead to
conditions on the physical quantities for safe and comfortable motion.
We have two distinct cases depending on whether the speed is zero at an end-point on not.

\subsubsection{Conditions for positive speeds}
\label{sec:conditions_for_positive_speeds}

Let $\Omega = [0,1]$ and $H^2(\Omega)$ be the Sobolev space of functions on $\Omega$
with square-integrable derivatives of up to order 2. Let $f : \Omega \to \mathbb{R}$.
Then
\begin{equation}
\label{eq:H2}
f \in H^2(\Omega)
\stackrel{\rm def}{\iff} \int_{\Omega} \left(\frac{d^j f}{d x^j}\right)^2 d x < \infty
\;\forall\; j = 0,1,2.
\end{equation}

First, we show that if $v, \theta \in H^2(\Omega)$, then the integrals of
squared tangential and normal jerk are finite. Using the
Sobolev embedding theorem~\citep{Adams_2003} it can be shown that if
$f \in H^2(\Omega)$, then $f' \in C^0(\Omega)$ and by extension
$f \in C^1(\Omega)$. Here $C^j(\Omega)$ is the space of
functions on $\Omega$ whose up to $j^{th}$ derivatives are
bounded and continuous.  Thus, if $v, \theta \in H^2(\Omega)$,
then all the lower derivatives are bounded and continuous.
Physically this means that quantities like the speed, acceleration, and
curvature are bounded and continuous $-$ all desirable properties
for comfortable motion.

Expanding all the jerk related terms in \Eq{eq:J_total},
bounding all the non-second derivative terms by a constant using
the results from the Sobolev embedding theorem, we immediately see that
the jerk part of discomfort is finite if $v, \theta \in H^2(\Omega)$.  This is a sufficient
condition only and not a necessary one as we shall see below.

We also need that the inverse of $v$ be integrable so that $J_\tau$ is finite.  This is trivially
true if $v$ is uniformly positive, that is, $v \ge \overline{v} > 0$ for some
constant positive $\overline{v}$ throughout the interval $[0, 1]$.  However,
$v$ can be zero at one or both end-points because of the imposed
conditions.  \Sec{sec:analysis_of_boundary_conditions} analyzes the boundary conditions in detail.
Here we assume that speed on both end-points is positive.
The cases with zero end-point speed are treated below in \Sec{sec:conditions_for_zero_speeds}.

Thus, consider the case that $v$ is positive on both end-points. Since $v$ is speed and always non-negative, it can
approach zero from above only. We make a justifiable assumption that $v$ can
be zero only at end-points if at all and not in the interior.  Otherwise, the
robot would stop and then start again.  This
is costly for discomfort since it increases travel time and leads to
acceleration and deceleration.  Of course, we {\em can} choose a motion in
which $v = 0$ in the interior and it can still be a valid motion
with finite discomfort. The assumption is that the trajectory that
actually minimizes discomfort will not have a halt in between.  Thus, if
$v > 0$ on end-points, it remains uniformly positive in the interior
and the discomfort is finite.

\subsubsection{Conditions for zero speed on boundary}
\label{sec:conditions_for_zero_speeds}

Consider the case in which $v(0) = 0$.  The case
$v(1) = 0$ can be treated in a similar manner. If $v(0) = 0$, $\frac{1}{v}$ must not blow up
faster than $\frac{1}{u^p}$ where $p < 1$.  This is to keep $J_\tau$
finite.  This can be seen as follows. Lets assume $v(u) = u^p$ for some $p > 0$ (so that $v(0) = 0)$.
This implies that $J_\tau = \frac{\lambda}{1-p}$ provided $p < 1$, otherwise it
is not defined.

For simplicity, assume a 1D motion so that $\theta(u) \equiv 0$.
Then $J_{\tny T} = \frac{1}{\lambda^3}\frac{(1-2 p)^2 p^2}{5 p - 3}$ provided $p > \frac{3}{5}$. Taking
all conditions into account, if $v(0) = 0$, the discomfort is finite if $v(u)$ behaves
like $u^p$ where $\frac{3}{5} < p < 1$.  However, in such a case,
$\int_0^1 v''^2 d u = \frac{(-1+p)^2 p^2}{2p-3}$ is defined and finite
only if $p > 3/2$.  This conflicts with the assumption that
$v \in H^2(\Omega)$.  Thus, we can have a finite discomfort even if
$v \notin H^2(\Omega)$.  
We see that the reason for this is
the zero speed boundary condition, which leads to $\int_0^1 v^3 v''^2 du$
being finite for $\frac{3}{5} < p < 1$ even though $\int_0^1 v''^2 du$ (which
is the highest order term in $J_{\tny T}$) is not finite for
such a range of $p$.

If we look at the integral $J_{\tny T} = \frac{1}{\lambda^3}\frac{(1-2 p)^2 p^2}{5 p - 3}$
carefully, we see that it can be finite even if $p < \frac{3}{5}$, provided
$p = \frac{1}{2}$.  This is a special case because $v v'' + v'^2$ is identically
zero for such a $p$ and tangential jerk discomfort is finite for a 1-D motion.

For a mathematically meaningful problem
we must treat zero speed boundary conditions separately from
non-zero speed boundary condition.  This analysis will be
done in more detail in Sections~\ref{sec:analysis_of_boundary_conditions}
and~\ref{sec:numerical_solution}
which are focused on boundary conditions and appropriate
singular finite elements respectively.

\subsubsection{Summary}
\label{sec:function_spaces_for_finite_discomfort_summary}
To summarize, the total discomfort is finite if $v, \theta \in H^2(\Omega)$
and the inverse of $v$ is integrable.  Inverse of $v$ is integrable if $v$ is
uniformly positive in $[0,1]$.  If zero speed boundary conditions are
imposed, we will have to choose $v$ outside $H^2(\Omega)$.
In such a case, at $u=0$, it is sufficient that $v$ approaches zero as $u^p$ where
$\frac{3}{5} < p < 1$ or $p = \frac{1}{2}$.  For the right end point, where $u=1$, replace $u$ with
$(1-u)$ in the condition.
We do not lose higher regularity of $v$ throughout the interval $\Omega$
just because $v \notin H^2(\Omega)$.  Assume $v > 0$ in the interior,
as justified above.  Then $v \ge \overline{v} > 0$ in $\Omega_\delta
\stackrel{\rm def}{=} [\delta, 1 - \delta]$ where $\delta = \delta(\overline{v}) > 0$.
Thus $v \in H^2(\Omega_\delta)$ is necessary to keep total discomfort finite.
This implies continuity and boundedness of velocity and acceleration
in $\Omega_\delta\; \forall \delta > 0$.

\subsection{Analysis of boundary conditions}
\label{sec:analysis_of_boundary_conditions}

The expression for the cost functional $J$ in \Eq{eq:J_total} shows that
the highest derivative order for $v$ and $\theta$ is two.  Thus, for
the boundary value problem to be well-posed we need two boundary
conditions on $v$ and $\theta$ at each end-point $-$ one on the
function and one on the first derivative.

We also have to impose that
the robot move from a specific starting point to a specific
ending point. This condition is a set of two equality constraints
on $\lambda$ and $\theta$ based on \Eq{eq:r_by_integral}.  If
the motion is from positions $\ve{r}_0$ to $\ve{r}_\tau$, then
\begin{equation}
\label{eq:r_by_integral_constr}
\ve{r}_\tau - \ve{r}_0 = \lambda \left\{ \int_0^1 \cth\, d u, \int_0^1 \sth\, d u \right\}.
\end{equation}

We now relate the mathematical requirement on $v$ and $\theta$ boundary values above
to expressions of physical quantities.
We do this for the starting point only.  The ending point relations
are analogous.

\subsubsection{Positive speed on boundary}
\label{sec:positive_speed_on_boundary}
First, consider the case when $v > 0$ on the starting point.
The speed $v$ needs to be specified, which is quite natural.  The $u$-derivative
of $v$, however, is not tangential acceleration.  The tangential acceleration
is the $t$-derivative and is given by \Eq{eq:aT}. It is $\frac{v v'}{\lambda}$.  Here $v$ is known
but $\lambda$ is not. Thus specifying
tangential acceleration gives us a constraint equation and not directly
a value for $v'(0)$.  This is imposed as an equality constraint.
Similarly, fixing a value for $\theta$ on starting point is natural.
We ``fix'' the values of $\theta'(0)$ by fixing the signed curvature $\kappa = \frac{\theta'}{\lambda}$.
As before, this leads to an equality constraint relating
$\theta'(0)$ and $\lambda$ if $\kappa \neq 0$.  Since choosing a meaningful non-zero
value of $\kappa$ is difficult, it is natural to impose $\kappa = 0$.  In this
case $\theta'(0) = 0$ can be imposed easily.

\subsubsection{Zero speed on boundary}
\label{sec:zero_speed_on_boundary}
We now discuss the $v = 0$ case.  If $v(0) = 0$, then, as seen in \Sec{sec:conditions_for_zero_speeds},
$v(u)$ must behave like $u^p$ for $\frac{3}{5} < p < 1$ or $p = \frac{1}{2}$ near $u=0$ and
$v'(u) \sim u^q$ for $-\frac{2}{5} < q < 0$ or $q = -\frac{1}{2}$ respectively.  This means the $\lim_{u \to 0}v'(u)$ is
infinite.  This leads to a difficulty in analyzing the expression for
the tangential acceleration ($\frac{v v'}{\lambda}$) without using limits.  We prove that if $v \sim u^p$
at boundary, then the tangential acceleration is 0 if $\frac{3}{5} < p < 1$
and it is finite but non-zero if $p = \frac{1}{2}$.
If $v(u) \sim u^p$, then, $v v' \sim u^{2p-1}$.  If
$\frac{3}{5} < p < 1$, it means $\frac{1}{5} < 2p - 1 < 1$.  Thus as
$u \to 0$, $v v' \to 0$ because of the allowable range of $p$.  If $p = \frac{1}{2}$,
$v v'$ behaves like a positive constant as $u \rightarrow 0$.  Hence $p = \frac{1}{2}$
corresponds to non-zero tangential acceleration.

We still have to decide with what strength does $v'(u)$ tend to infinity
at an end-point. If $v = 0$ and $a \neq 0$,
it is clear that $v'(u) \sim u^{-1/2}$.
If $v = 0$ and $a = 0$,
the analysis above has only shown that 
$\lim_{u \to 0} v(u) v'(u) = 0$, and $\lim_{u \to 0} v'(u) = \infty$.
In this case, we need to use the time domain.  The reason we have such a
singularity is because of working in the arc-length domain.
Consider starting from origin with zero speed and acceleration
at zero time ($t$) in 1D. Expanding the distance traveled ($s$) as a function
of time, we see that
\begin{equation*}
\label{eq:s_taylor}
s(t) = 0 + 0 t + \frac{1}{2} 0 t^2 + \frac{1}{6} j t^3 + \ldots.
\end{equation*}
Here $j > 0$ is the jerk at $t = 0$.  We ignore the higher order terms.
Then, to the lowest power of $t$, the speed as a function of $t$ is
$$
v(t) = \frac{1}{2} j t^2.
$$
Eliminating $t$ to relate $v$ and $s$, we get
$$
v = \frac{6^{2/3}}{2} j^{1/3} s^{2/3}.
$$
Now $s = \lambda u$ because $u$ is the scaled arc-length parameter. Using this we
get $v = C u^{2/3}$, where all the constants are absorbed in $C$.  Thus,
$v(u) \sim u^p$ for $p = \frac{2}{3}$.  This value of $p$
is within the acceptable range of $p$, the open interval ($\frac{3}{5}, 1$).
This also tells us that
\begin{equation}
\label{eq:vdash_strength}
v'(u) \sim u^{-1/3}
\end{equation}
is the appropriate
strength of the singularity.

\subsubsection{Summary}
\label{sec:analysis_of_boundary_conditions_summary}
To summarize, we must specify following boundary conditions at
both end points: position,
orientation, curvature, and speed. If a specified speed is non-zero, the tangential
acceleration must be specified.  If the speed is zero, the tangential
acceleration can be zero.  If tangential
acceleration is non-zero, it must be positive if it is 
the starting point or must be negative if it is the ending point.

\subsection{Obstacle avoidance}
\label{sec:obstacle_avoidance}

For safe motion, it is necessary that the robot avoid
obstacles while navigating.  Simply speaking, obstacles are regions in the
plane of motion through which the geometric path must not
pass.  Obstacles can be represented in a variety of ways. For example, convex
polygons, rectangular cells, simple closed shapes like ellipses,
or level sets of implicitly defined simple functions of two
arguments are some possibilities.

\subsubsection{Modeling obstacles as star-shaped domains}
\label{sec:obstacles_as_star_shaped_domains}
We have chosen to model the ``forbidden'' region formed by the obstacles as a union
of star-shaped domains with boundaries that are closed curves with piecewise continuous second derivative.
A set in $\mathbb{R}^n$ is called a star-shaped domain if there exists at least one point
${\bf x}_0$ in the set such that the line segment connecting ${\bf x}_0$ and ${\bf x}$ lies in the set
for all ${\bf x}$ in the set. Intuitively this means that there exists at least one point
in the set from which all other points are ``visible''. We will refer to such a point 
${\bf x}_0$ as a \emph {center} of the star-shaped domain.

\begin{figure}
\begin{center}
\includegraphics[scale=0.6]{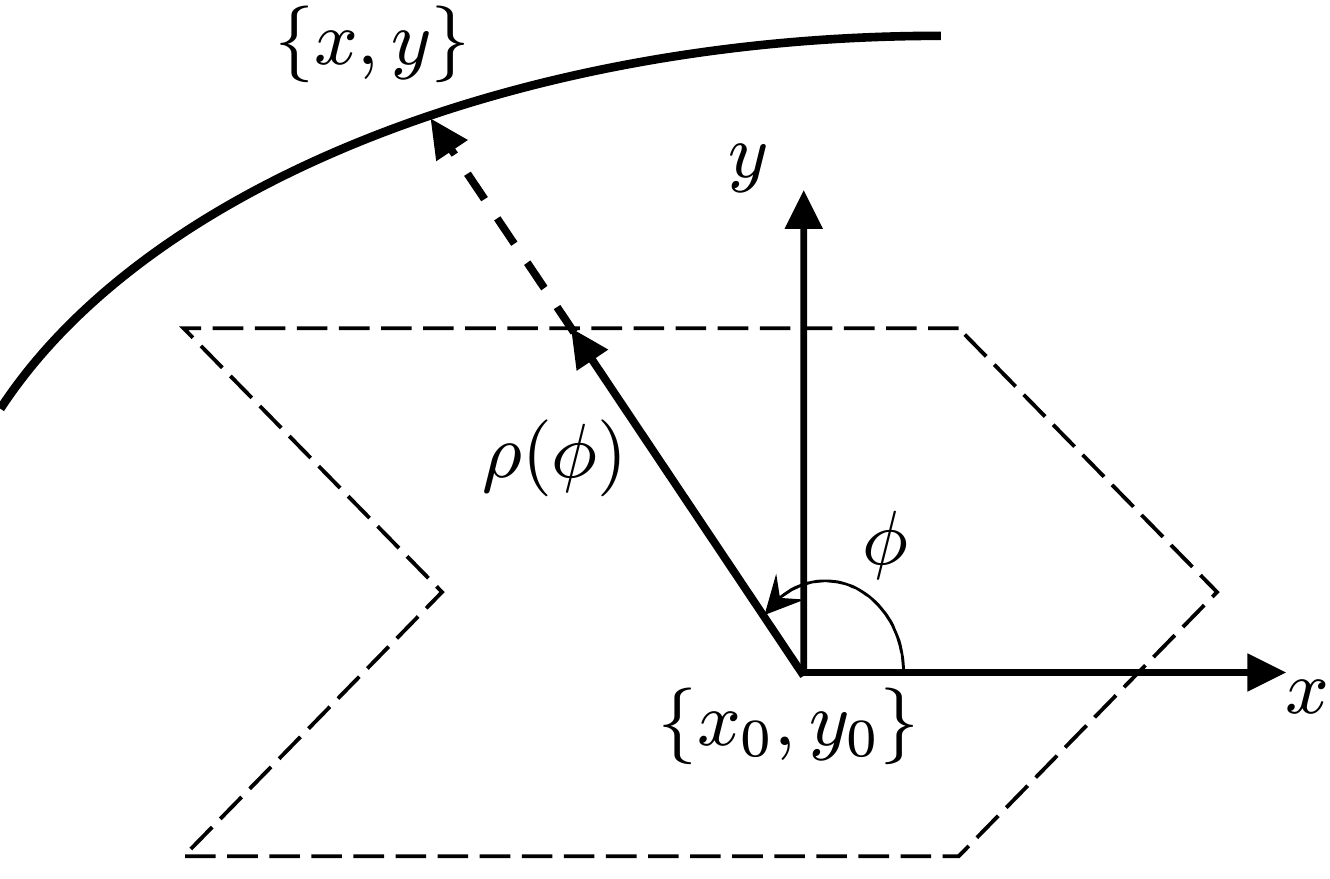}
\end{center}
\caption
{Notation for star-shaped obstacles.}{\small{A non-convex star-shaped obstacle is shown with
its ``center'' $\{ x_0, y_0 \}$ and a distance function $\rho = \rho(\phi)$.
The distance function gives a single point on the boundary for $\phi \in [0,2\pi]$.
The robot trajectory must lie outside the obstacle.}}
\label{fig:star_shaped}
\end{figure}

The choice of using star-shaped domains is made so that
each point on the boundary of an obstacle can be treated
as coming from a well-defined function in polar coordinates
centered within the particular obstacle.  See \Fig{fig:star_shaped}.
This also allows treatment of non-convex obstacles without subdividing
them into a union of convex shapes.  A big advantage is that
we reduce the number of imposed constraints since the number of
inequality constraints is proportional to the number of obstacles.
This leads to a faster optimization process.

This approach is a special case of using level sets of an implicitly defined function
as an obstacle boundary.  What is different here is that given the
description of the boundary in polar coordinates, which is easy to specify
for common shapes, we {\em construct} an implicit function (see the
following section).  This is done based on the
assumption that the boundary encloses a star-shaped region.
The piecewise smoothness property is required to impose the obstacle constraint
in a numerical optimization method.  Since up to second derivative
of constraint can be required, the obstacle boundary should also
be smooth to that order (or at least piecewise smooth).

If an obstacle is not
star-shaped, our framework can still handle it if
it can be expressed as a finite union of piecewise smooth star-shaped domains.
Efficient algorithms to decompose any polygon into a finite number
of star-shaped polygons exist~\citep{Avis_1981}.

\subsubsection{Incorporating constraints for obstacle avoidance}
\label{sec:obstacle_avoidance_constraints}

We now derive a function for the inequality constraint that a given
point in the plane is not inside the boundary of one star-shaped obstacle.
It is easy to extend this to multiple points
and multiple obstacles by just repeating the inequality with different
parameters.

Let an obstacle be specified by its boundary in polar coordinates
that are centered at $\ve{r}_0 = \{x_0, y_0\}$.  Each $\phi \in [0, 2\pi)$ gives a
point on the boundary using the distance $\rho(\phi)$ from the
obstacle origin. The distance function
$\rho$ must be periodic with a period $2 \pi$. See \Fig{fig:star_shaped}.

\begin{figure}
\begin{center}
\subfigure[Obstacle shape]{\label{fig:star_shape_rho}
\includegraphics[scale=0.4]{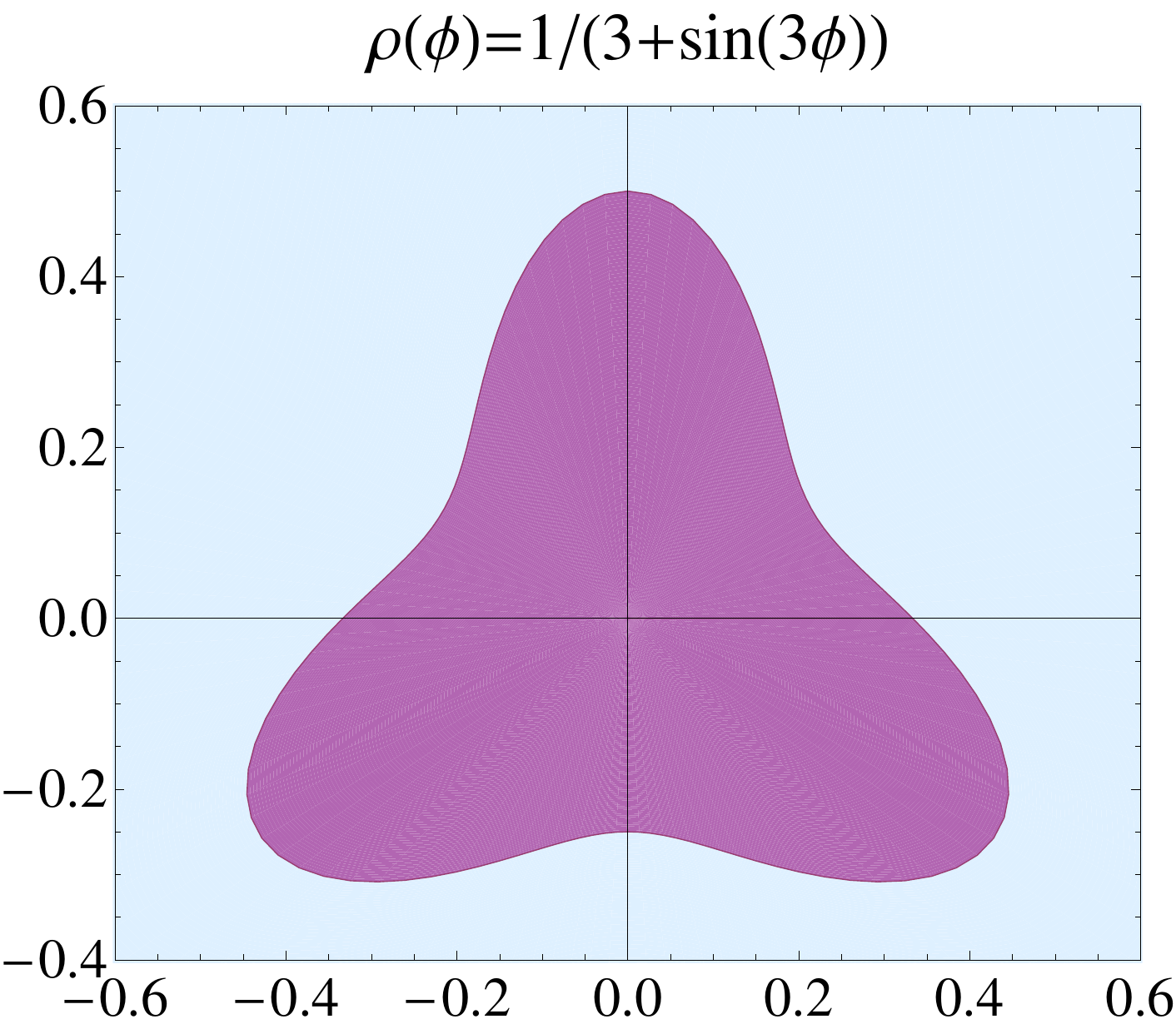}}
\subfigure[Obstacle as a 1-D curve]{\label{fig:star_shape_flat}
\includegraphics[scale=0.42]{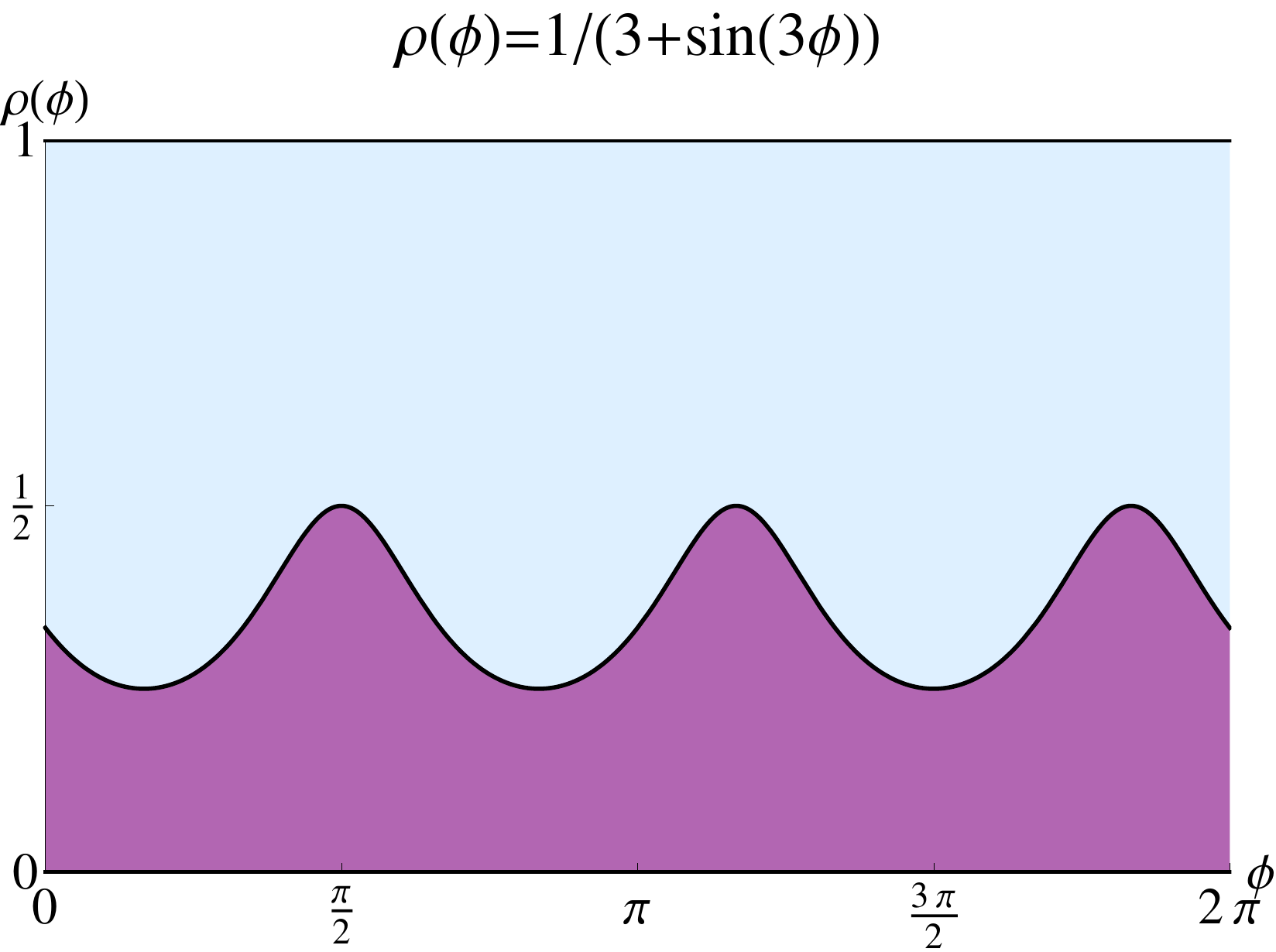}}\\
\subfigure[Surface plot of constraint]{\label{fig:star_shape_implicit}
\includegraphics[scale=0.42]{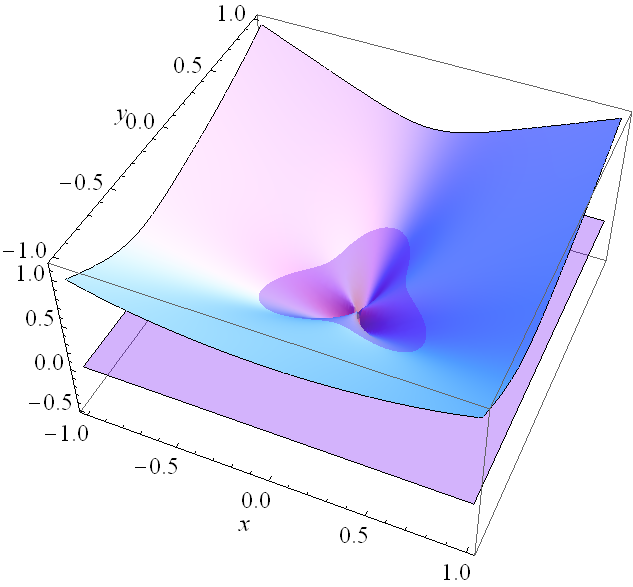}}
\subfigure[Level sets of constraint]{\label{fig:star_shape_contours}
\includegraphics[scale=0.5]{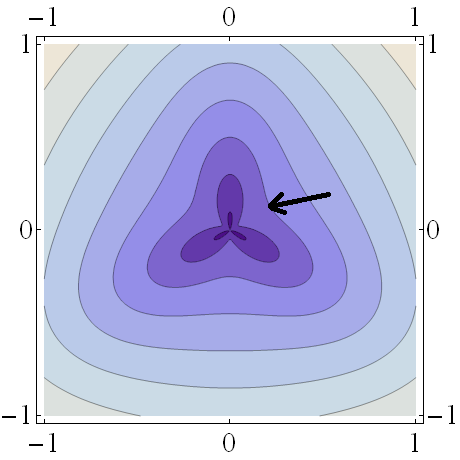}}
\end{center}
\caption
{Obstacle and constraint plots.}{\small{
The figures show an obstacle in polar coordinates in (a), and its
1-D representation in (b).  The region with darker shade is the interior
and a feasible trajectory must not pass through it. The surface plot
of the corresponding constraint function $C(\ve{r})$ of \Eq{eq:star_shaped}
is shown in (c) and its level set is shown in (d). The arrow marks
the zero level set, which is the obstacle boundary.
}}
\label{fig:star_shape_example}
\end{figure}

Suppose we want a point $\ve{r} = \{x, y\}$ to be outside the obstacle boundary.
Define $C(\ve{r})$ as
\begin{equation}
\label{eq:star_shaped}
C(\ve{r}) = \norm{\ve{r} - \ve{r}_0}_2 - \rho( \mbox{arctan2}(\ve{r} - \ve{r}_0))
\end{equation}
where the subscript 2 refers to the Euclidean norm.
It is obvious that $C(\ve{r}) \ge 0 \iff$ the point $\ve{r}$ is outside the obstacle.
This can be seen using a 1D graph of $\rho(\phi)$.  For example, let an
obstacle be represented as shown in \Fig{fig:star_shape_rho}.  \Fig{fig:star_shape_flat}
shows the same obstacle flattened out as a 1D curve.  Then $C(\ve{r})$ is
positive in the top region and negative below.  The star-shaped property leads to
a single-valued curve $\rho(\phi)$ when flattened like this.  The vector $\ve{r}$
is related to the primary variables in our optimization
problem using \Eq{eq:r_by_integral}.

\subsubsection{Derivatives of obstacle avoidance constraint}
\label{sec:obstacle_avoidance_constraint_derivatives}

We will need derivatives of $C(\ve{r})$ with respect to $\ve{r}$ for incorporating
$C(\ve{r}) \ge 0$ as a constraint in the trajectory optimization problem.  Here
$\ve{r}$ is any point on the path that we want to lie outside a given
obstacle.  We can derive the following expressions for first and second derivatives
of $C(\ve{r})$.  The derivatives of $\rho$ below are evaluated at
$\phi = \mbox{arctan2}(\ve{r} - \ve{r}_0)$.  To simplify the expressions,
$x, y, \ve{r}$ refer to the offsets from obstacle origin $\ve{r}_0$
instead of absolute positions in the plane.
\begin{equation}
\label{eq:star_shaped_deriv1}
\frac{\partial C}{\partial \ve{r}} = \frac{\ve{r}}{\norm{\ve{r}}_2}
- \rho'(\phi) \frac{\{-y,x\}}{\norm{\ve{r}}^2_2}
\end{equation}

\begin{equation}
\label{eq:star_shaped_deriv2}
\frac{\partial^2 C}{\partial \ve{r}^2} = 
\frac{1}{\norm{\ve{r}}^3_2}
\left(1 - \frac{\rho''(\phi)}{\norm{\ve{r}}_2}\right)
\begin{bmatrix}
 y^2  & - x y\\
 -x y & x^2\\
\end{bmatrix}
- \frac{\rho'(\phi)}{\norm{\ve{r}}^4_2}
\begin{bmatrix}
 2 x y & y^2 - x^2\\
 y^2 - x^2 & - 2 x y\\
\end{bmatrix}
\end{equation}
Obviously, the second derivative is a $2 \times 2$ matrix.

The constraint function $C(\ve{r})$ is piecewise differentiable for all
$\ve{r}$ except at a single point $\ve{r} = \ve{r}_0$.  If $\ve{r} = \ve{r}_0$
by chance, which is easily detectable, we know that the $\ve{r}$ is inside the obstacle and can
perturbed to avoid this undefined behavior. Note that  $C(\ve{r})$ remains bounded inside
the obstacle.  It is the derivatives that are not bounded as $\ve{r} \rightarrow \ve{r}_0$
\Fig{fig:star_shape_implicit} shows a surface plot of $C(\ve{r})$
for the obstacle shown in \Fig{fig:star_shape_rho}.  The contours
of constant values are shown in \Fig{fig:star_shape_contours}.

\subsubsection{Incorporating robot shape}
\label{sec:robot_shape}
The discussion on obstacle avoidance constraints so far has assumed that the robot
is a point.
In reality, the robot is not a point. To impose obstacle avoidance constraints
in this case, the robot can be modeled as a closed curve that encloses the projection of its boundary
in the plane of motion. We can choose a set of points on this curve and impose the constraint that all these points
be outside all obstacles. The distance between any pair of points can be 
smaller
than the smallest obstacle. We have currently not implemented this and this is part
of future work.

\subsection{The full nonlinear constrained optimization problem}
\label{sec:full_nonlinear_constrained_optimization_problem}

We now summarize the nonlinear and constrained trajectory optimization problem
taking into account all input parameters, all the boundary conditions,
and all the constraints. This is the ``functional'' form of the problem (posed in function
spaces).  We will present an appropriate discretization procedure valid for
all input combinations in the next chapter.

Minimize the discomfort functional $J$, where
$$
J(v, \theta, \lambda) =
\int_0^1 \frac{\lambda}{v} d u
+ w_{\tny T} \int_0^1 \frac{v}{\lambda^3} (v'^2 + v v'' - v^2 \theta'^2)^2 d u
+ w_{\tny N} \int_0^1 \frac{v^3}{\lambda^3} (3 v' \theta' + v \theta'')^2 d u,
$$

given the following boundary conditions for both starting point and ending point
\begin{itemize}
	\item position ($\ve{r}_0$, $\ve{r}_\tau$),
	\item orientation ($\theta_0$, $\theta_\tau$),
	\item signed curvature ($\kappa_0$, $\kappa_\tau$),
	\item speed ($v_0 \ge 0$, $v_\tau \ge 0$),
	\item tangential acceleration ($a_{{\tny T},0}$, $a_{{\tny T},\tau}$),
\end{itemize}
and constraints on allowable range of
\begin{itemize}
	\item speed ($v_{min} = 0, v_{max}$),
	\item tangential acceleration ($a_{{\tny T},min}, a_{{\tny T},max}$),
	\item normal acceleration ($a_{{\tny N},min}, a_{{\tny N},max}$),
	\item angular speed ($\omega_{min}, \omega_{max}$),
	\item curvature, if necessary ($\kappa_{min} = 0, \kappa_{max}$),
\end{itemize}
and
\begin{itemize}
\item number of obstacles $N_{obs}$, 
\item locations of obstacles $\{ \ve{c}_i \}_{i=1}^{N_{obs}}$ 
\item representation of obstacles that allows computation of $\{ \rho_i(\phi)\}_{i=1}^{N_{obs}}$, for $\phi \in [0,2\pi)$ 
\end{itemize}
and
\begin{itemize}
\item an initial guess for $(v(u), \theta(u), \lambda)$, in $u \in [0,1]$, 
\item weights $w_{\tny T} > 0$ and $w_{\tny N} > 0$.
\end{itemize}

The constraint on starting and ending position requires that
$$
\ve{r}_\tau - \ve{r}_0 = \lambda \left\{ \int_0^1 \cth\, d u, \int_0^1 \sth\, d u \right\}
$$
Staying outside all obstacles requires that
$$
\norm{\ve{r}(u) - \ve{c}_i}_2 - \rho_i( \mbox{arctan2}(\ve{r}(u) - \ve{c}_i)) \ge 0
\;\forall \; i \in 1, \ldots, N_{obs}, \mbox{ and} \; \forall \; u \in [0,1]
$$
where
$$
\ve{r}(u) = \ve{r}(0) + \lambda \left\{ \int_0^u \cth\, d u, \int_0^u \sth\, d u \right\}.
$$
As a post-processing step, we compute time $t$ as a function of $u$ using
$$
t = t(u) = \int_0^u \frac{\lambda}{v(u)} du
$$
and convert all quantities ($v, \theta, \ve{r},$ and their derivatives) from
$u$ domain to $t$ domain.

\section{Numerical solution}
\label{sec:numerical_solution}
\begin{figure}
\begin{center}
\includegraphics[scale=0.5,trim = 50 120 50 100]{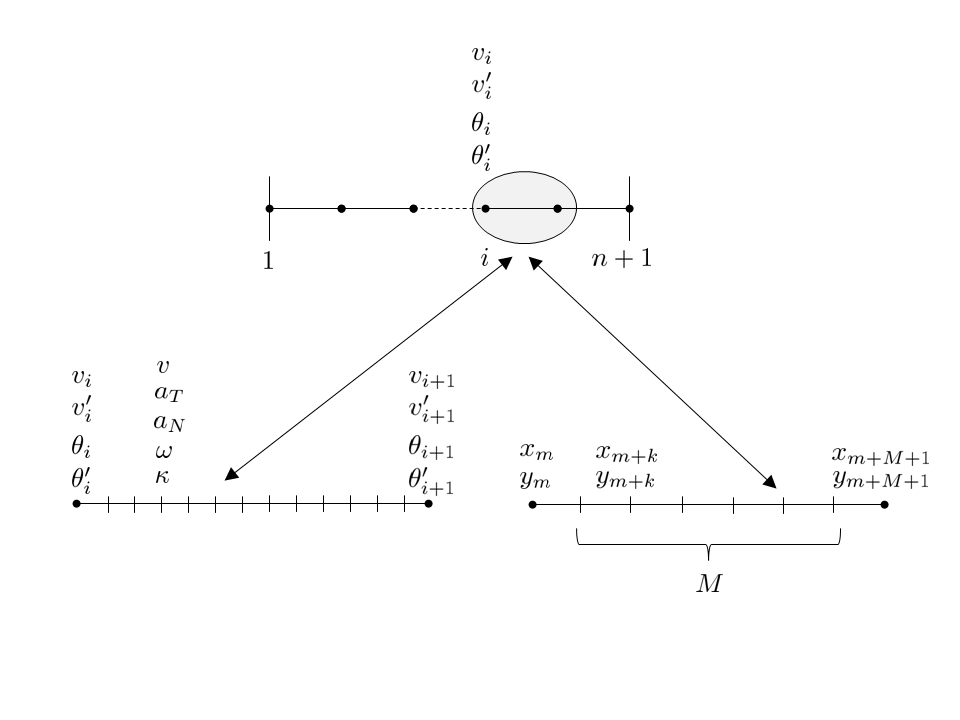}
\end{center}
\caption{Finite-element mesh for the optimization problem.}{\small The figure at the top shows the finite-element 
mesh consisting of $n$ elements and $n+1$ nodes. There are 4 primary unknowns at each node -- speed, orientation, and their derivatives with respect to the scaled-arc length parameter $u$. The figure on the bottom left shows how dynamic constraints are imposed. $P$ points are chosen in the interior of each element, and speed, angular speed, curvature, and tangential and normal accelerations are computed at each of these points in terms of the primary unknowns. Bounds are then imposed as constraints on these computed values. The figure on the bottom right shows how obstacle constraints are imposed. Addition variables $(x,y)$ representing position are introduced at $M$ points in the interior of each element and
at each node. Each $(x,y)$ pair is related to each of its neighboring pairs by a constraint. For obstacle avoidance, a constraint is imposed on each $(x,y)$ pair so that it is stays outside all obstacles.}
\label{fig:mesh}
\end{figure}

The optimization problem posed in \Sec{sec:full_nonlinear_constrained_optimization_problem}
is infinite dimensional since it is posed on infinite dimensional function spaces.
This means that we must discretize it as a finite dimensional
problem before it can be solved numerically. We saw in \Sec{sec:analysis_of_boundary_conditions} that we must be able to impose
two kinds of boundary conditions. In the first kind, the problem is set in
the Sobolev space of functions whose up to second derivatives are square integrable.
In the second kind, we must allow functions that are singular at
the boundary (with a known strength) but still lie in the same Sobolev space
in the interior.
 Keeping the problem setting and
requirements mentioned above in mind, it is natural to use the Finite Element Method (FEM) to
discretize it.

For the first kind of boundary conditions, where speed is non-zero
at both boundaries, $v(u)$ must belong to $H^2(\Omega)$. 
It is natural to use the basis functions in $C^1(\Omega)$, the space of
functions that are continuous and have continuous first derivatives.
In accordance with
standard FEM practice~\citep{Hughes_2000}, we choose
the cubic Hermite shape functions to discretize $v(u)$.
For the second kind of boundary conditions, when either one or both 
the boundary points have
zero speed specified, we must allow functions that are singular at
the boundary (with a known strength) but still belong to $H^2(\Omega)$
in the interior. We use special shape functions for $v(u)$ on
the boundary element where speed is zero so that $v(u)$ has the 
appropriate strength of
singularity at the boundary, and use regular cubic Hermite shape 
functions in the interior. For both kinds of boundary
conditions, we use
cubic Hermite shape functions for $\theta(u)$ on all elements.

With the above choice of shape functions, an $n$ element mesh 
for the problem consists of the following unknowns 
at each node -- $v, v', \theta, \theta'$ resulting in a total of
$4(n + 1)$ unknowns. In addition, the path length $\lambda$ is also
an unknown.
For optimization, the values of the objective function, its gradient and Hessian, the values of constraints,
and the gradient and Hessian of each constraint are required. For efficiency,
it is desirable that objective function and constraint Hessians be sparse. We show in the second
paper of this series
that the Hessian of obstacle avoidance constraints can be kept sparse if we introduce $2N$ additional unknowns in the form of position 
$\ve{r}(u_i) = \left\{x(u_i), y(u_i)\right\}_{i=1}^{N}$ at $N$ points on the mesh
and a constraint 
$\ve{r}(u_i) - \ve{r}(u_{i-1}) = \lambda \left\{ \int_{u_{i-1}}^{u_i} \cth\, d u, \int_{u_{i-1}}^{u_i} \sth\, d u\right\}$ between each pair of $\ve{r}(u_{i-1})$ and $\ve{r}(u_{i})$.
We choose uniformly separated $N$ points and let $N = nM +(n+1)$ so that
obstacle avoidance constraints are imposed at $M$ points in the interior
of each element and at each of the $n+1$ nodes.

With the discretization above, the infinite dimensional problem
of \Sec{sec:full_nonlinear_constrained_optimization_problem} is converted into a finite-
dimensional nonlinear constrained optimization problem.
The objective now is to determine the values of the $4(n+1) + 2N + 1$ 
unknowns that minimize the discomfort cost functional and satisfy the boundary
conditions and constraints described in (\Sec{sec:full_nonlinear_constrained_optimization_problem}).
To impose boundary conditions, we
impose constraints on some of the degrees of freedom and 
eliminate some of the degrees of freedom at the end points.
To impose dynamic constraints, we compute
the speed $v$, tangential acceleration $a_T$, normal acceleration $a_N$,
angular speed $\omega$, and curvature $\kappa$ in terms of the unknowns
at $P$ points in the interior of each element and impose bounds
on these quantities as constraints.  See \Fig{fig:mesh}.

To numerically compute the integrals in the
cost functional and constraints, we use Gauss quadrature formulas
with 12 integration points. In our implementation, the number of elements in the 
finite element mesh, $n = 32$, number of intervals per element for obstacle
avoidance constraints, $M = 20$, the number
of points per element on which to impose dynamic constraints $P = 12$.
The rationale for choosing these particular values is discussed
in the second paper in this series.

\section{Initial guess for the optimization problem}
\label{sec:initial_guess}
Because of the non-linearity in the optimization problem, and the
presence of both inequality and inequality
constraints, it is crucial that a suitable initial guess of the trajectory
be computed and provided to an optimization algorithm. Many software packages 
can generate their own ``starting points'', but a good
initial guess that is within the feasible region can easily reduce the
computational effort (measured by number of function and derivative
evaluation steps) many times.  Not only that, reliably solving a nonlinear
constrained optimization problem without a high quality initial guess can be
extremely difficult.  Because of these reasons, we invest considerable
mathematical and computational effort to generate a high quality initial guess of the
trajectory.  

Our optimization problem is to find the
scalar $\lambda$ and the two functions $\theta$ and $v$ that minimize the discomfort. 
We compute the initial guess of trajectory by computing $\lambda$ and $\theta$ first and
then computing $v$ by solving a separate optimization problem.  We
emphasize that the initial guess computation process must deal with
arbitrary inputs and reliably compute the initial guesses.

\subsection{Initial guess of path}
\label{sec:path_initial_guess}
To compute initial guess for $\theta(u)$ and $\lambda$ for any pair of specified
initial and final orientations $(\theta_0, \theta_\tau)$, we solve an
auxiliary (but simpler) nonlinear constrained optimization problem
for the four pairs of orientations $(\theta_0, \theta_\tau)$,
$(\theta_0, \theta_\tau)$, $(\theta_0, \theta_\tau + 2\pi)$, and 
$(\theta_0, \theta_\tau - 2\pi)$. We minimize
\begin{equation}
\label{eq:th_init_guess_opt}
J(\theta, \lambda) = \lambda + w \int_0^1 \theta''^2 d u
\end{equation}
where $w := \max(\Delta L, R)$, and $\theta$ must satisfy the
boundary conditions, the two equality constraints of
\Eq{eq:r_by_integral_constr}, and the curvature constraint 
$
\left| \theta'(u) \right| \le \lambda \kappa_{\max} \; \forall u \in [0,1].
$
We do not impose obstacle related constraints in this problem.
Here $\Delta L$ is the distance between start and end positions and
$R$ is the minimum allowed radius of curvature. Note that this is a geometric
and time is absent.

This method of computing four different initial guesses is based on a special 
kind of non-uniqueness of paths. 
This particular kind of non-uniqueness
arises because a single physical orientation $\theta$ can correspond to multiple
numerical values of $\theta$ ($\theta \pm 2 n \pi \; \forall n
\in \mathbb{N}$). This is because $\theta$ is continuous and cannot jump to a
different value in between. Of course, this optimization problem needs its own
initial guess, and we use paths similar to Dubins curves
to compute a suitable initial guess. Our method generates two different
initial guesses for the auxiliary problem for the pair $(\theta_0, \theta_\tau)$,
leading to 4 initial
guesses of $\theta(u)$ for the discomfort minimization problem. 
$\lambda$ is the length of the paths computed above.
The non-uniqueness in paths and the computation of initial guess is
discussed in greater detail in the second paper (Part II) of this series.

\subsection{Initial guess of speed}
\label{sec:speed_initial_guess}
For the case when both end-points have non-zero speed, we compute initial
guess of $v(u)$ by solving an auxiliary optimization problem.
We minimize
\begin{equation}
\label{eq:v_init_guess}
J(v) = \int_0^1 v''^2 d u
\end{equation}
subject to boundary constraints $v(0) = v_0 > 0$, $v(1) = v_1 > 0$,
$v'(0) = \frac{a_0 \lambda}{v_0}$, $v'(1) = \frac{a_1 \lambda}{v_1}$
and inequality constraints $v_{\min}(u) \le v(u) \le v_{\max}(u)$
and $A_{\min}(u) \le v'(u) \le A_{\max}(u)$.
The expressions for $v'(0)$ and $v'(1)$ come from the relation
in~\Eq{eq:aT}. The length $\lambda$ is computed when the initial
guess for $\theta$ is computed.  Here we choose $v_{\min}(u) = \min(v_0, v_1)/2$
and $v_{\max}(u)$ is a constant that comes from
the hardware limits.  The function $A_{\min}(u)$ is chosen to
be the constant $10 a_{\min} \lambda / \min(v_0, v_1)$ where
$a_{\min}$ is the minimum allowed physical acceleration.
$A_{\max}(u)$ is chosen similarly using $a_{\max}$.

If both end-points have zero speeds, the function
\begin{equation}
\label{eq:zero_v_init_guess_function}
v(u) = v_{\max} \left( 4 u (1 - u) \right)^{2/3}
\end{equation}
satisfies the boundary conditions and singularities and has a maximum
value of $v_{\max}$.  This case does not require any optimization.

If only one of the end-points has a zero speed boundary condition,
we split the initial guess for $v$ into a sum of two functions.
The first one takes care of the singularity and the second takes
care of the non-zero speed boundary condition on the other end-point.
We now maintain only the $v_{\max}$ constraint because
$v'(u)$ is unbounded and $v_{\min} = 0$ naturally.
If the right end-point has zero speed, we choose
\begin{equation}
\label{eq:non_zero_v_init_guess_function}
v(u) = v_{\mbox{singular}}(u) + v_{\mbox{non-singular}}(u)
\end{equation}
where
\begin{equation}
v_{\mbox{singular}}(u) = \frac{16}{9} 2^{1/3} v_{\max} u^2 (1-u)^{2/3}.
\end{equation}
This function has the correct singularity behavior and its maximum
value is $v_{\max}/2$.  The non-singular part is computed via optimization
so that the sum is always less than $v_{\max}$.  For the other case,
when left end-point has zero speed, the singular part (using symmetry) is
$$
\frac{16}{9} 2^{1/3} v_{\max} (1-u)^2 u^{2/3}.
$$

\subsection{Summary}
For any given pair of orientations, we compute four initial guesses of path
$\theta(u)$ and corresponding path lengths $\lambda$. To compute
initial guess of speed $v(u)$, we treat zero speed and non-zero speed boundary
conditions differently. When speed is zero at both ends, we use
\Eq{eq:zero_v_init_guess_function}. When speed is non-zero at either ends, we
use \Eq{eq:non_zero_v_init_guess_function}. In this case, we compute
$v_{\mbox{non-singular}}(u)$ by solving the optimization problem
\Eq{eq:v_init_guess}. In this case, four initial guesses
of speed are computed corresponding to each guess of $\lambda$.

\section{Evaluation and results}
\label{sec:evaluation_and_results}
The motion planning framework described in this paper is expected
to reliably plan trajectories for different types of boundary conditions.
These trajectories should satisfy dynamic constraints and the corresponding 
geometric paths should not intersect obstacles. Further, this framework should 
reliably compute trajectories between a given pair of boundary
conditions for a range of weights, $w_{\tny T}$ and $w_{\tny N}$,
so that users can customize the motion by changing these weights.
In the following discussion, we refer to optimization problem
of \Sec{sec:full_nonlinear_constrained_optimization_problem}
as the {\em discomfort minimization problem}.

We begin by describing the input to the discomfort minimization 
problem. Some quantities in the input 
such as dynamic bounds
are fixed, while others such as boundary conditions and obstacle
locations and shapes are problem dependent. We also provide some
implementation details.

Next, we present illustrative examples showing the various steps 
of the solution method, and demonstrate some of the strengths of our method
such as the ability to plan trajectories for a wide variety of boundary 
conditions and obstacle shapes.

We then analyze how varying the weight factors $f_{\tny T}$ and $f_{\tny N}$
affect the solution trajectory. Our objective is to find qualitative 
relationships between these weight factors and each of the terms in the
discomfort measure (total travel time, integral of
squared tangential jerk, and integral of squared normal jerk). 
These relationships should provide guidelines for user customization.

Next, to evaluate the reliability of our method, we construct a 
set of problems by varying boundary conditions 
and find the success rate. We also analyze the run-time and number of 
iterations to solve the discomfort minimization problem.

\subsection{Experimental setting}
\label{sec: experimental_setting}
The input to the discomfort minimization problem described in 
\Sec{sec:full_nonlinear_constrained_optimization_problem} consists of:
\begin{enumerate}

\item Number of elements, $n$, for finite element discretization.
We choose $n = 32$ based on a numerical experiment based on convergence
to the ``exact'' solution of the infinite dimensional optimization
problem as the maximum finite element size is reduced. Details are in the second 
paper (Part II) of this series.

\item Number of intervals per element $M$, to compute the $\left\{x, y\right\}$ 
pairs for imposing obstacle constraints (see \Sec{sec:obstacle_avoidance} and \Sec{sec:numerical_solution}).
We choose $M = 20$ when obstacles are present, otherwise the choice is
irrelevant.

\item Values of bounds on curvature, speed, angular speed, tangential acceleration,
and normal acceleration (See \Sec{sec:full_nonlinear_constrained_optimization_problem}). Curvature bound
should be determined from the robot's geometry. We choose the value for a typical assistive wheelchair.
While we assume a point robot
and do not consider robot shape for obstacle-avoidance, we do include curvature
constraints based on the dimensions of a typical wheelchair. All other bounds should be
chosen for comfort. In the absence of relevant comfort studies for assistive robots,
we choose bounds on linear and angular speed based on our experience with
an intelligent wheelchair~\citep{Gulati_2008,Gulati_2009,Murarka_2009b} and
studies of comfort in ground vehicles (see \Sec{sec:comfort}). 
All these values are shown in Table~\ref{tab:bounds}.

\begin{table}[h]
\begin{center}
  \begin{tabular}{ | l | r | r | }
    \hline
    {\bf Quantity} 			& {\bf Lower Bound} & {\bf Upper Bound} \\ \hline
    Curvature (1/m) &  $ -1.8$ 					& 1.8					\\ \hline
    Speed (m/s)& 0.0 & 3.0 \\ \hline
    Angular speed (rad/s) & -1.57 & 1.57 \\ \hline
    Tangential acceleration (m/s$^2$) & -1.0 & 1.0 \\ \hline
    Normal acceleration (m/s$^2$) & -1.0 & 1.0 \\ \hline
  \end{tabular}
\end{center}
\caption{Lower and upper bounds on curvature, speeds, and accelerations used in experiments.} {\small{Curvature bounds are based on a minimum turning radius of 0.55 m.}}
\label{tab:bounds}
\end{table}

\item Non-dimensional multiplying factors for weights, $f_{\tny T} > 0$ and 
$f_{\tny N} > 0$. Both these values are set to $1$ unless mentioned otherwise.

\item Representation of obstacles as star-shaped domains with piecewise $C^2$ boundary (see \Sec{sec:obstacle_avoidance_constraint_derivatives}). In our experiments, we use circular, elliptical,
and star-shaped polygonal obstacles. See Figures~\ref{fig:corridor_example} and~\ref{fig:star_shaped_example}.

\item Boundary conditions on position, orientation, curvature, speed and 
tangential acceleration (see \Sec{sec:analysis_of_boundary_conditions}). These are problem specific and
we describe these for each of the experiments.

\end{enumerate}

We have implemented our code in C++. We use Ipopt, a robust large-scale 
nonlinear constrained
optimization library~\citep{Wachter_2006}, also written in C++, to solve the
optimization problem. We explicitly compute gradient and Hessian for the optimization
problem in our code instead of letting Ipopt compute these using finite difference.
This leads to greater robustness and faster convergence.  We set
the Ipopt parameter for relative tolerance as $10^{-8}$ and set the maximum number
of iterations to $500$.

After optimization, the outputs
are the nodal values of $v$, $v'$, $\theta$, and $\theta'$, and the curve
length $\lambda$ (see \Sec{sec:numerical_solution}). The functions $v(u)$ and $\theta(u)$, $u \in [0,1]$ are 
known in terms of these nodal values. We use 
\Eq{eq:t} to construct a table of $u$ values for
$u \in [0,1]$ and the 
corresponding $t$ values for $t \in [0,\tau]$. The value of any of the quantities 
of interest (orientation, speed, etc.) at any time $t \in [0, \tau]$ is computed
using this table by linear interpolation.

\subsection{Illustrative examples}
\label{sec:illustrative_examples}
\begin{figure}[p]

\begin{center}
	\includegraphics[scale=0.5]{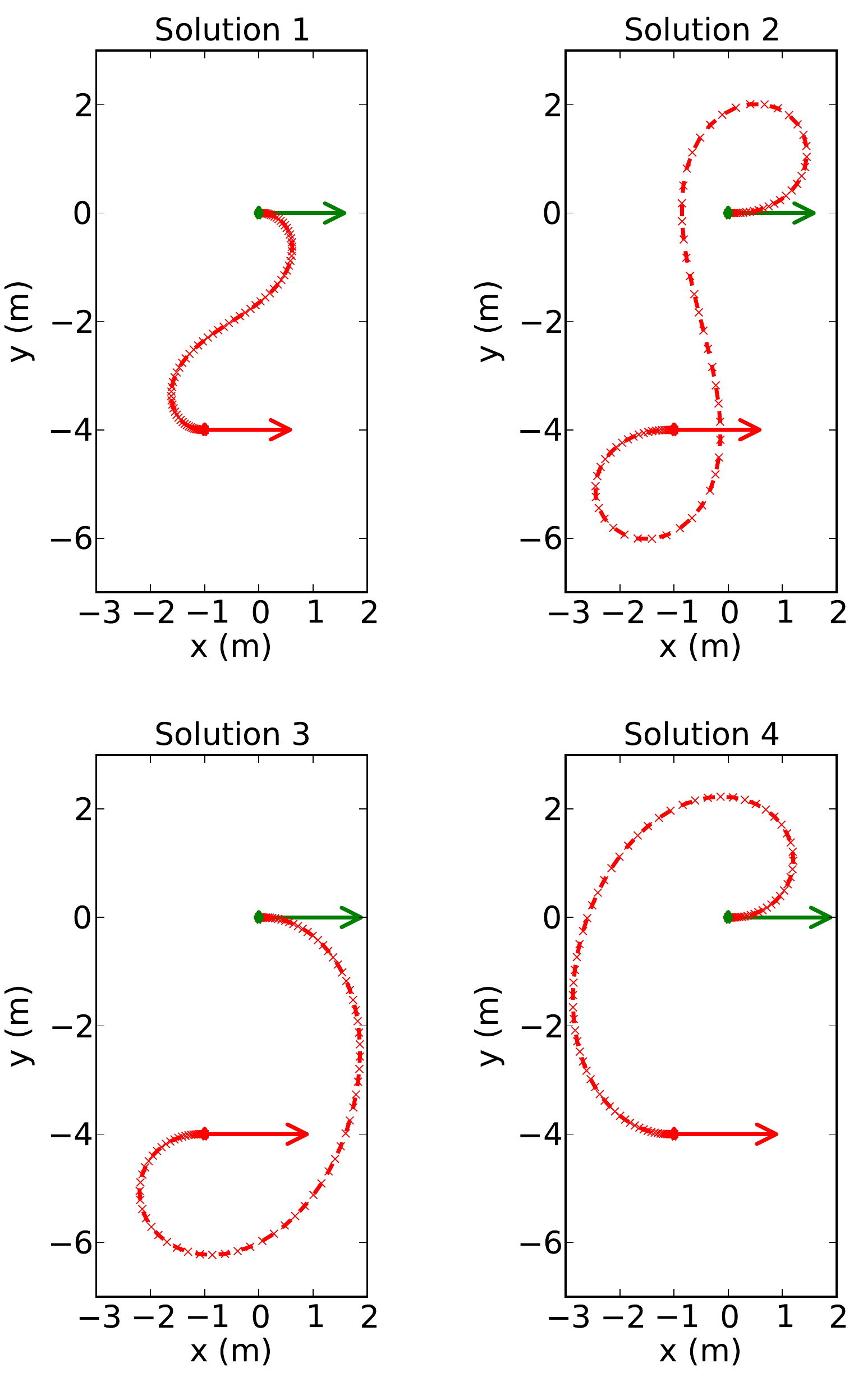}
\end{center}
\caption{Four initial guess of path.}{\small{Problem input is as follows: initial position = $\left\{0,0\right\}$, orientation = 0,
speed = 0, tangential acceleration = 0; final position = $\left\{-1,-4\right\}$, orientation = 0,
speed = 0, tangential acceleration = 0. The four initial guesses of path are computed
using the method described in \Sec{sec:path_initial_guess} so that final orientation 
in (a),(b),(c) and (d)
is 0, 0, $-2\pi$ and $2\pi$ respectively. All quantities have appropriate units in terms 
of meters and seconds. Initial and final positions are shown by markers and
orientations are indicated by arrows. While the path is parameterized by $u$, for
ease of visualization, we show markers at equal intervals of time. Thus distance between
markers is inversely proportional to speed. }}
\label{fig:path_init-s_shape}

\begin{center}
	\includegraphics[scale=0.45]{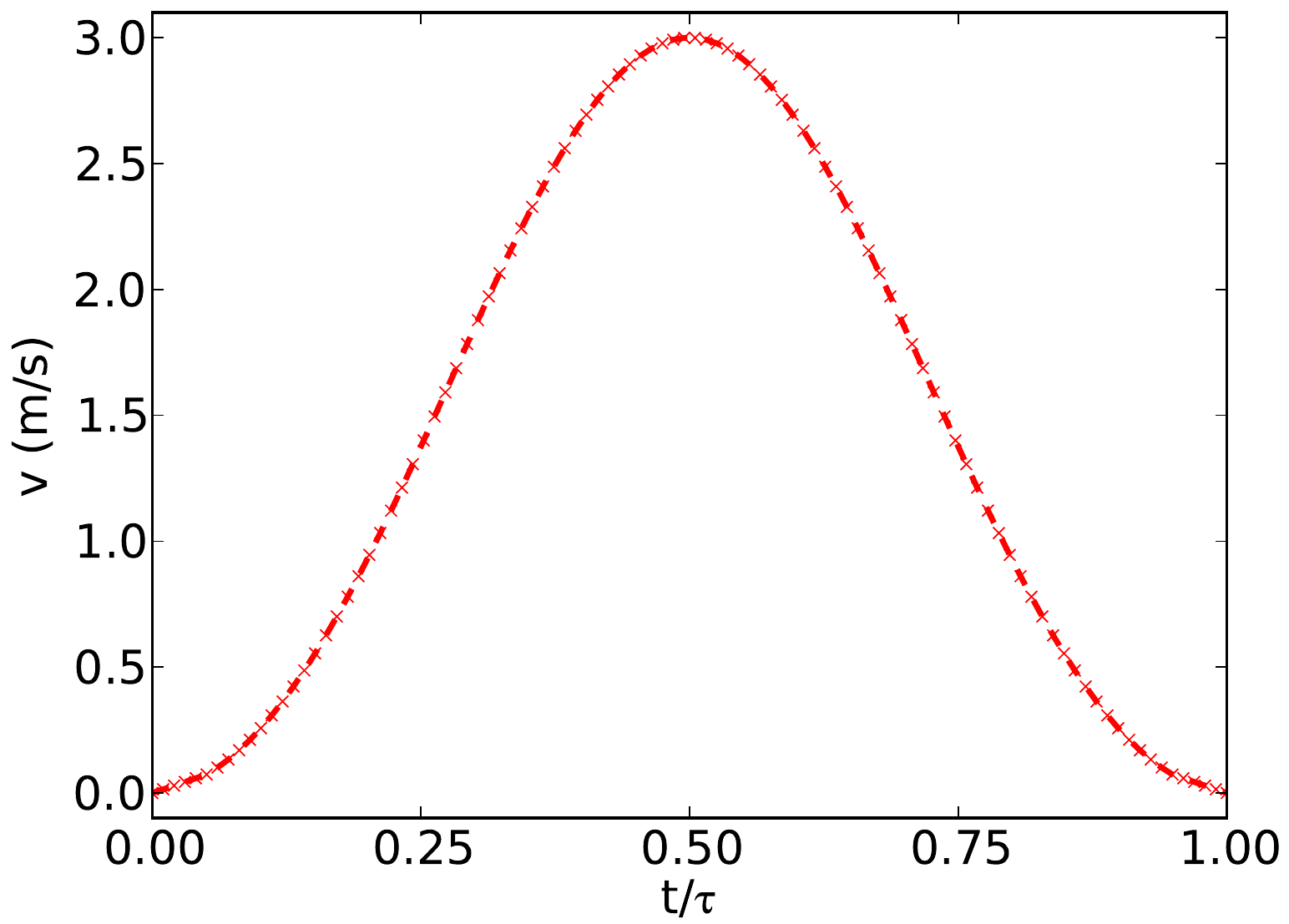}
\end{center}
\caption{Initial guess of speed for problem of \Fig{fig:path_init-s_shape}.}{\small{In this case, because of zero speed boundary condition
on both ends, the same initial guess of speed is produced for each path guess. When
speed is non-zero on one or both ends, four distinct guesses of speed may be produced.}}
\label{fig:v_init-s_shape}
\end{figure}

We begin by presenting an example that illustrates the optimization process.
In \Fig{fig:path_init-s_shape}, the initial position is $\left\{0, 0\right\}$
and final position is $\left\{-1, -4\right\}$. The initial and final orientations are
both zero. The speed and tangential acceleration at both ends are also zero.

First,
four initial guesses of path ($\theta(u), u \in [0,1]$ and $\lambda$) are computed
as described in \Sec{sec:path_initial_guess}. These four initial guesses are shown in
\Fig{fig:path_init-s_shape}.
The first two guessed have $\theta_\tau = 0$, the third guess
has $\theta_\tau = -2\pi$, and the fourth guess has $\theta_\tau = 2\pi$.
An initial guess of speed, $v(u)$, is computed as described in
\Sec{sec:speed_initial_guess}. In this example, speed at both
ends is zero and hence $v(u)$ is computed using \Eq{eq:zero_v_init_guess_function}. 
Thus we
get the same function $v(u)$ for all guesses of path. See \Fig{fig:v_init-s_shape}.

\begin{figure}
\begin{center}
	\includegraphics[scale=0.62]{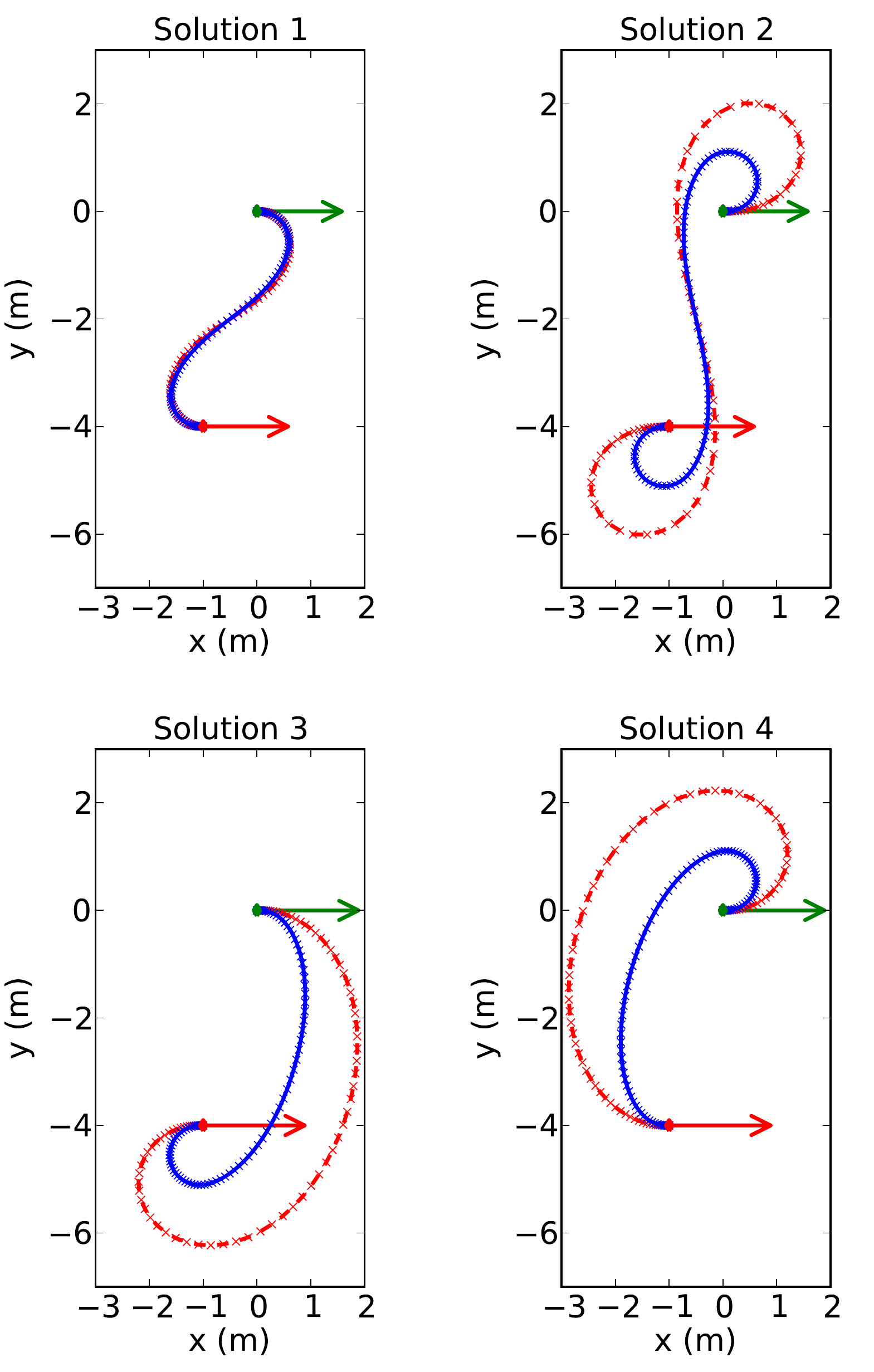}
\end{center}
\caption{Solution paths of the problem of \Fig{fig:path_init-s_shape}.}{\small{Final (optimal) path for each solution 
is shown as solid curve. Initial guess is shown as dashed curve.
The number of DOFs for the discomfort minimization problem were $1403$ and number of constraints were $3232$.
The total cost and travel time for the four solutions are shown in Table~\ref{tab:s_shape_costs}.}
\label{fig:path_init_and_final-s_shape}}
\end{figure}

The discomfort minimization problem is solved for
each of these four initial guesses. 
The four solution paths that minimize discomfort are shown in 
\Fig{fig:path_init_and_final-s_shape}. The travel time and
costs for the four solution paths are shown in Table~\ref{tab:s_shape_costs}.
The path corresponding to Solution 1 has the minimum cost, and
is thus in agreement with our intuitive notion of the best path among these
four. Notice the circular arcs at the start and end of the path of Solution 2. These 
arcs have a constant radius equal to the minimum
turning radius of the robot because of curvature constraints. If curvature 
constraints are not imposed, these arcs have a smaller radius and the path has 
a smaller length. Note that it is not always true that all four solutions 
are distinct since two or more problems starting from different initial 
guesses may converge to the same solution.

\begin{figure}
\begin{center}
	\includegraphics[scale=0.45]{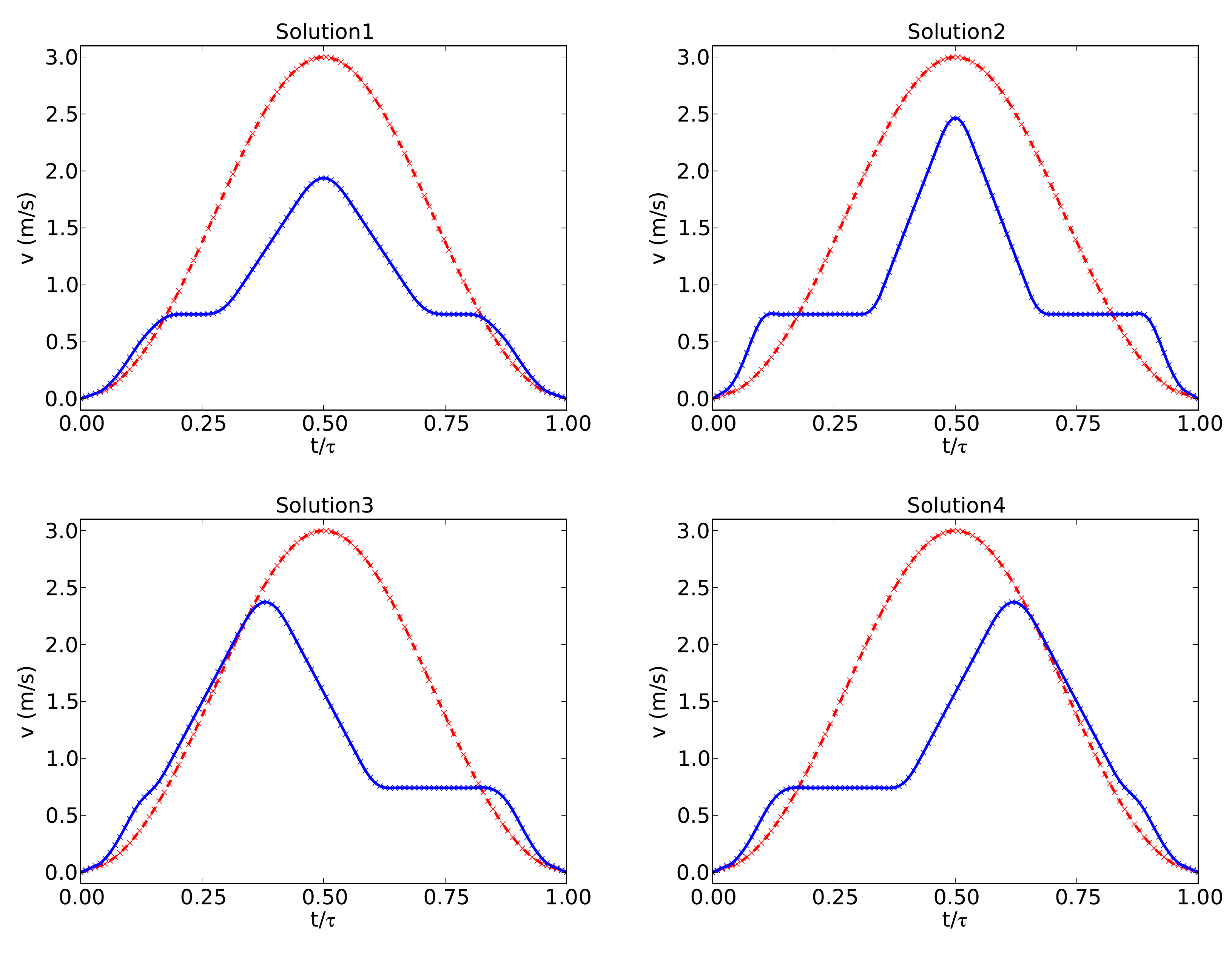}
\end{center}
\caption{Solution speeds of the problem of \Fig{fig:path_init-s_shape}.}{\small{Final (optimal) speed for each solution 
is shown as the solid curve. Initial guess is shown as the dashed curve.}}
\label{fig:v_init_and_final-s_shape}
\end{figure}

The solution speeds are shown in \Fig{fig:v_init_and_final-s_shape}. 
The final speeds in Solution 1 and Solution 2 are symmetric about 
$t = \frac{\tau}{2}$ because of the inherent ``symmetry'' due
to zero orientation, speed, and acceleration at both ends. The final speeds 
in Solution 3 and Solution 4 are mirror images of each other about
$t = \frac{\tau}{2}$ because the final orientations in these two are 
$-2\pi$ and $2\pi$ respectively. The figures also show that the initial guesses
of the paths and speeds are quite good, which is important for nonlinear optimization.

\begin{figure}
\begin{center}
		\includegraphics[scale=0.6]{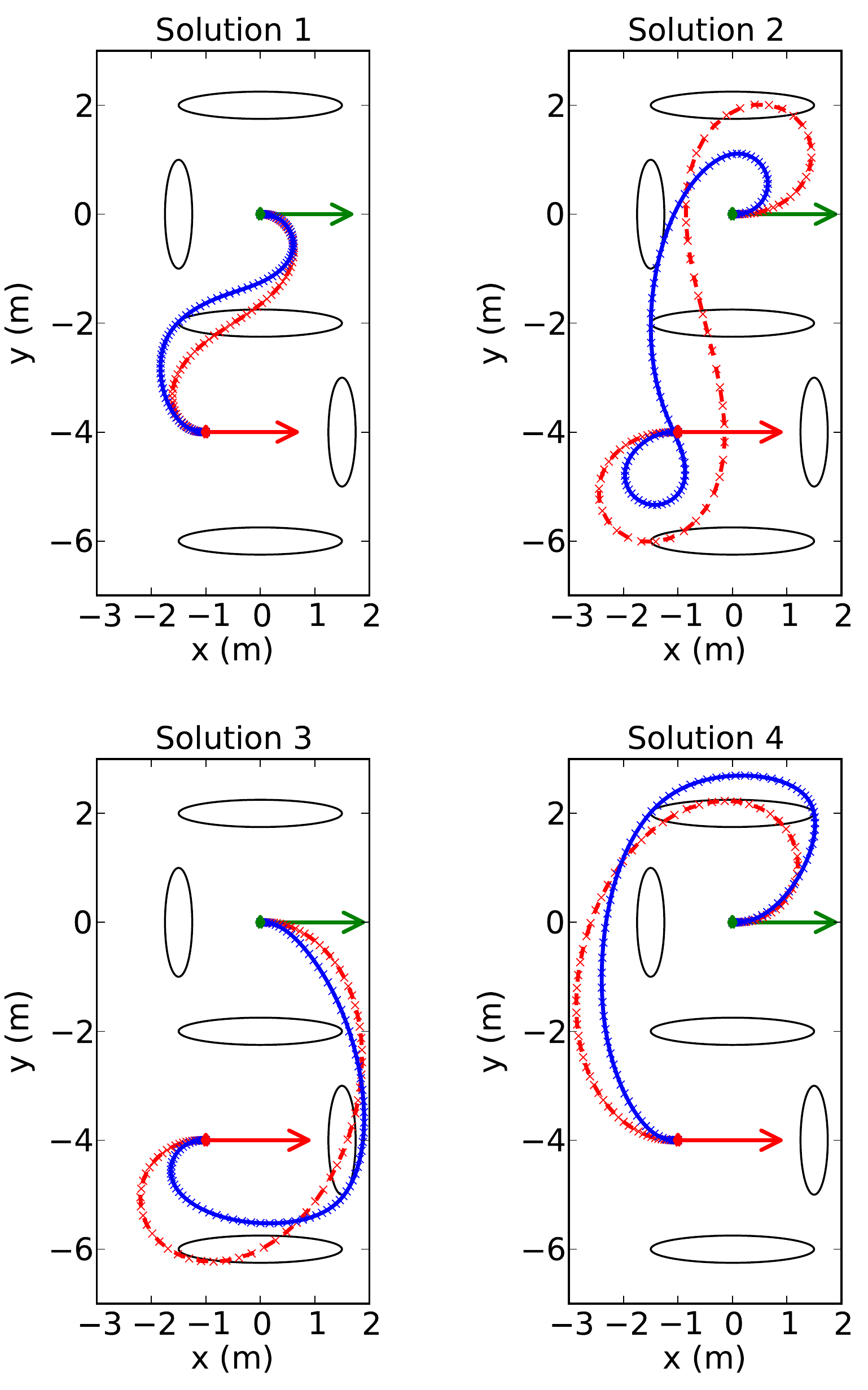}
	\end{center}
\caption{Solution paths to a problem with five elliptical obstacles.} {\small{The boundary conditions 
of this problem are identical to the problem of \Fig{fig:path_init-s_shape}. 
Four distinct solution
paths in the neighborhood of the four initial guesses are found. Initial guesses are the
dotted curves while the final solutions are the solid curves. This problem had
$3195$ constraints for obstacle-avoidance in addition to the constraints in \Fig{fig:path_init-s_shape}.
The total cost and travel time for the four
solutions are shown in Table~\ref{tab:s_shape_obs_costs}.}}
\label{fig:path_init_and_final-s_shape_obs}
\end{figure}

In \Fig{fig:path_init_and_final-s_shape_obs}, we introduce five 
elliptical obstacles for the same boundary conditions.
All four initial guesses of path and solution paths are shown. The 
initial guesses of path and speed do not consider obstacles and hence are identical to 
those in Figures~\ref{fig:path_init_and_final-s_shape} and~\ref{fig:v_init_and_final-s_shape}
respectively.
Four distinct solution paths are found.  The travel 
time and total cost for all four solutions is shown in Table~\ref{tab:s_shape_obs_costs},
and is greater than for the problem of \Fig{fig:path_init_and_final-s_shape} (see Table~\ref{tab:s_shape_costs}). The minimum cost path is that of Solution 1 which again agrees with our intuition. Notice
how the path of Solution 3 passes above the lowermost elliptical obstacle, while
the path of Solution 4 passes below the uppermost elliptical obstacle. 
Our experience
with this and other examples shows that once the optimization algorithm 
takes a step that brings an iterate to one side of the obstacle, further
iterations keep it on the same side.  We believe that this is because
paths passing an obstacle on different ``sides'' cannot be transformed to each other via
a continuous deformation of the path and lie in disjoint
feasible regions.  The iterates in the optimization process 
cannot jump from one feasible region to a different feasible region
in general.

\clearpage

\clearpage

\clearpage

\begin{table}
\begin{center}
  \begin{tabular}{ | c | r | r | }
    \hline
    {\bf Solution Number} 			& {\bf Travel time (s)} & {\bf Total cost (s)} \\ \hline
    1 &  $ 6.3$ 					& 6.5					\\ \hline
    2 & 10.0 & 11.0 \\ \hline
    3 & 7.9 & 8.0 \\ \hline
    4 & 7.9 & 8.0 \\ \hline
  \end{tabular}
\end{center}
\caption{Travel time and total cost for problem of \Fig{fig:path_init-s_shape}.}
\label{tab:s_shape_costs}
\end{table}

\clearpage

\clearpage
\begin{table}[t!]
\begin{center}
  \begin{tabular}{ | c | r | r | }
    \hline
    {\bf Solution Number} 			& {\bf Travel time (s)} & {\bf Total cost (s)} \\ \hline
    1 &  $ 7.0$ 					& 7.1					\\ \hline
    2 & 11.7 & 11.9 \\ \hline
    3 & 9.1 & 9.4 \\ \hline
    4 & 10.3 & 10.4 \\ \hline
  \end{tabular}
\end{center}
\caption{Travel time and total cost for problem of \Fig{fig:path_init_and_final-s_shape_obs}.}
\label{tab:s_shape_obs_costs}
\end{table}

\begin{figure}
\begin{center}
	\subfigure[]
	{
	    \includegraphics[scale=0.195]{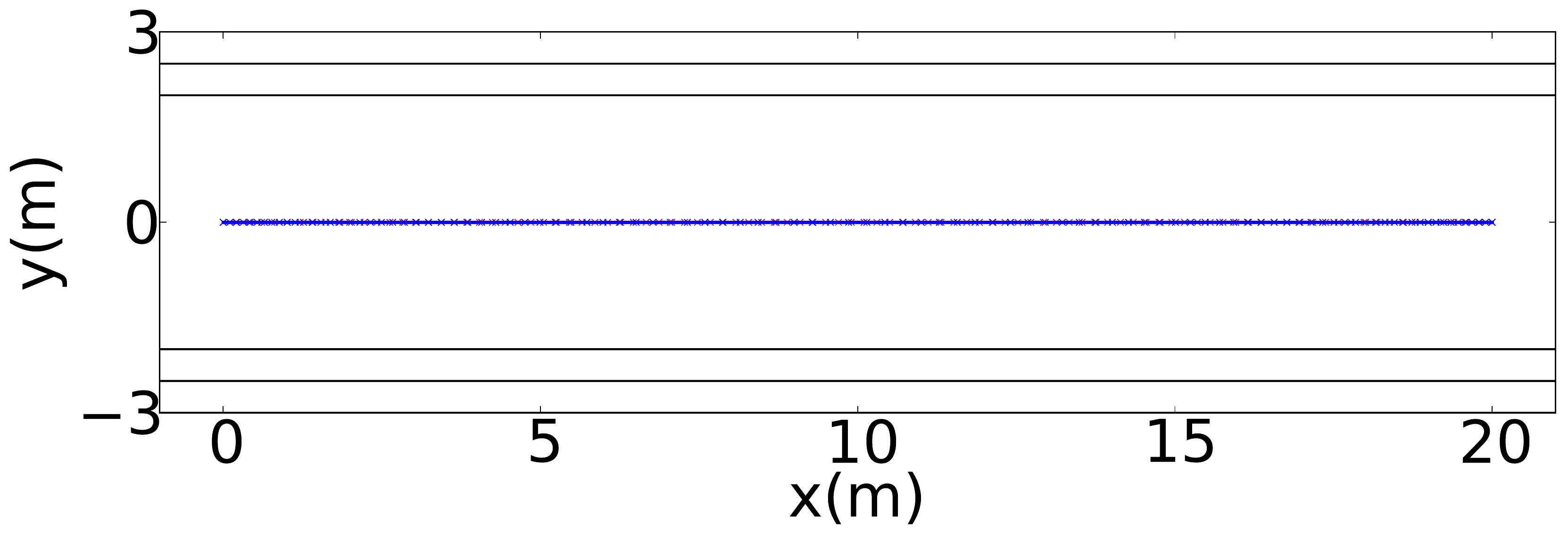}
	}
	\subfigure[]
	{   
	    \includegraphics[scale=0.195]{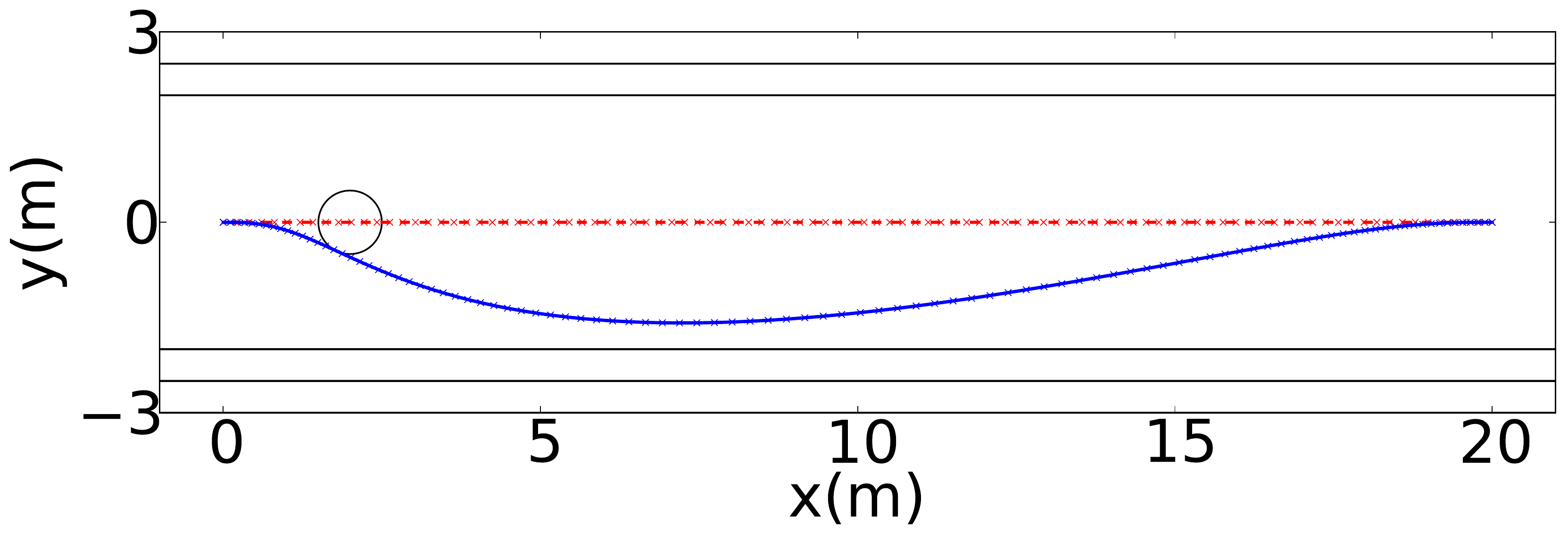}
	}\\
	\subfigure[]
	{
	    \includegraphics[scale=0.195]{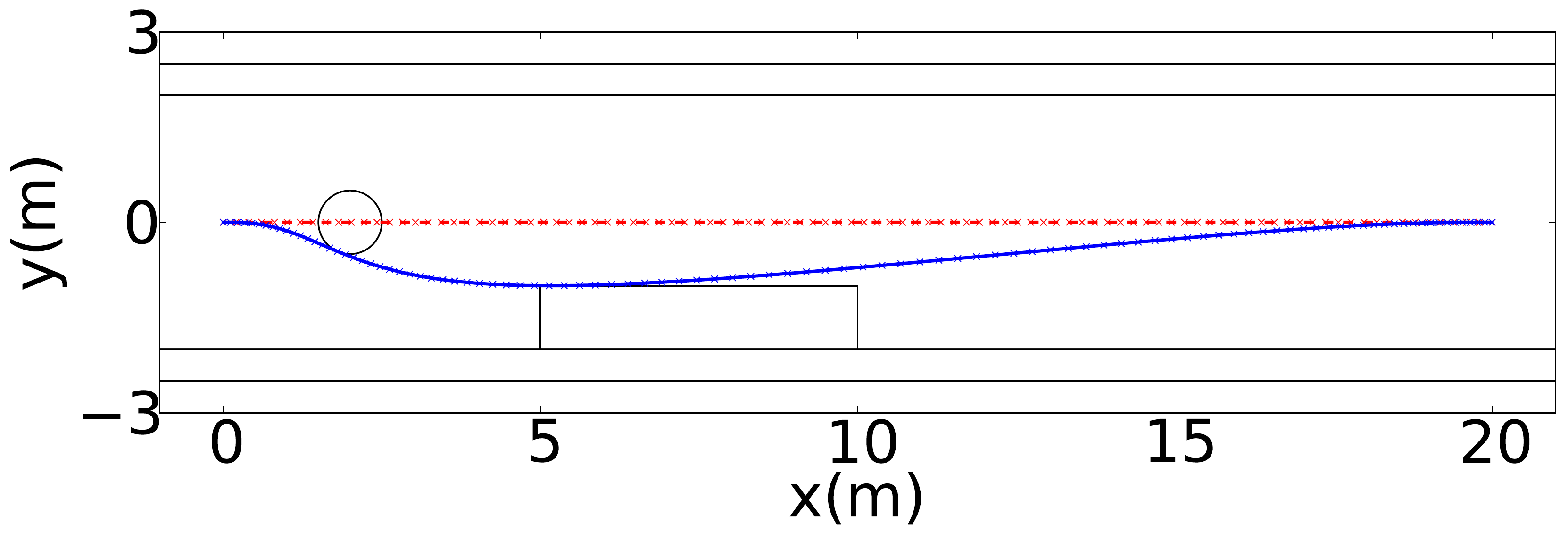}
	}
	\subfigure[]
	{   
	    \includegraphics[scale=0.195]{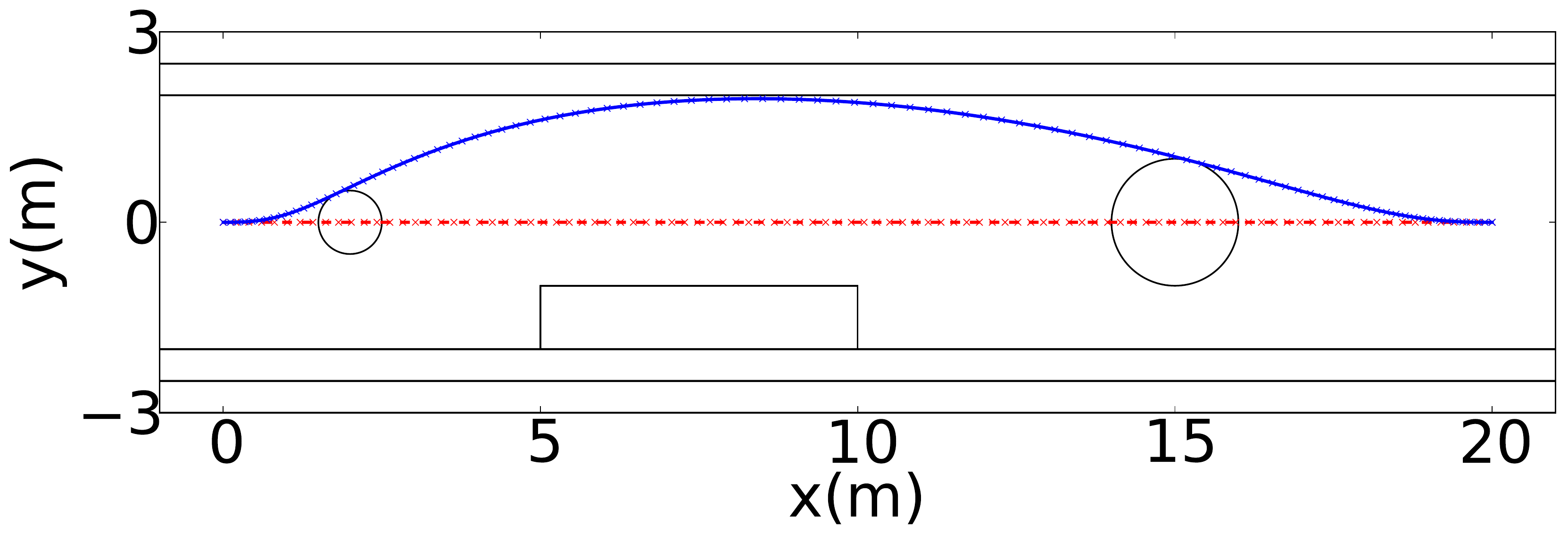}
	}
	\subfigure[]
	{
	    \includegraphics[scale=0.195]{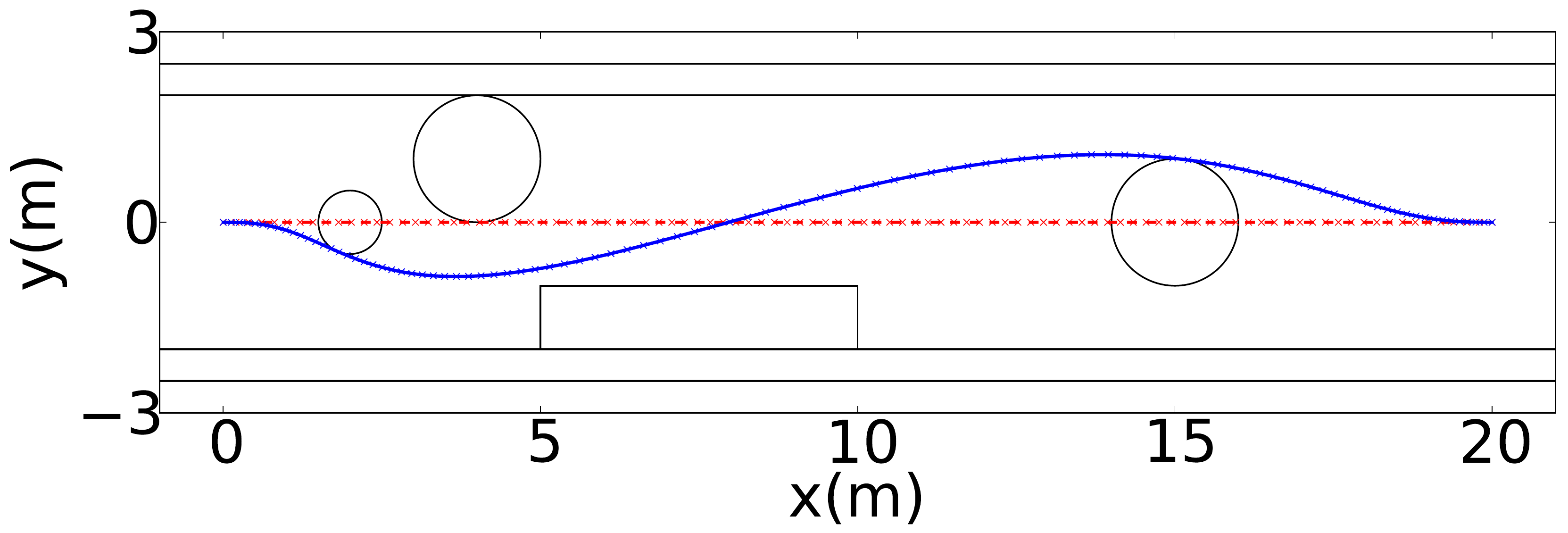}
	}
	\subfigure[]
	{   
	    \includegraphics[scale=0.195]{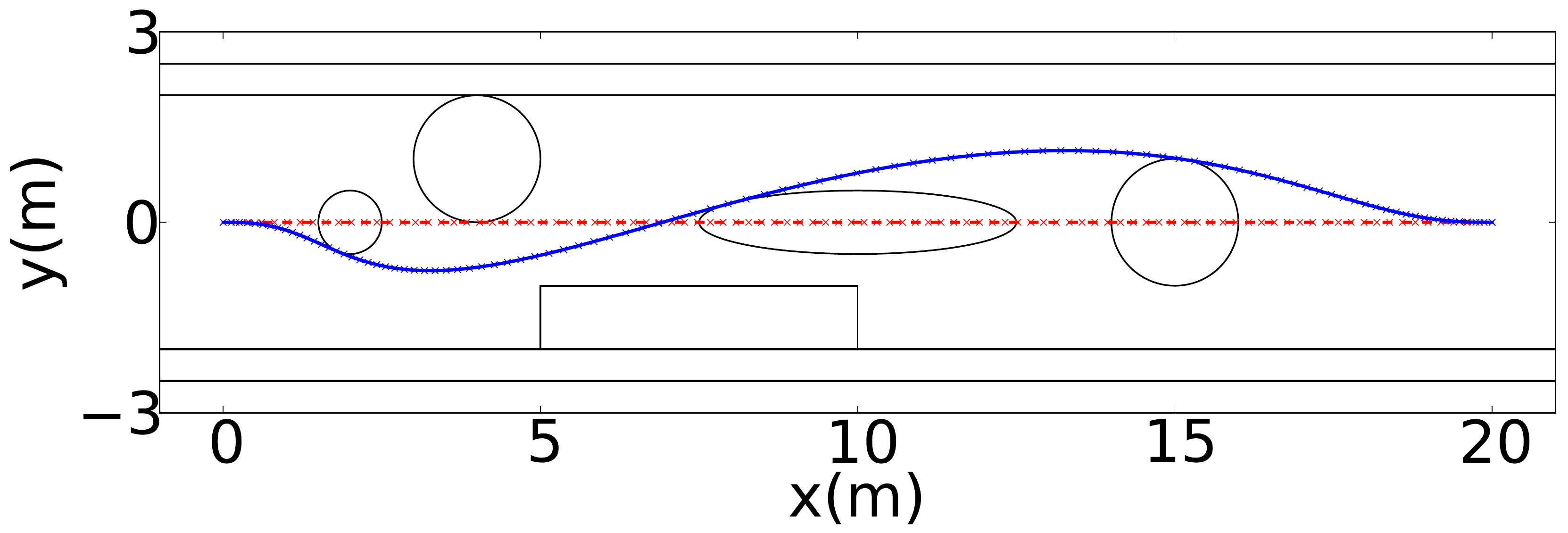}
	}
\end{center}
\caption{Obstacle avoidance in a corridor-like setting with
non-zero speed at both ends.}
{\small{Problem input is as follows: initial position = $\left\{0,0\right\}$, orientation = 0,
speed = 1, tangential acceleration = 0; final position = $\left\{20,0\right\}$, orientation = 0,
speed = 1, tangential acceleration = 0. One of the four solution paths is shown.
Initial guess is shown as the dashed curve while solution is shown as the solid
curve. (a) Only two rectangular obstacles, comprising the corridor
walls are present. The solution path is a straight line. (b) Addition of
a circular obstacle results in a path that passes below the obstacle. Another
solution path that passes above the obstacle and is symmetric to this path
about the center line would also be a solution with same cost. (c),(d),(e),(f)
One more obstacle is added and the same problem is solved starting from the
same initial guess as in (a). All quantities have appropriate units in terms 
of meters and seconds.}}
\label{fig:corridor_example} 
\end{figure}

\Fig{fig:corridor_example} show an example where the initial and final
speeds are both non-zero. This scenario exemplifies one of the common navigation
tasks for an autonomous mobile robot -- that of navigating in a corridor or sidewalk
or driving in a lane on a road.
We show only one solution out of four in this case.
\Fig{fig:corridor_example}(a) has two rectangular obstacles, signifying a wall. 
In the sequence \Fig{fig:corridor_example}(b)--(f), one obstacle is
added at a time, and each time a path is found that avoids all the obstacles.

\begin{figure}
\begin{center}
	\includegraphics[scale=0.4]{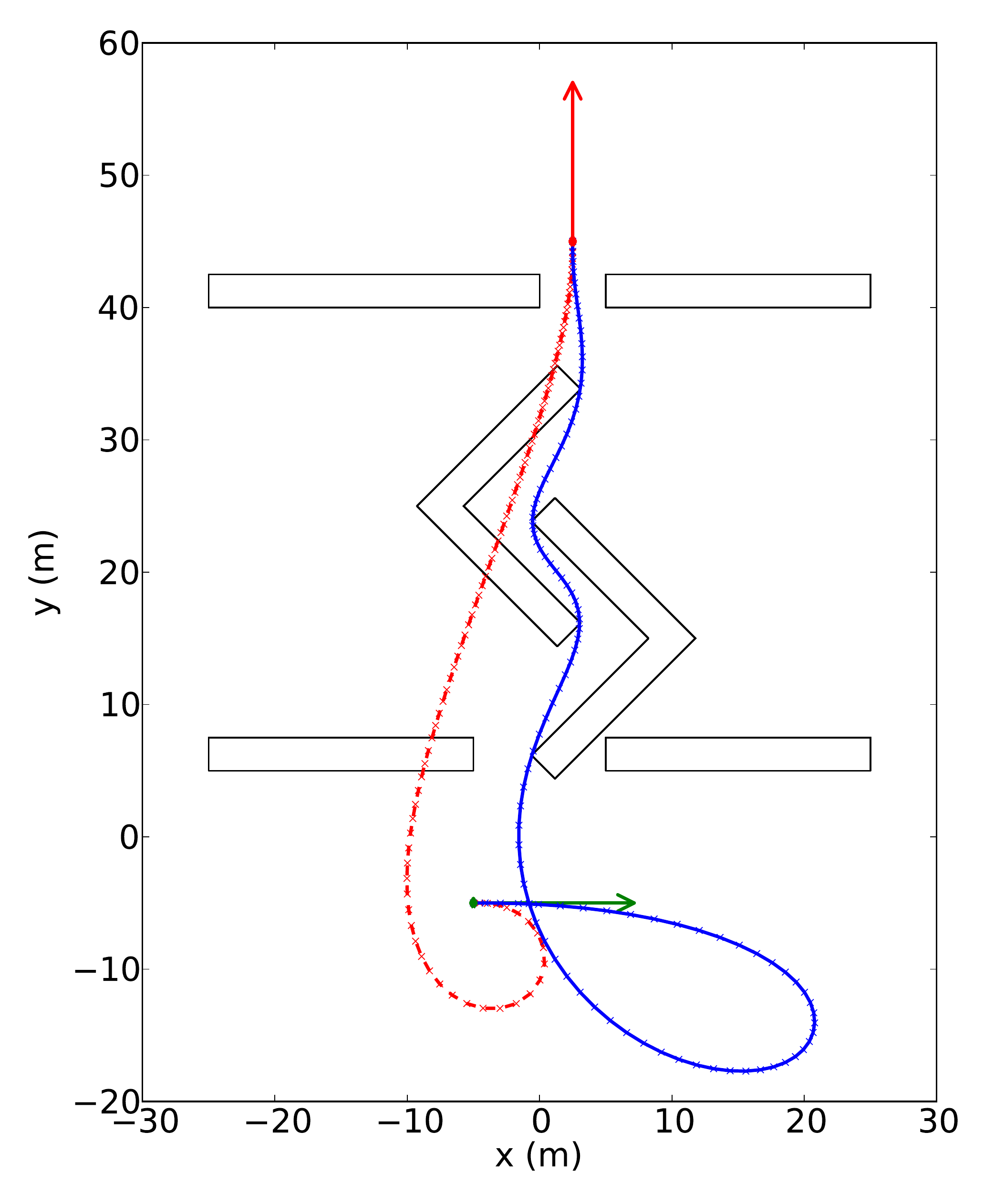}
\end{center}
\caption{Illustrative example showing passage through narrow space between star-shaped obstacles with
non-zero speed at both ends and high positive acceleration at start.}
{\small{Problem input is as follows: initial position = $\left\{-5,-5\right\}$, orientation = 0,
speed = 1, tangential acceleration = 0.5; final position = $\left\{2.5,45\right\}$, orientation = $\pi/2$,
speed = 1, tangential acceleration = 0. Four rectangular and two star-shaped obstacles are present.
Initial guess is shown as the dashed curve while solution is shown as the solid
curve. The loop in the solution at the beginning of the path is because the initial acceleration is high and hence it is not
possible to make a sharp turn without violating dynamic constraints. All quantities have
appropriate units in terms of meters and seconds.}}
\label{fig:star_shaped_example}
\end{figure}

\Fig{fig:star_shaped_example} shows an example when the initial speed
is non-zero and the initial acceleration is positive. There are four rectangular 
and two star-shaped obstacles. This is a particularly difficult
case because it involves a non-zero speed and high acceleration(0.5 m/s$^2$, half the
maximum allowable acceleration) at the beginning and a narrow
passage between obstacles. In this case, only one of the four initial
guesses resulted in a solution. Notice the
loop in the path near the
start. This is because the initial speed and acceleration are non-zero, 
and hence a sharp 90 degree left turn is not possible without violating
dynamic bounds. If dynamic bounds are removed, another path, without
a loop, starting
from another initial guess is also found as a solution. This path
does not have a loop. Also notice how the path just touches the vertices 
of obstacles so that its length is as small as is consistent with comfort.

\clearpage

\clearpage

\subsection{Effect of weights on discomfort}
\label{sec:effect_of_weights}
In this section we analyze the effect of the two dimensionless
factors $f_{\tny T}$ and $f_{\tny N}$ on the individual terms comprising the cost functional
(travel time, integral of squared tangential jerk, and 
integral of squared normal jerk) (see \Eq{eq:cost_functional}). This analysis
provides us with guidelines for choosing the values of weights for customization by human
users.
Henceforth, for conciseness, we will refer to the three terms -- travel time, 
integral of squared tangential jerk, and 
integral of squared normal jerk as $\tau$, $J_{\tny T}$ and $J_{\tny N}$ respectively.
Thus, the cost functional of \Eq{eq:cost_functional} is 
$$J = \tau + f_{\tny T} J_{\tny T} + f_{\tny N} J_{\tny N}$$

For this experiment, we construct a problem with identical boundary conditions as that of 
the example in \Fig{fig:path_init-s_shape}. In order to 
delineate the effect of weights, we remove all constraints and solve the unconstrained problem
for a range of factors $f_{\tny T}$ and $f_{\tny N}$ for each of
the four initial guesses. $f_{\tny T}$ is varied from
$2^{-13}$ to $2^{13}$ in a geometric sequence, each subsequent value being obtained by
multiplying the current value by 10. For each value of $f_{\tny T}$, $f_{\tny N}$ is 
varied from $2^{-13}$ to $2^{13}$ in a similar manner. Thus each weight roughly ranges between
0.0001 and 10000. This results in $4 \times 27 \times 27 = 2916$ 
problems out of which $97\%$ were successfully solved. We show plots corresponding 
to only one of these four solutions. Plots for the
remaining solutions are similar, although the number of problems that converge
is different for each initial guess.

\begin{figure}
\begin{center}
	\subfigure[]
	{
		\includegraphics[scale=0.43]{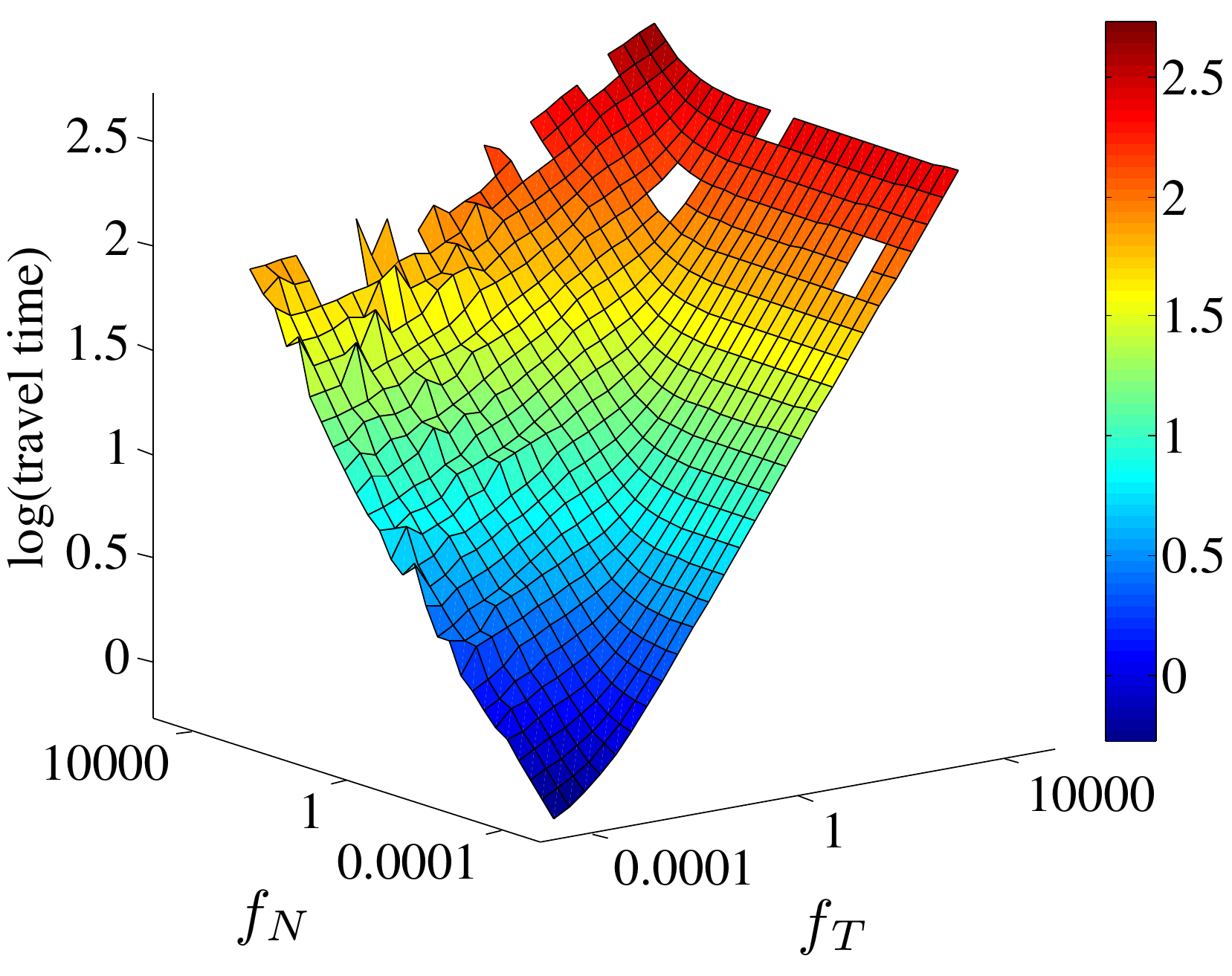}
	}
	\subfigure[]
	{
		\includegraphics[trim = 0mm -20mm 0mm 0mm, clip,scale=0.34]{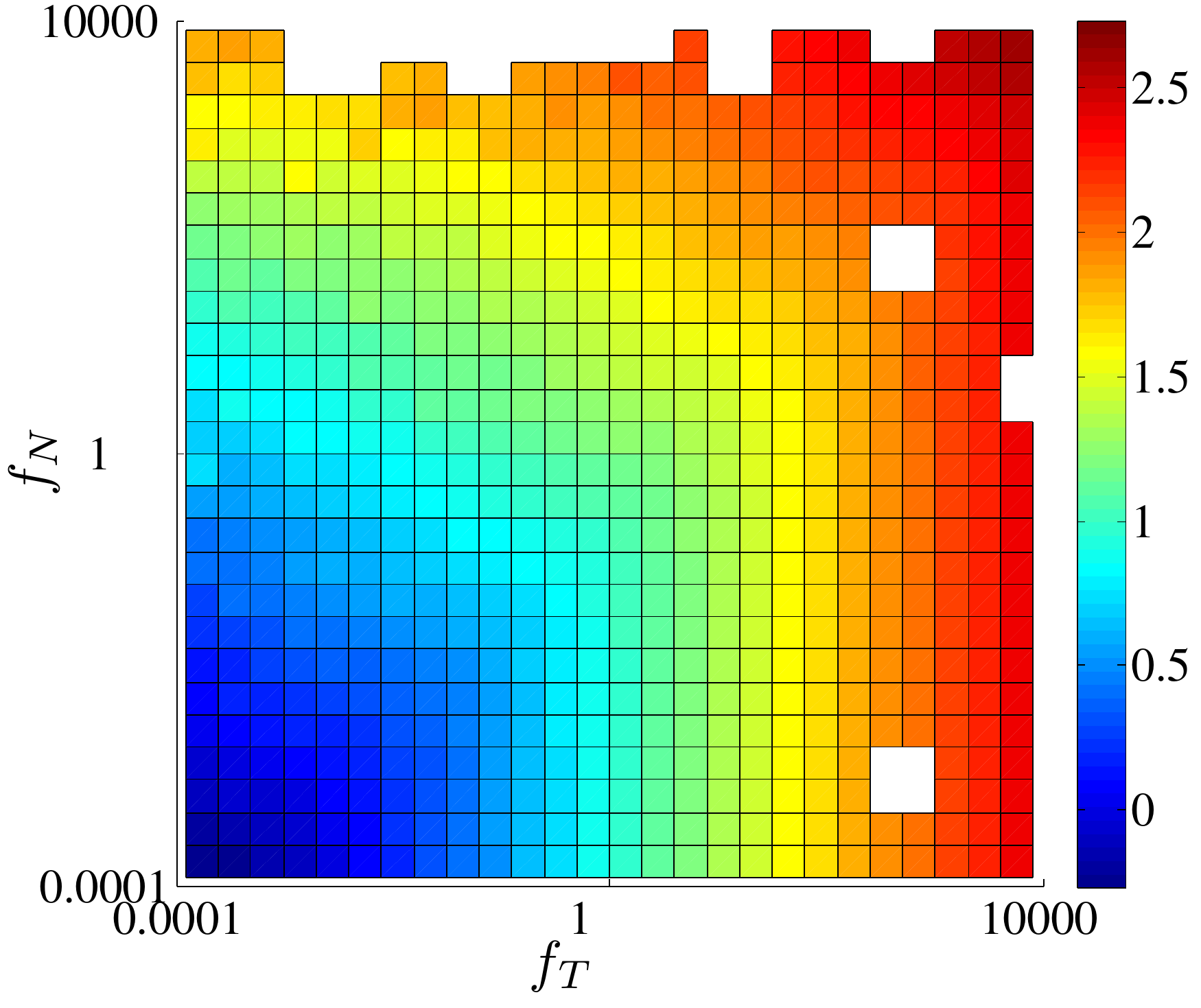}
	}\\
	\subfigure[]
	{   
	    \includegraphics[scale=0.3]{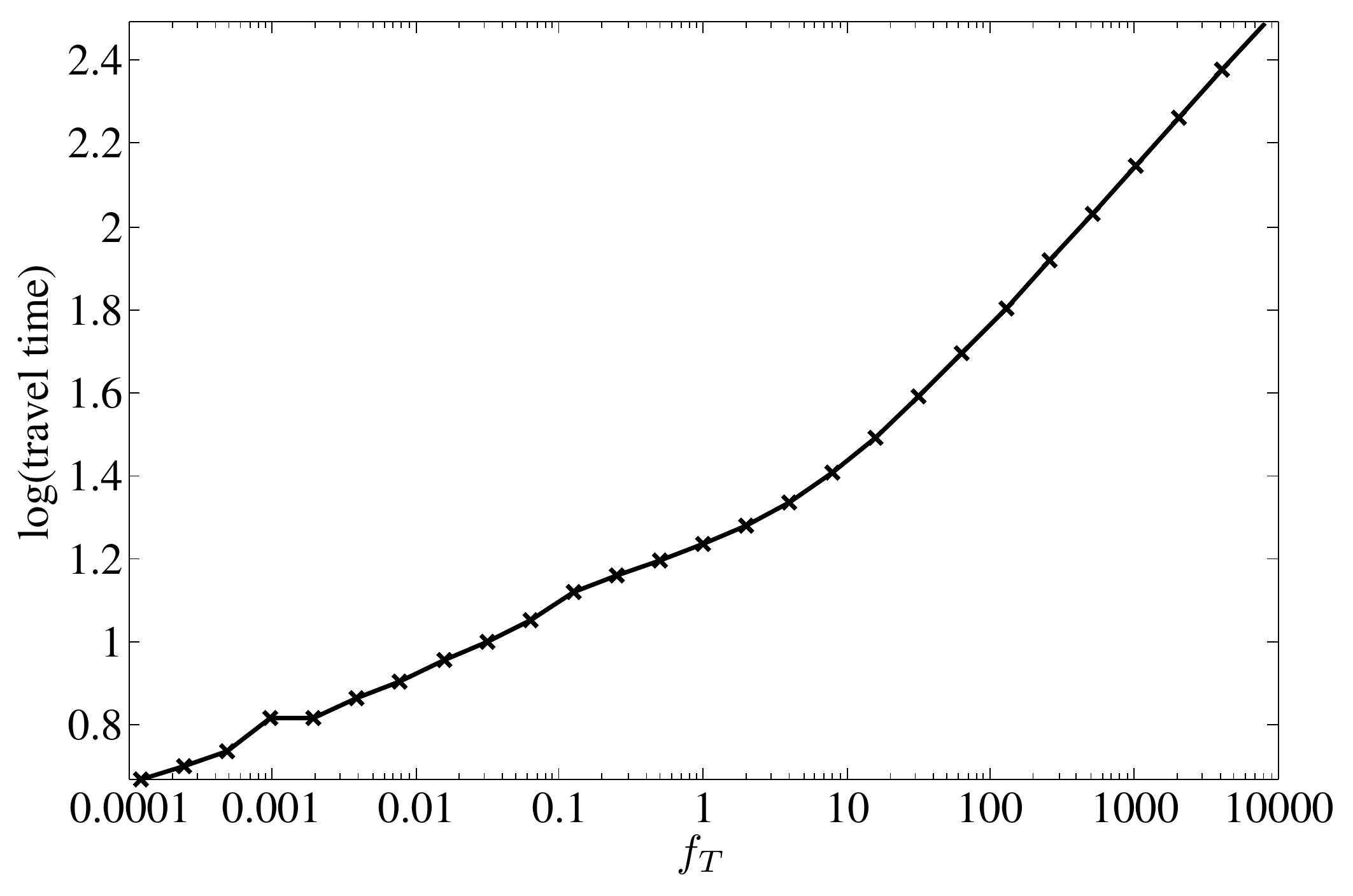}
	}
	\subfigure[]
	{
	    \includegraphics[scale=0.3]{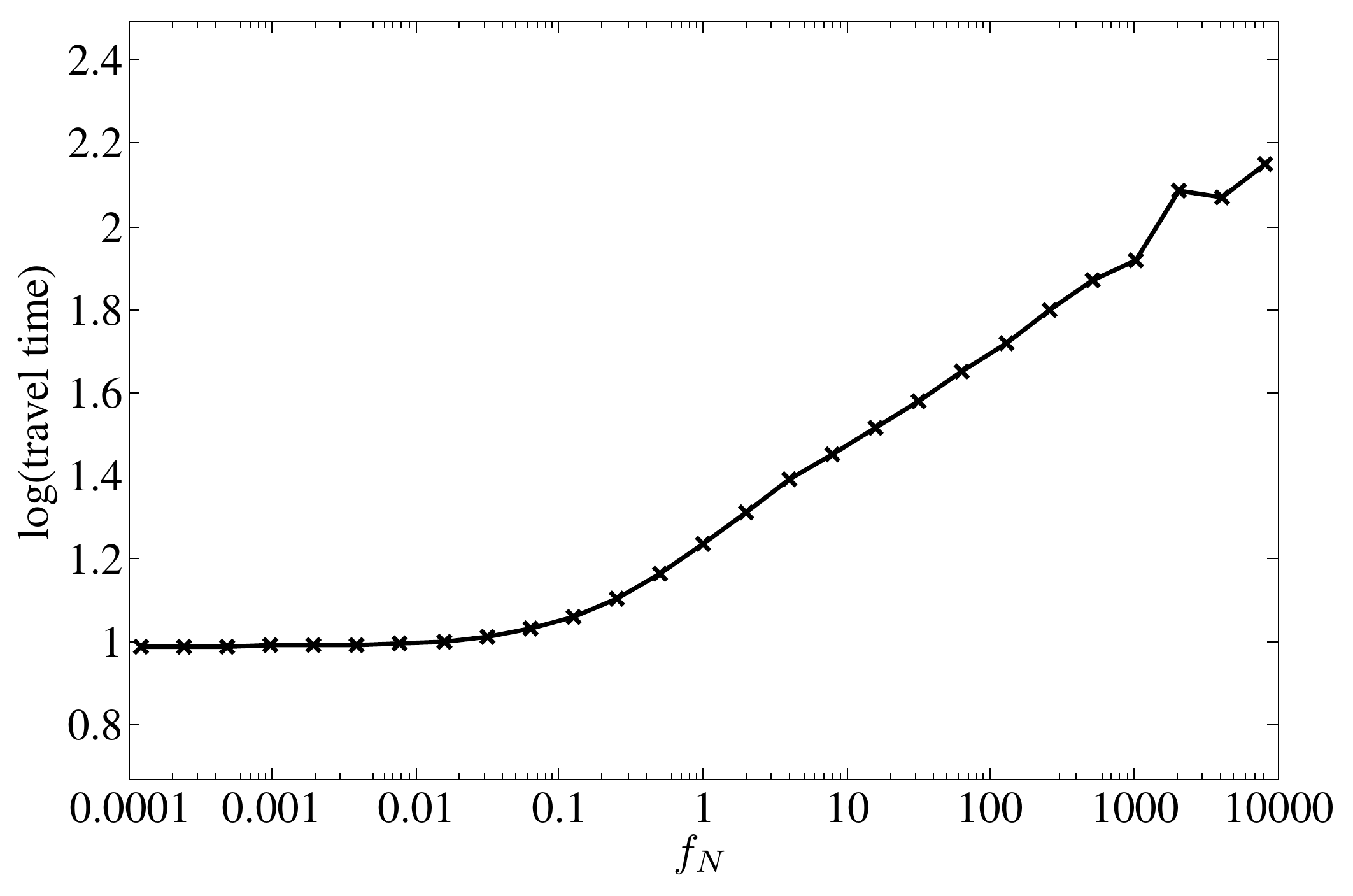}
	}
\end{center}
\caption{Effect of weights on travel time.}{\small{
(a) Surface plot of $\log \tau$ as a function of $f_{\tny T}$ and $f_{\tny N}$ on
a log-log scale. (b) Top
view of the surface plot. (c) Slice of the surface plot at $f_{\tny N} = 1$. 
(d) Slice of the surface plot at $f_{\tny T} = 1$.}} 
\label{fig:weights_and_tau} 
\end{figure}

\begin{figure}[p]
\begin{center}
	\subfigure[]
	{
		\includegraphics[scale=0.34]{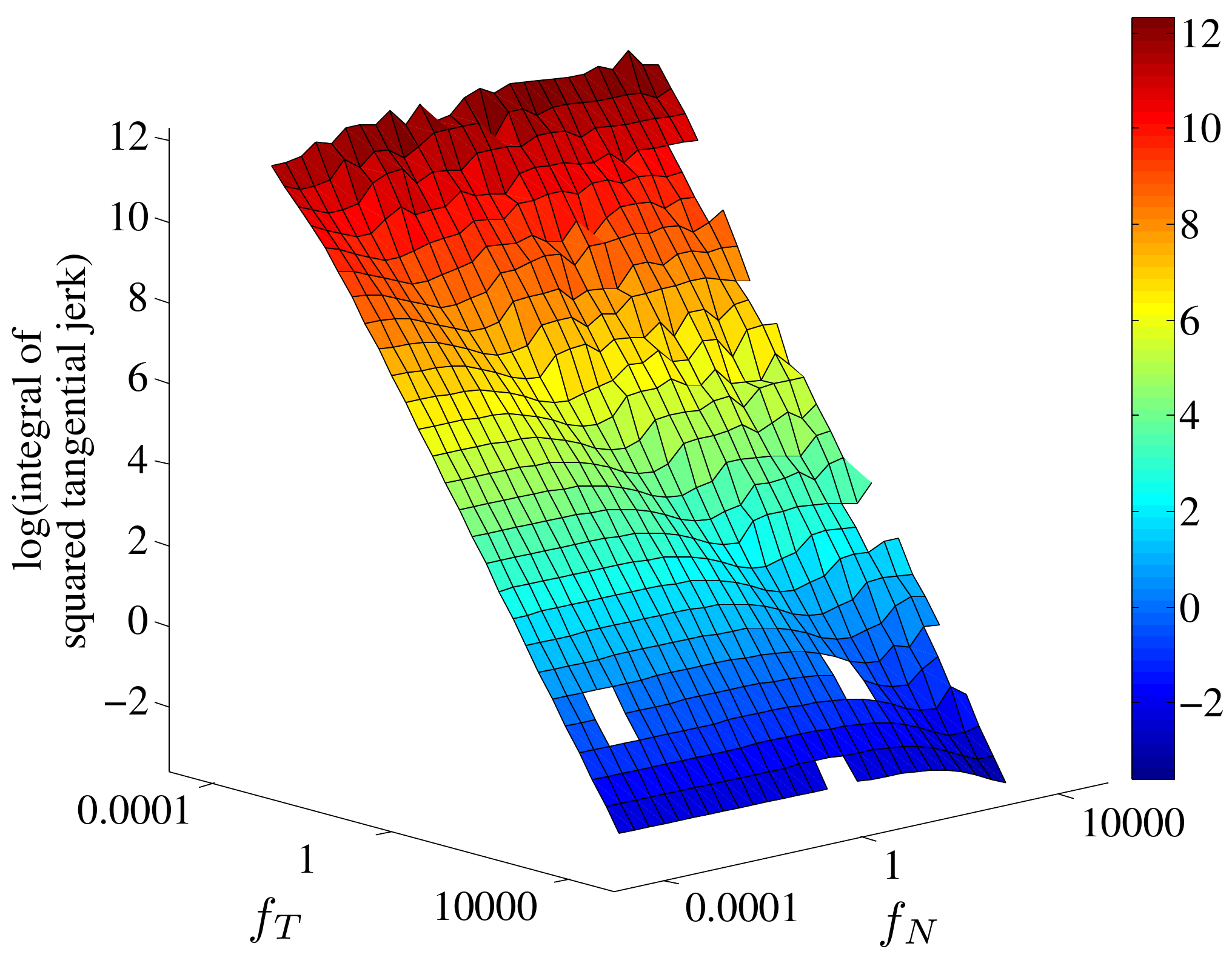}
	}
	\subfigure[]
	{
		\includegraphics[trim = 0mm -15mm 0mm 0mm, clip,scale=0.34]{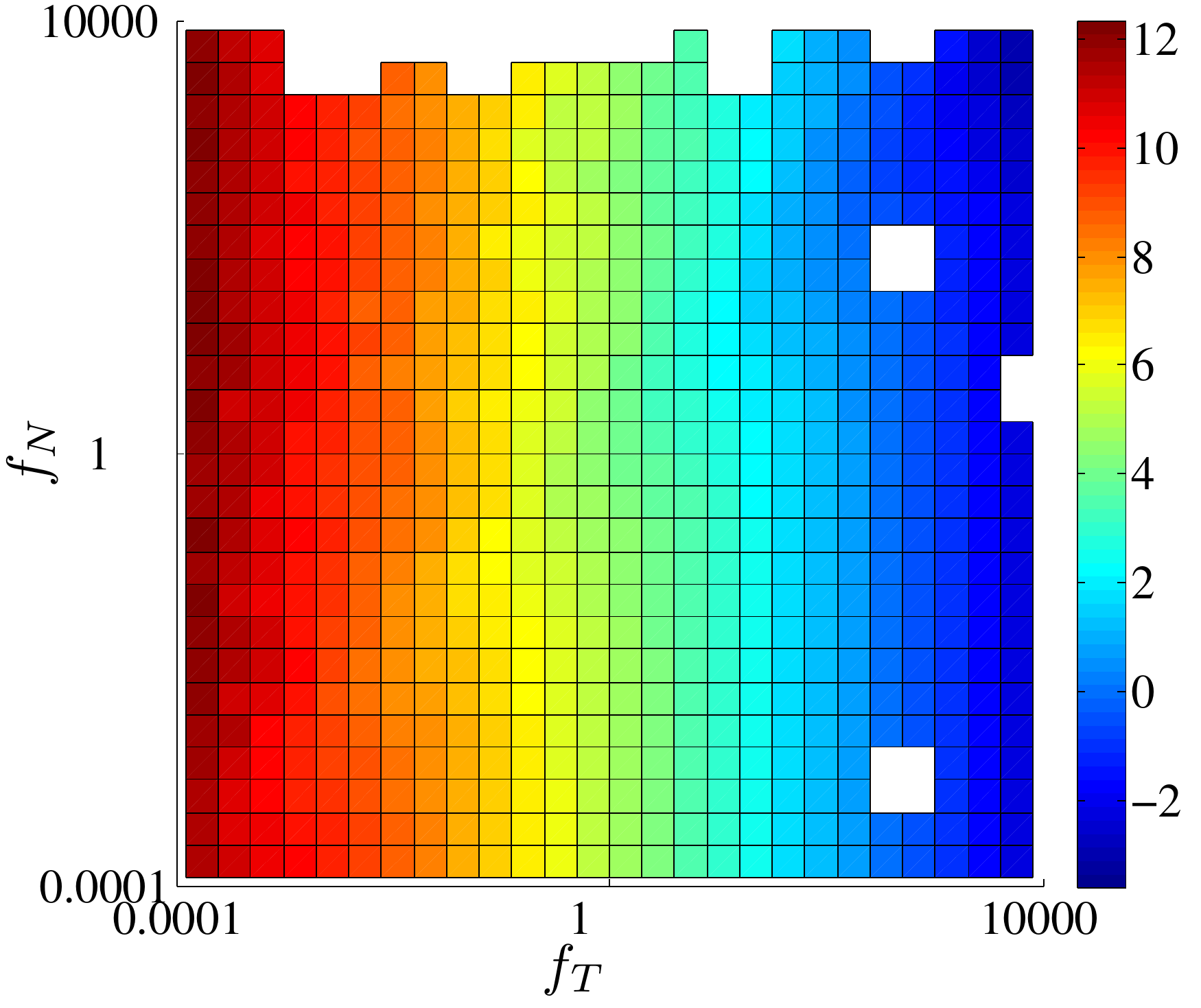}
	}\\
	\subfigure[]
	{   
	    \includegraphics[scale=0.3]{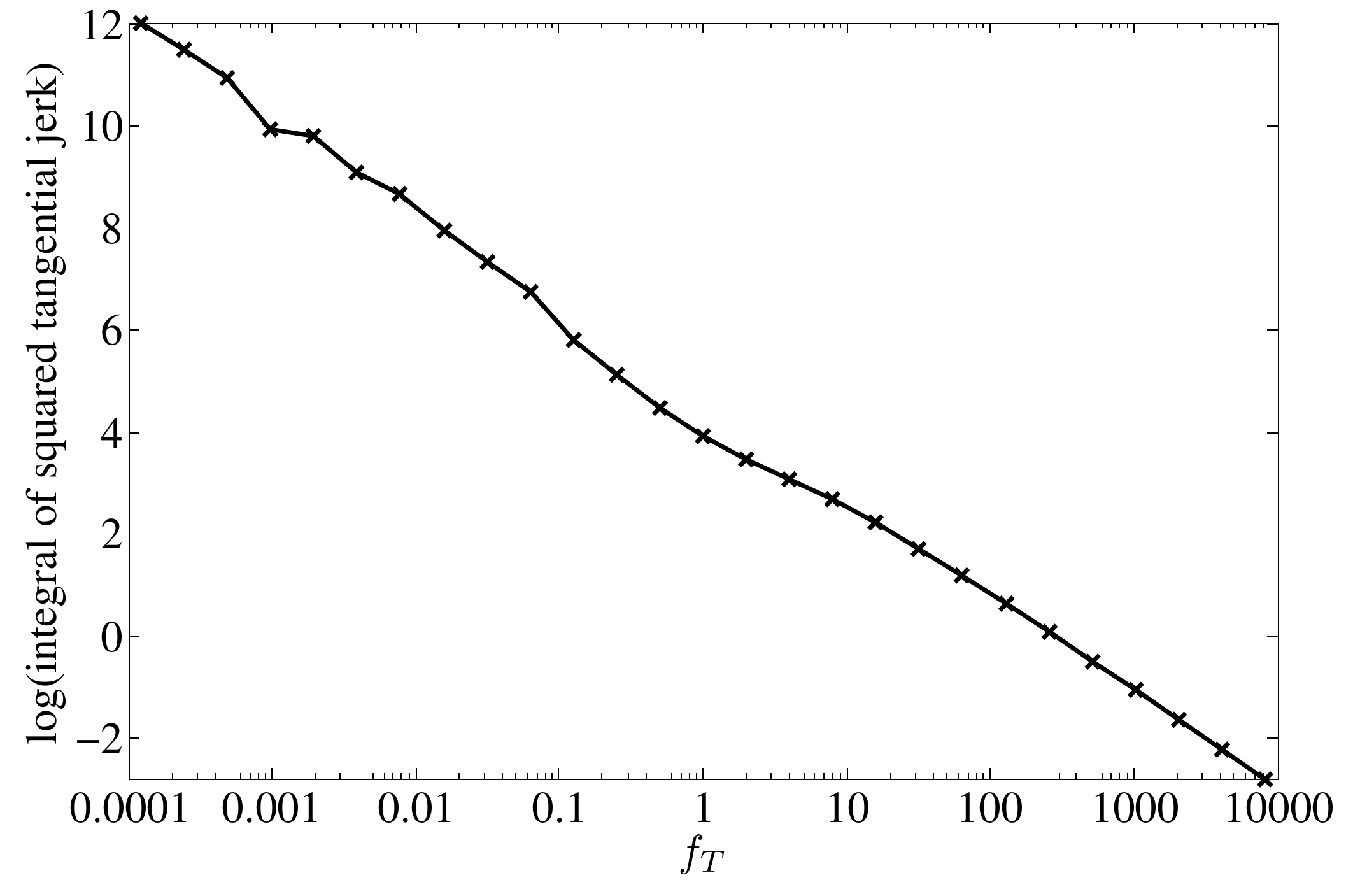}
	}
	\subfigure[]
	{
	    \includegraphics[scale=0.3]{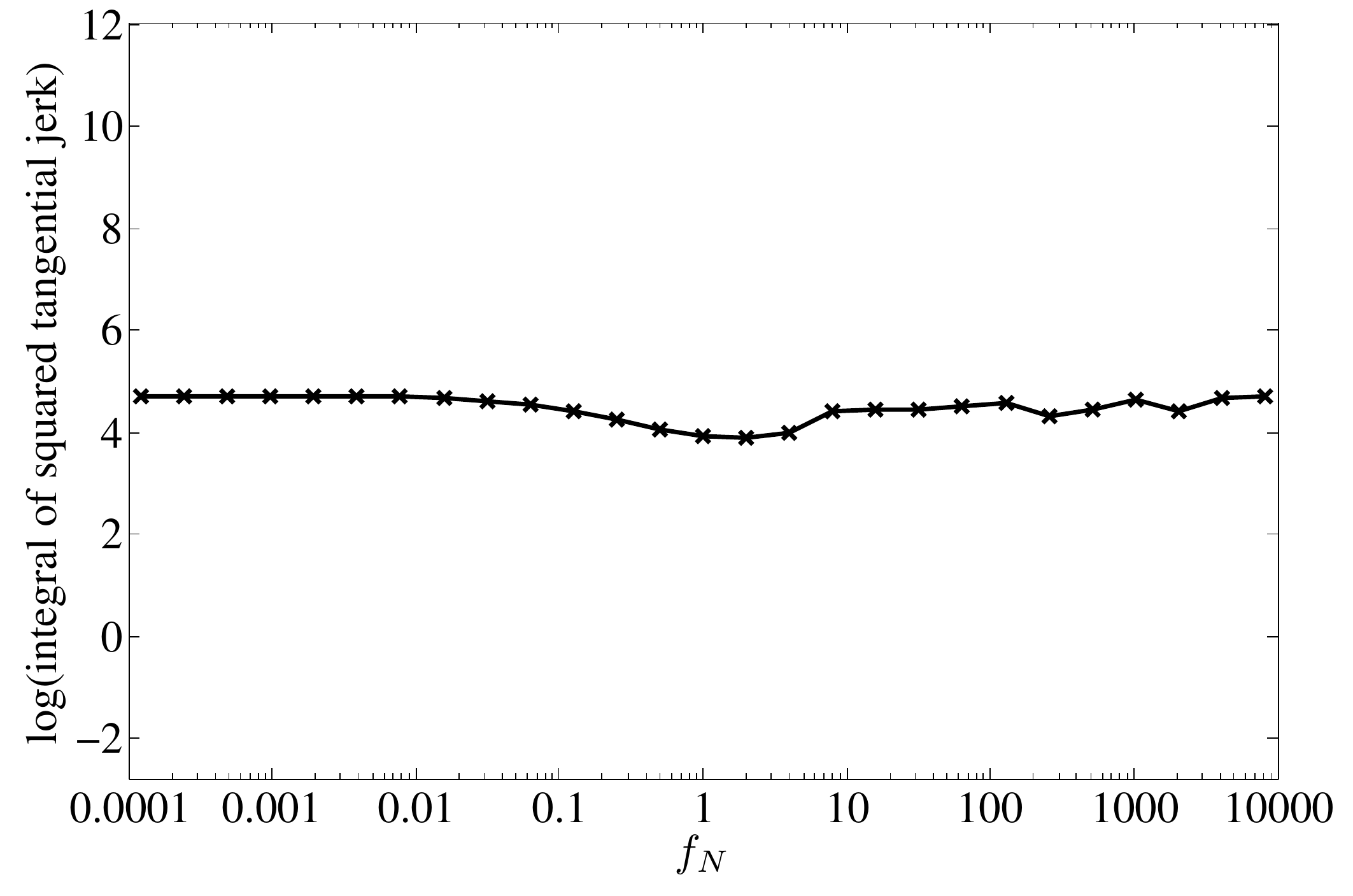}
	}
	\end{center}
\caption{Effect of weights on integral of squared tangential jerk.}{\small{
(a) Surface plot of log(integral of squared tangential jerk) as a function of $f_{\tny T}$ and $f_{\tny N}$ on
a log-log scale. (b) Top
view of the surface plot. (c) Slice of the surface plot at $f_{\tny N} = 1$. 
(d) Slice of the surface plot at $f_{\tny T} = 1$.}} 
\label{fig:weights_and_jt_sq} 

\begin{center}
	\subfigure[]
	{
		\includegraphics[scale=0.34]{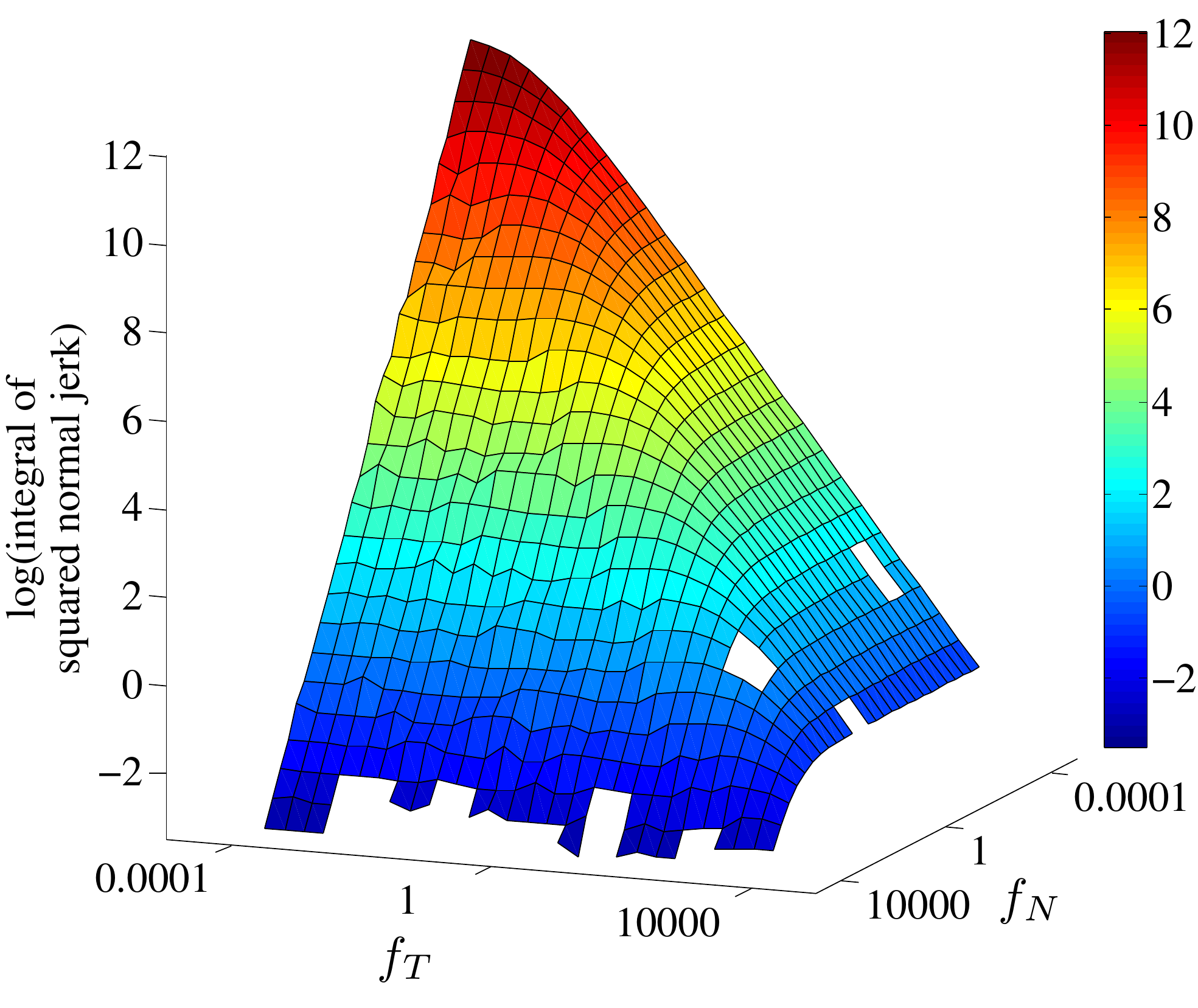}
	}
	\subfigure[]
	{
		\includegraphics[trim = 0mm -25mm 0mm 0mm, clip,scale=0.34]{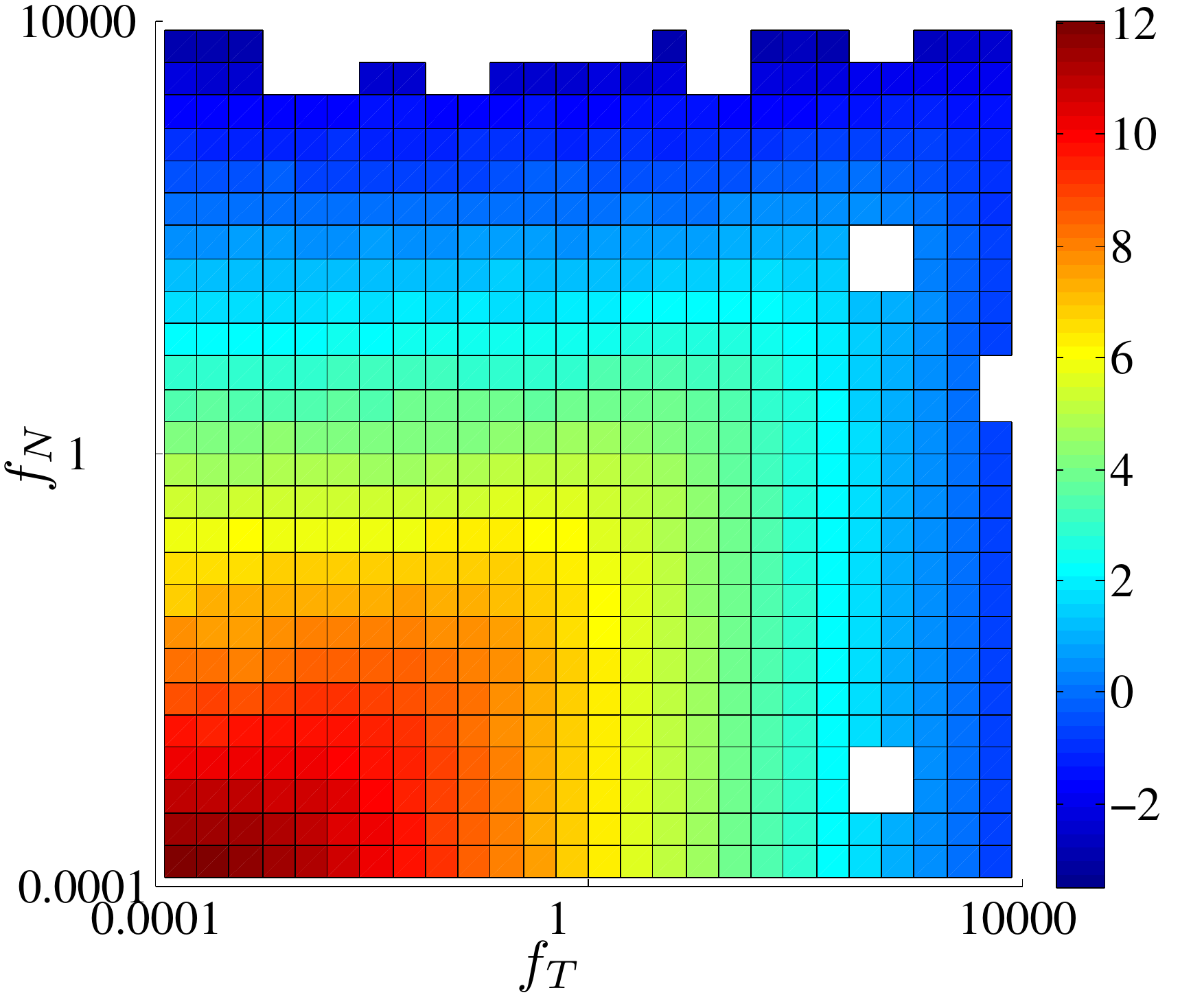}
	}
	\subfigure[]
	{   
	    \includegraphics[scale=0.3]{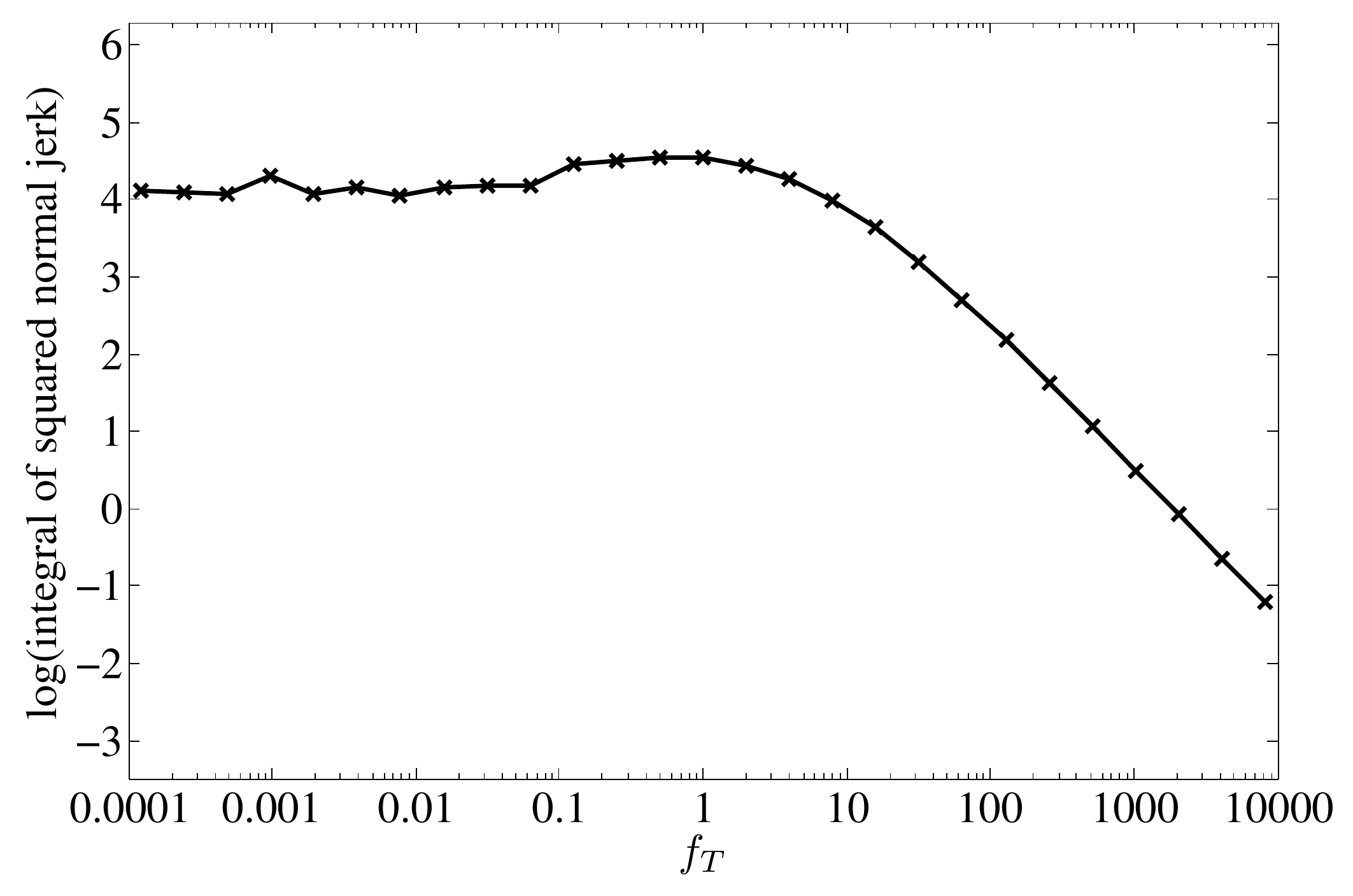}
	}
	\subfigure[]
	{
	    \includegraphics[scale=0.3]{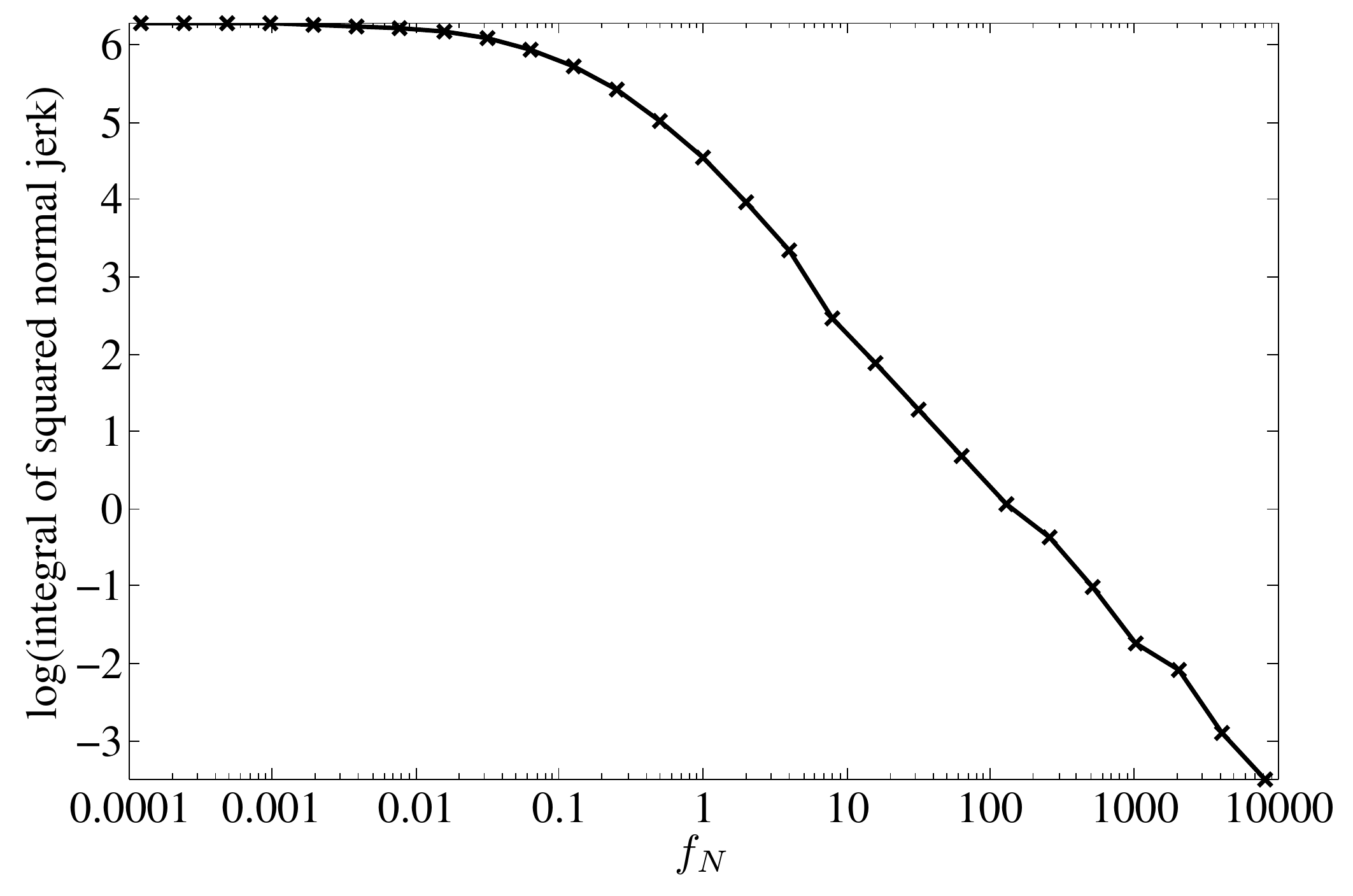}
	}
\end{center}
\caption{Effect of weights on integral of squared normal jerk.}{\small{
(a) Surface plot of log(integral of squared tangential jerk) as a function of $f_{\tny T}$ and $f_{\tny N}$ on
a log-log scale. (b) Top
view of the surface plot. (c) Slice of the surface plot at $f_{\tny N} = 1$. 
(d) Slice of the surface plot at $f_{\tny T} = 1$.}} 
\label{fig:weights_and_jn_sq} 
\end{figure}

Figures~\ref{fig:weights_and_tau},~\ref{fig:weights_and_jt_sq}, and~\ref{fig:weights_and_jn_sq}
show $\tau$, $J_{\tny T}$ and $J_{\tny N}$ respectively. In each figure, part
(a) shows log of the respective quantity as a function of $f_{\tny T}$ and $f_{\tny N}$ on a log-log-log scale.
Part (b) is a  top view of the surface plot above. Part (c) shows slices of this surface plot 
at $f_{\tny N} = 1$ and $f_{\tny T} = 1$ respectively.

\begin{figure}
\begin{center}
	\includegraphics[scale=0.74]{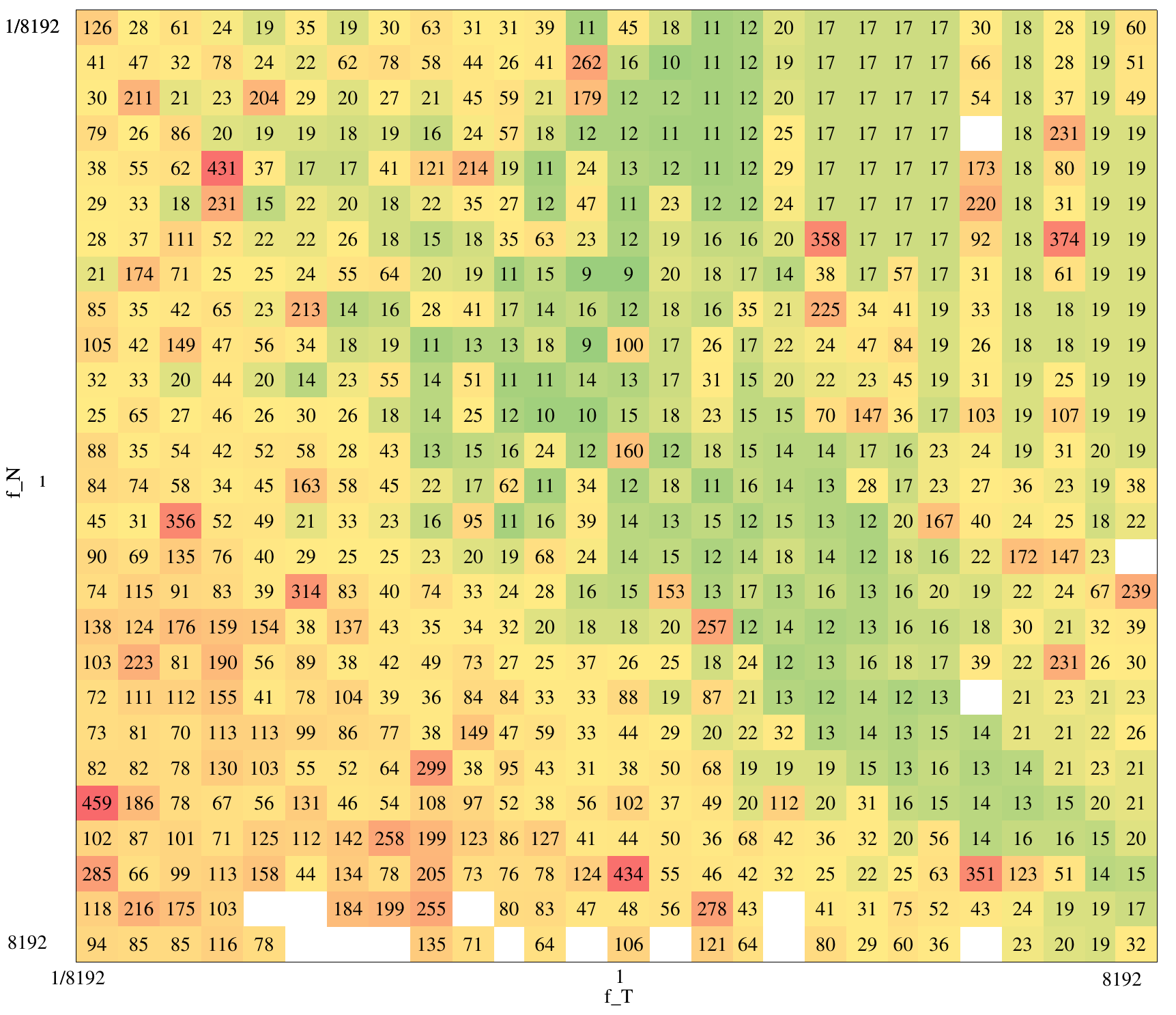}
\end{center}
\caption{Number of iterations for the range of weight factors.}{\small{The green cells
indicate smaller number of iterations compared to red cells. It is clear that
the least number of iterations are taken in the region where both factors are
greater than 1/32 and one factor is roughly within 1/16 to 16 times the other.
The empty cells correspond to problems that failed to converge.}}
\label{fig:weights_and_n_iters}
\end{figure}

The ``holes'' in the surface plots correspond to the problems that did not converge
to a solution. In general, the surfaces are rougher and there are more failures
when $f_{\tny N}$ is much larger than $f_{\tny T}$. This indicates that
the problem becomes less ``stable'' as the weight factors are too imbalanced.
(In reality there are more holes in the surfaces than there are non-convergent
problems. This is an unfortunate artifact of the plotting software
that we use. In the surface plot, a vertex corresponds to a problem rather than a 
cell. Thus, one non-convergent problem causes all the cells that share that vertex
to be removed. The actual non-convergent problems correspond to the
empty cells of \Fig{fig:weights_and_n_iters}). 

In this experiment, the ratio of tangential jerk weight to normal jerk weight has
been varied by nearly 8 orders of magnitude and we get solutions in almost all
cases.

From \Fig{fig:weights_and_tau}, we see that the travel time 
increases with increase in weights. This is expected since
large weights mean that the contribution of travel time to total discomfort is
relatively low compared to the contribution of the terms due to jerk. We also see 
that $\tau$ monotonically increases with $f_{\tny T}$. For low values of
$f_{\tny N}$, $\tau$ does not change appreciably with $f_{\tny N}$. As the value of 
$f_{\tny N}$ increases beyond a threshold, $\tau$ monotonically increases 
with $f_{\tny N}$. The rate of increase of $\tau$ with respect to
$f_{\tny T}$ is higher than it is with respect to $f_{\tny N}$.

From \Fig{fig:weights_and_jt_sq}, we observe that $\log J_{\tny T}$ decreases 
linearly with $\log f_{\tny T}$ while it is almost constant with
respect to $\log f_{\tny N}$.  Thus, the integral of squared tangential jerk, $J_{\tny T}$, is related to $f_{\tny T}$ by a power law.

From \Fig{fig:weights_and_jn_sq} we see that for low values of
$f_{\tny T}$, $J_{\tny N}$ does not change appreciably with $f_{\tny T}$. As the value of 
$f_{\tny T}$ increases beyond a threshold, $J_{\tny N}$ monotonically decreases 
with $f_{\tny T}$. A similar behavior is observed with respect to $f_{\tny N}$
although the threshold value appears lower than that for $f_{\tny T}$. Once
the values exceed the threshold, the rate
of change of $J_{\tny N}$ with respect to both $f_{\tny N}$ and $f_{\tny T}$ is
almost the same.

Thus, we see that the integral of squared tangential jerk, $J_{\tny T}$ is a function of $f_{\tny T}$ alone,
and travel time changes more rapidly by changing $f_{\tny T}$ compared to $f_{\tny N}$.
Integral of squared normal jerk, $J_{\tny N}$ is a function of both $f_{\tny T}$ and
$f_{\tny N}$. Whenever a relationship exist between $f_{\tny T}$ or $f_{\tny N}$
and any of the quantities travel time, integral of squared tangential jerk, and integral
of squared normal jerk, it is of the form of a power law.

From this analysis, we can draw some useful guidelines for customizing weights for comfort
even though the effect of weight on discomfort is nonlinear. Since $J_{\tny T}$ is
a function of $f_{\tny T}$ alone, we can devise experiments that allow a
user to choose $f_{\tny T}$ that keeps tangential jerk to an acceptable level.
For example, we can devise experiments that consist primarily of straight line
motion, and has zero speeds on both ends. In such a motion, normal component
of jerk will make none or minimal contribution to discomfort. Hence, it would
be easy to set $f_{\tny T}$. Next, we can devise experiments that consist
of at least some curved segments. The user can choose $f_{\tny N}$ to keep
normal jerk during this curved motion to an acceptable level. Because
of power law relationships, the weights should be varied in a geometric
manner rather than a linear manner for faster customization.

\Fig{fig:weights_and_n_iters} shows the number of iterations taken
by Ipopt to find a solution.  Apart from a few isolated outliers that require
large number of iterations, it is clear that the number of iterations is
small in the region where $f_{\tny T}$ is not too small compared to $f_{\tny N}$
and both factors are not too small either.
If $f_{\tny N}$ is much larger than $f_{\tny T}$, the problems still converge
in most cases but require many iterations.  Most of the failures are
when $f_{\tny N}$ is too large compared to unity.  Hence, we recommend
that for customization $f_{\tny N}$ should not be too large compared to $f_{\tny T}$
and both should be not too small compared to unity.
\vspace{1in}

\subsection{Reliability}
\label{sec:reliability}
To evaluate the reliability of our method, we construct a set of $7500$ problems with different
boundary conditions and solve the full constrained optimization problem
corresponding to each of the 4 initial guesses for each problem.
We do not include obstacles in this test. 

We generate the problem set as follows. Fix the initial position as 
$\left\{0,0 \right\}$ and orientation as 0. Choose final position
at different distances along radial lines from the origin.
Choose 10 radial lines that start from 0 degrees and go up to 
180 degrees in equal increments. The distance on the radial line
is chosen from the set $\left\{1, 2, 4, 8, 16\right\}$. The angle
of the radial line and the distance on the line determines the
final position. Choose 30 final orientations starting from 
0 up to 360 degrees (360 degrees not included) in equal increments. The 
speed, $v$, and tangential acceleration,
$a_{\tny T}$, at both ends are varied by choosing $\left\{v, a_{\tny T}
\right\}$ pairs from the set $\left\{ \left\{0, 0\right\}, \left\{1, -0.1\right\},
\left\{1, 0\right\}, \left\{1, 0.1 \right\}, \left\{3, 0 \right\} \right\}$.
Thus we have 10 radial lines, 5 distances on each radial line, 30 orientations,
5 $\left\{v, a_{\tny T} \right\}$ pairs, resulting in
$10 \times 5 \times 30 \times 5 = 7500$ cases.

Each problem has $189$ degrees of freedom, $2018$ constraints, out of
which $66$ are equality constraints and $1952$ are inequality constraints.
For computation of initial guess of path, we set the maximum number
of iterations to 100. For discomfort minimization problem we set
the maximum number of iterations to 200. An average of 3.6 solution 
paths were found for each problem. This average would be higher
if we set the maximum number of iterations even higher. However,
since we wanted to evaluate how reliably our method performed in a
reasonable amount of computation time, we kept the maximum
number of iteration as 200.

\begin{figure}[b!]
\begin{center}
	\subfigure[][99\% solved within 4 s.]
	{
		\includegraphics[trim = 0mm 0mm 20mm 0mm, clip, scale = 0.18] {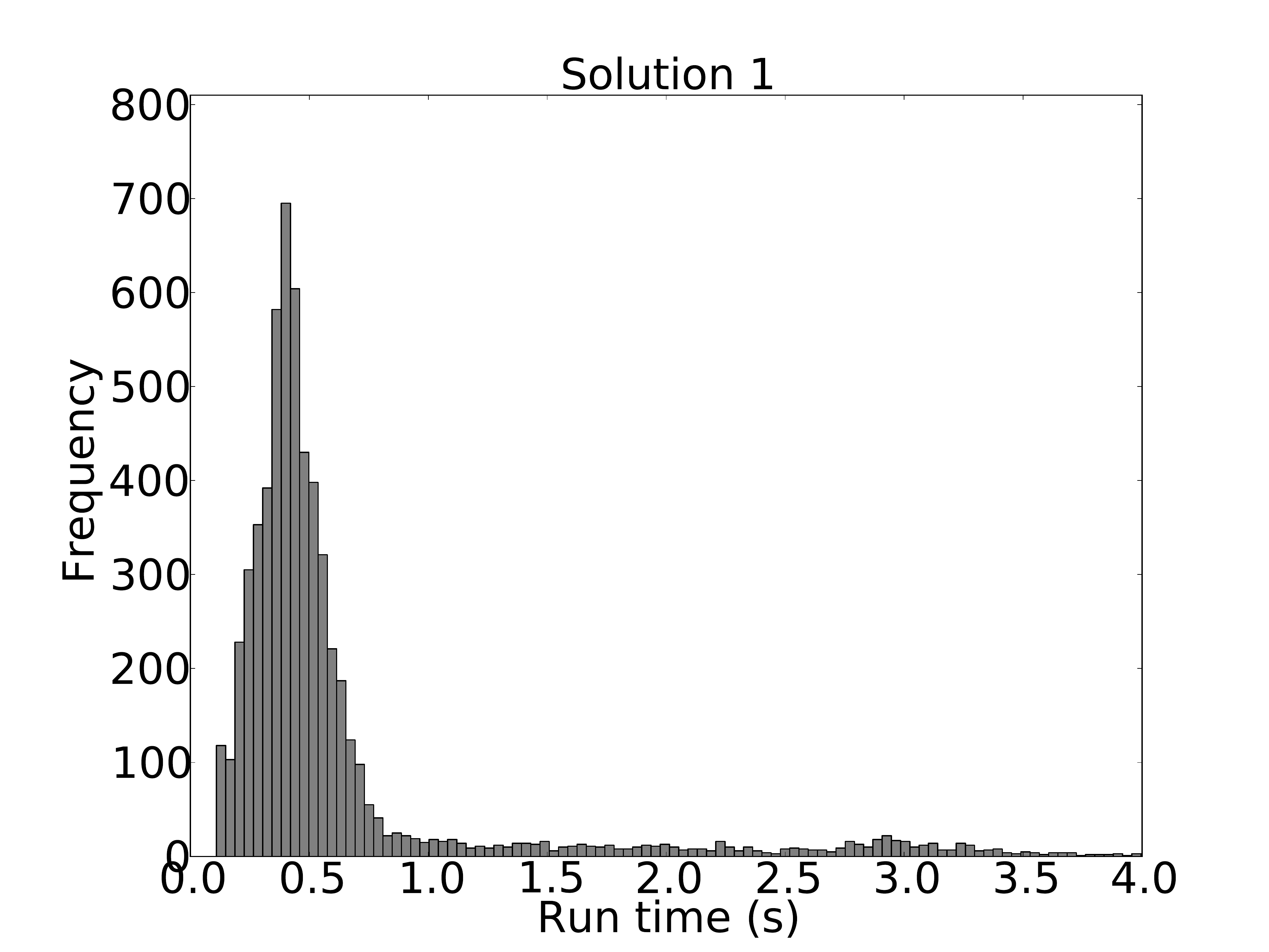}
	}
	\subfigure[][99\% solved within 4 s.]
	{   
	    \includegraphics[trim = 0mm 0mm 20mm 0mm, clip, scale = 0.18] {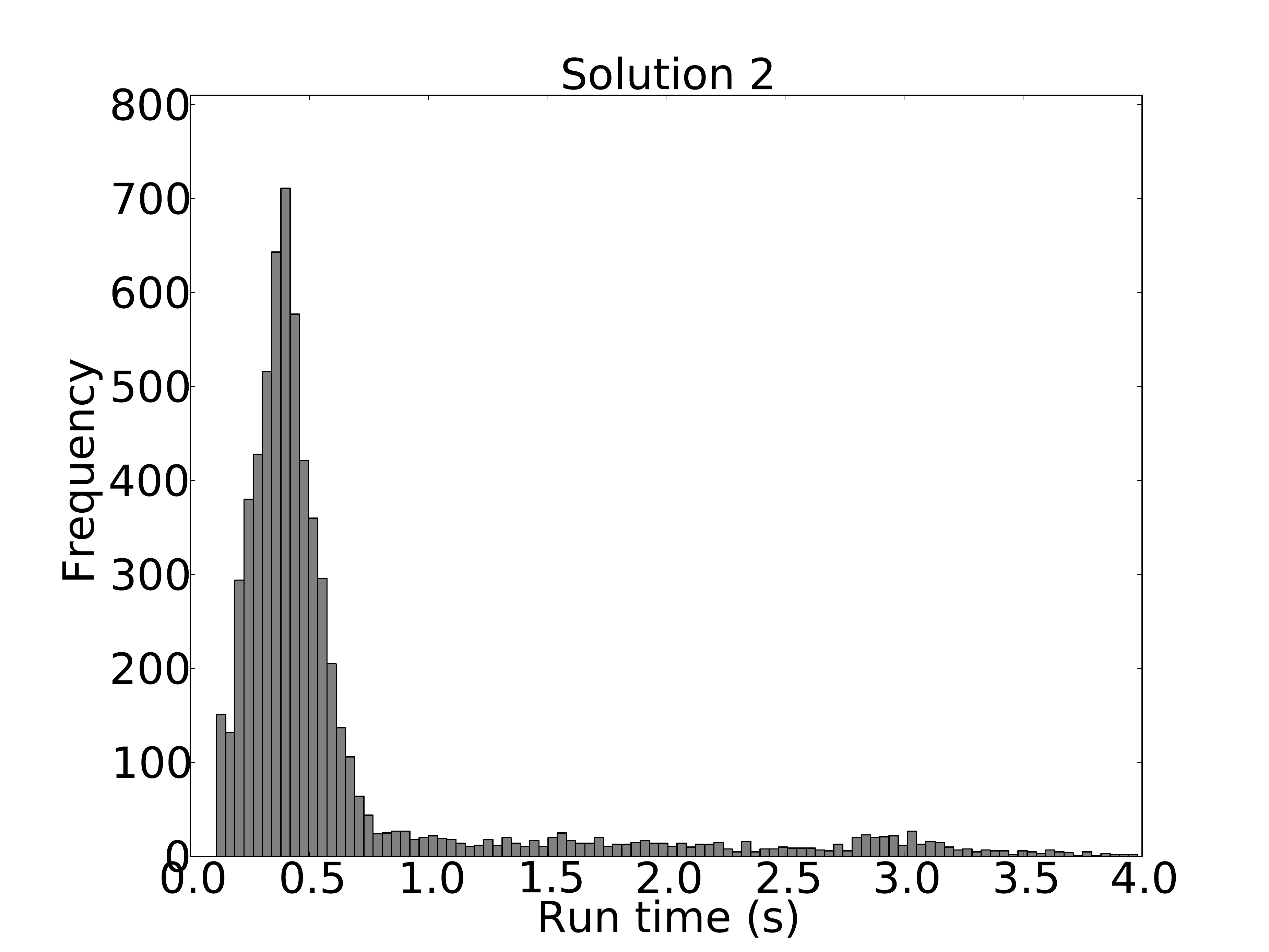}
	}\\
	\subfigure[][99\% solved within 4 s.]
	{
	    \includegraphics[trim = 0mm 0mm 20mm 0mm, clip, scale = 0.18]
	    {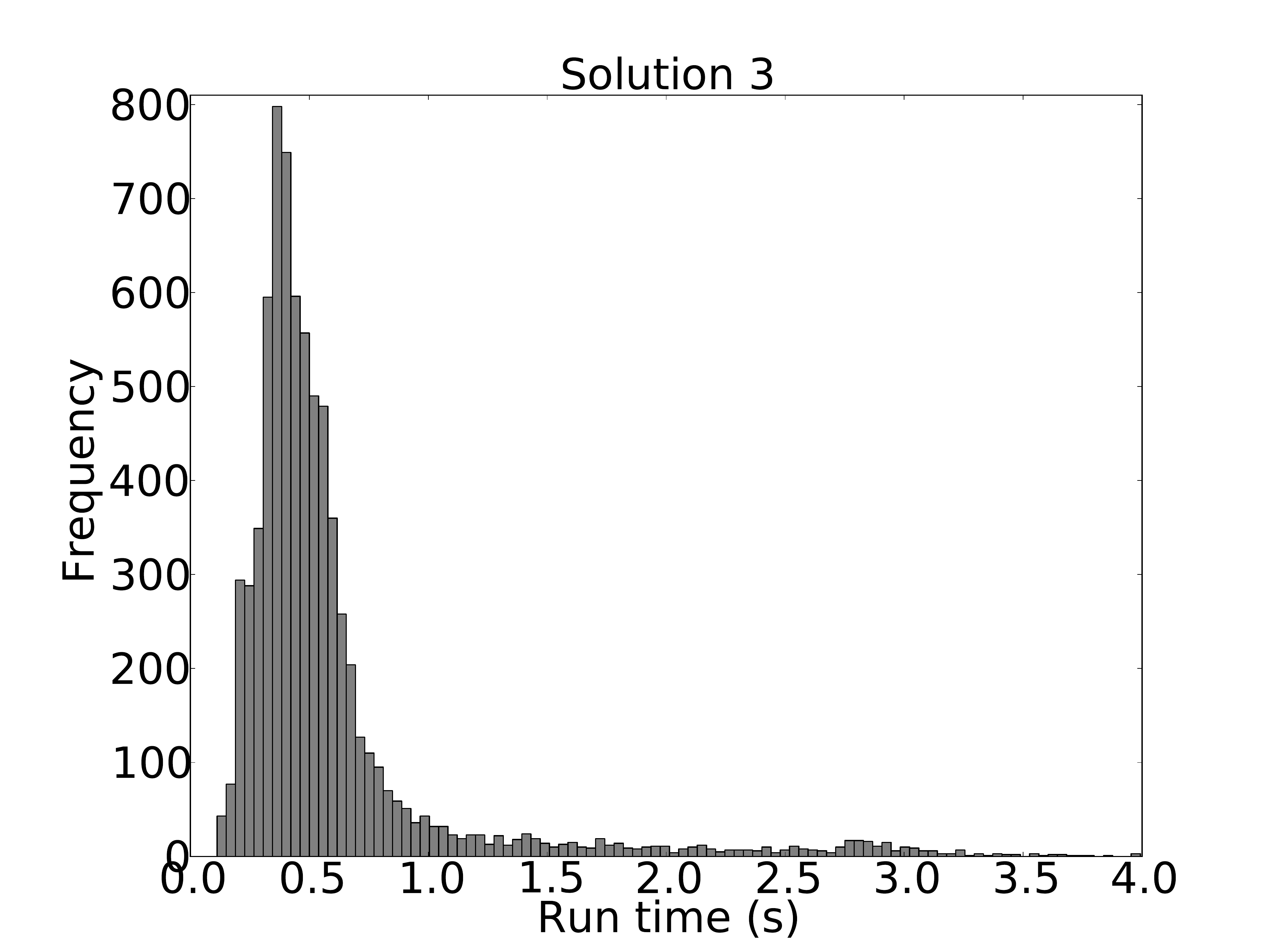}
	}
	\subfigure[][100\% solved within 4 s.]
	{
	    \includegraphics[trim = 0mm 0mm 20mm 0mm, clip, scale = 0.18]
	    {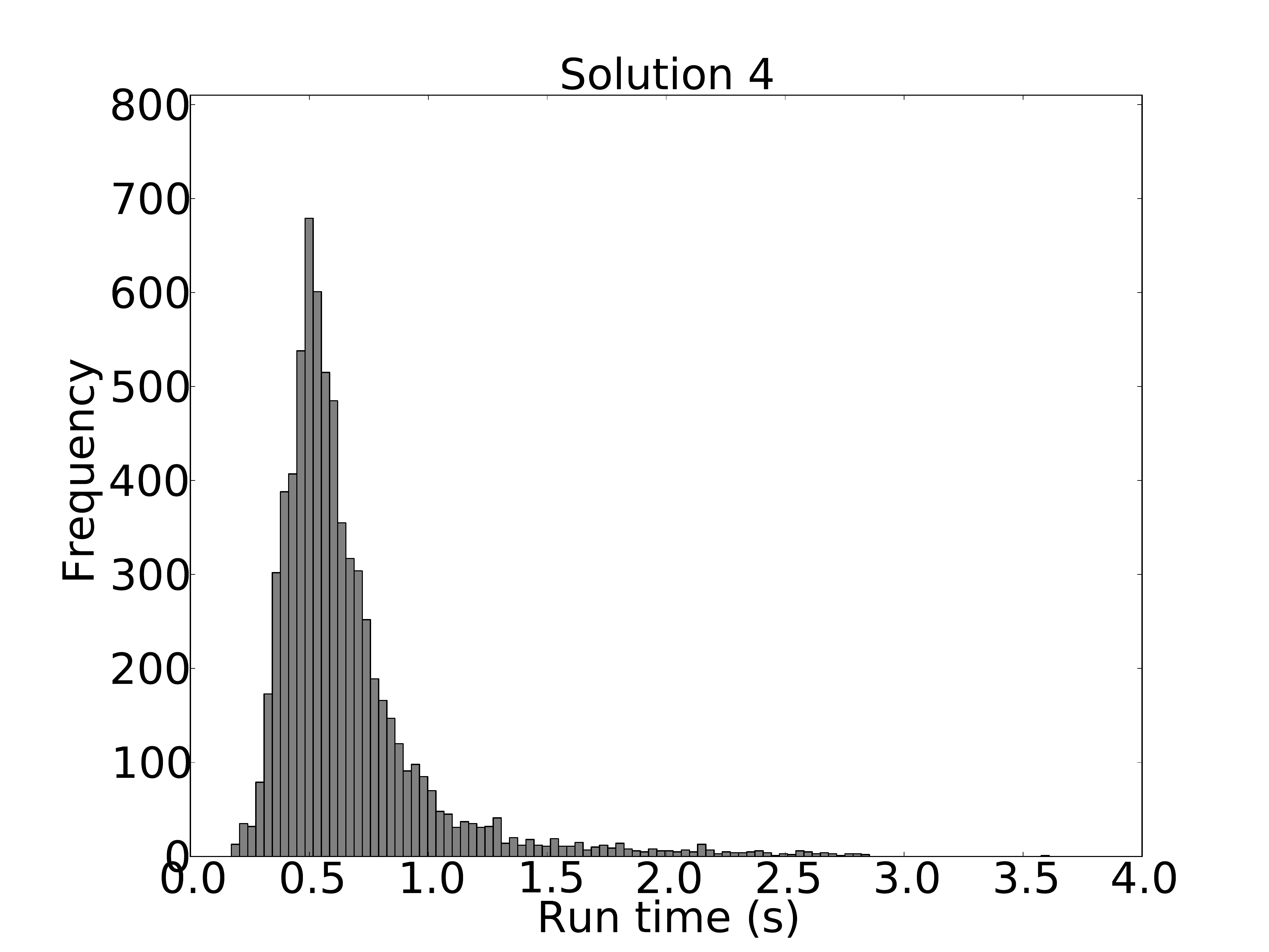}
	}
\end{center}
\caption{Histogram of time taken to compute solution of discomfort minimization problem.}
{\small{This includes both successful and unsuccessful cases. Total 7500 cases.}}
\label{fig:full_problem_time}
\end{figure}

All the problems were solved on a computer with an Intel Core i7 CPU running
at 2.67 GHz, 4 GB RAM, and 4 MB L-2 cache size. Histogram of run-time for 
computing the solution of the discomfort minimization problem is shown in 
\Fig{fig:full_problem_time} respectively. 
In this histogram, we have removed 1\% of cases that lie 
outside the range of the axis shown for better visualization. 
This histogram shows both successful and unsuccessful
cases. 

For each of the 7500 problems, each of the successful initial guesses was 
used to compute a solution of the discomfort minimization problem. 
For all the problems, at least one successful solution 
was computed. Table~\ref{tab:successful_prob_stats} shows the number of problems for which
one, two, three, or four solutions were successfully computed.

\begin{figure}
\begin{center}
	\subfigure[99\% solved within 4 s.]
	{
		\includegraphics[trim = 0mm 0mm 20mm 0mm, clip, scale = 0.18] {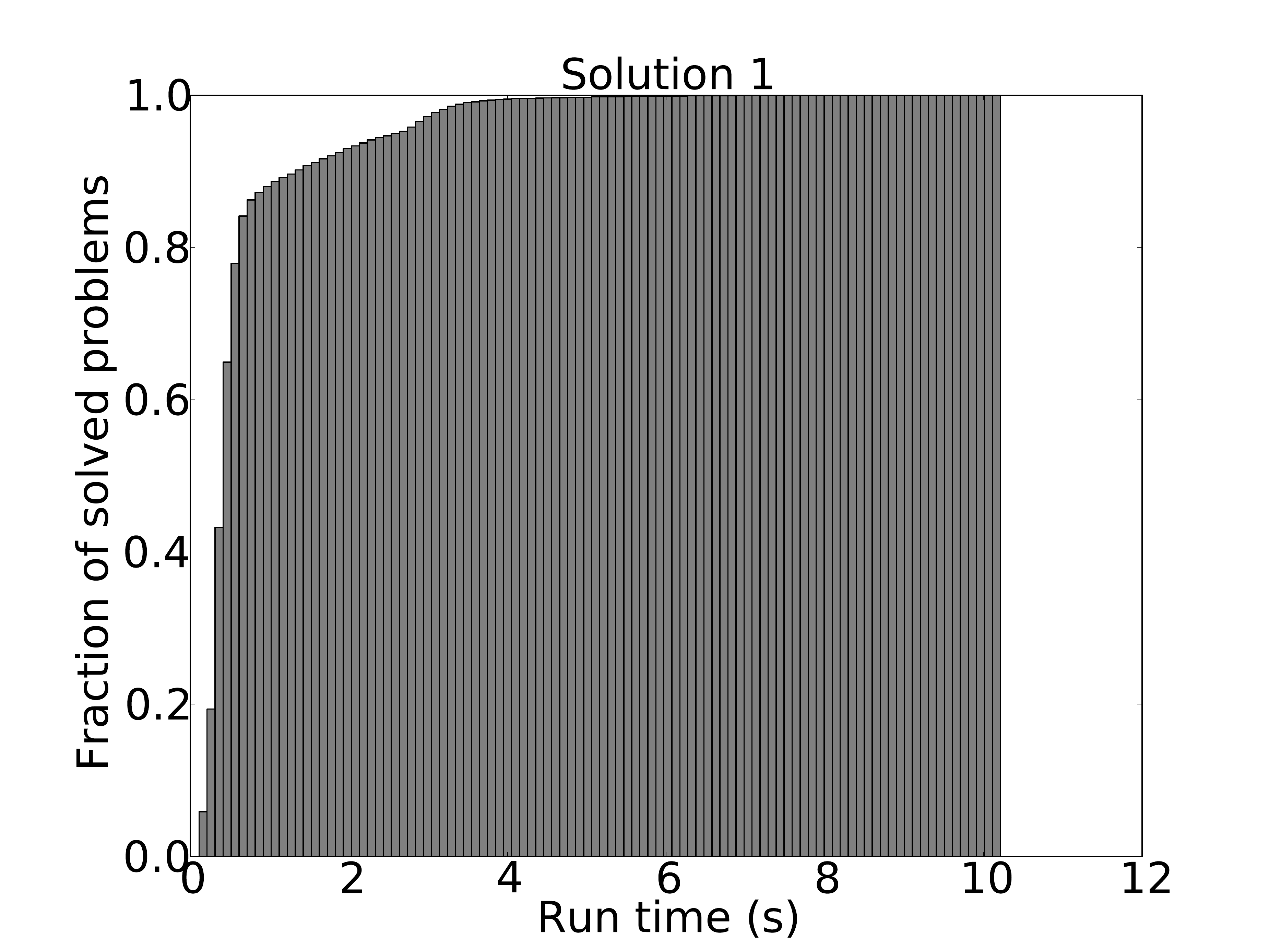}
	}
	\subfigure[99\% solved within 4 s.]
	{   
	    \includegraphics[trim = 0mm 0mm 20mm 0mm, clip, scale = 0.18] {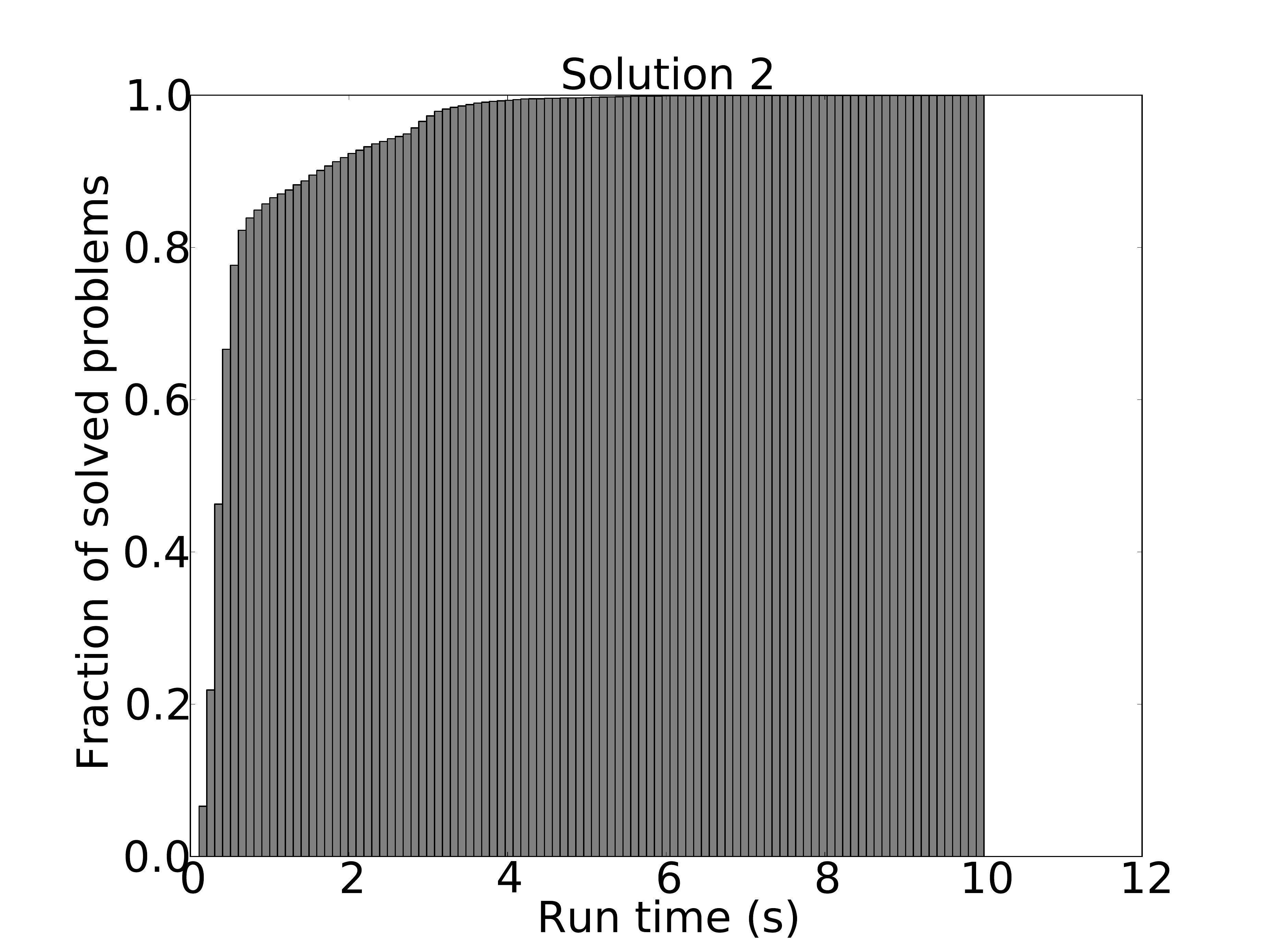}
	}\\
	\subfigure[99\% solved within 4 s.]
	{
	    \includegraphics[trim = 0mm 0mm 20mm 0mm, clip, scale = 0.18]
	    {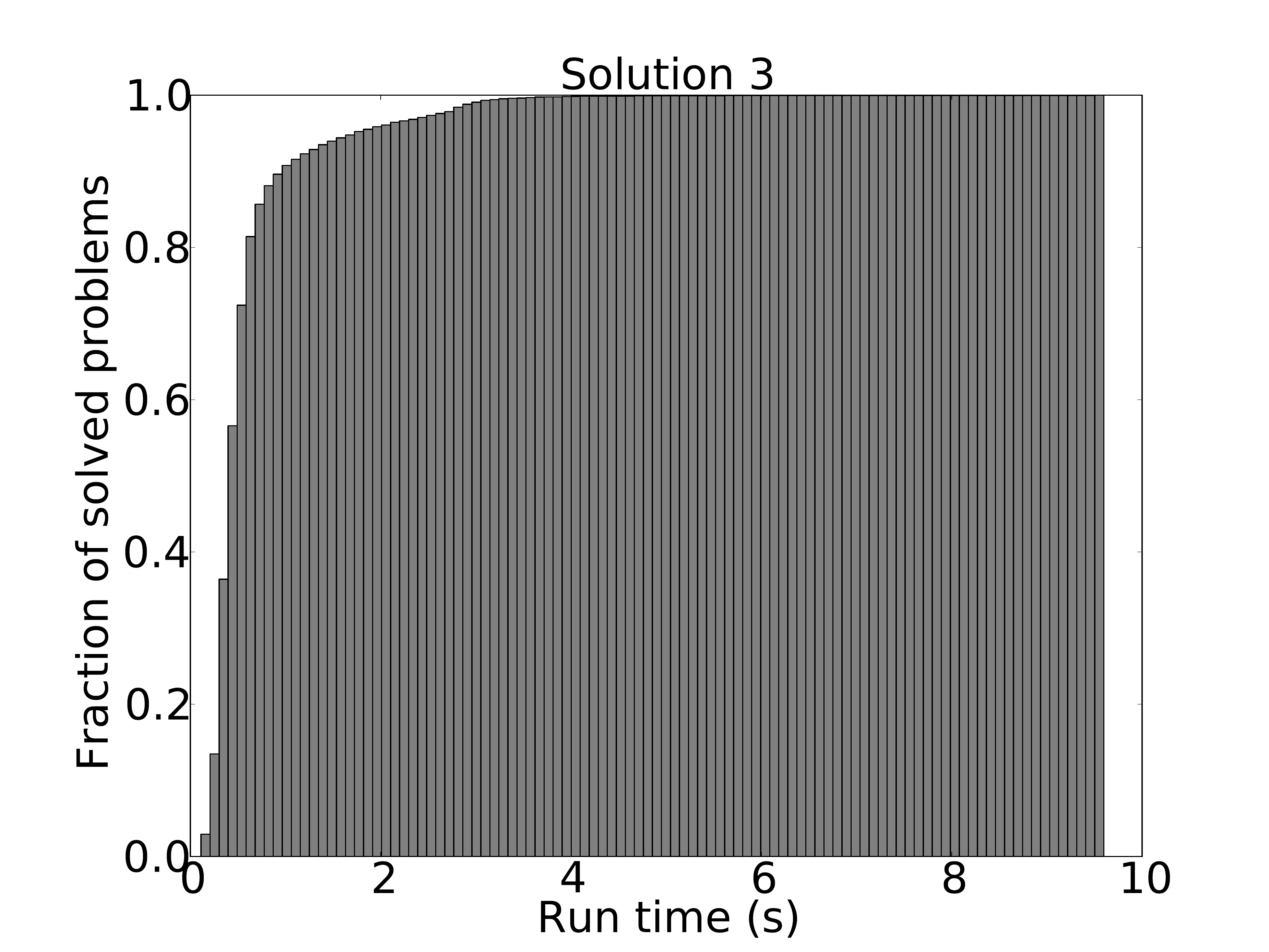}
	}
	\subfigure[100\% solved within 4 s.]
	{
	    \includegraphics[trim = 0mm 0mm 20mm 0mm, clip, scale = 0.18]
	    {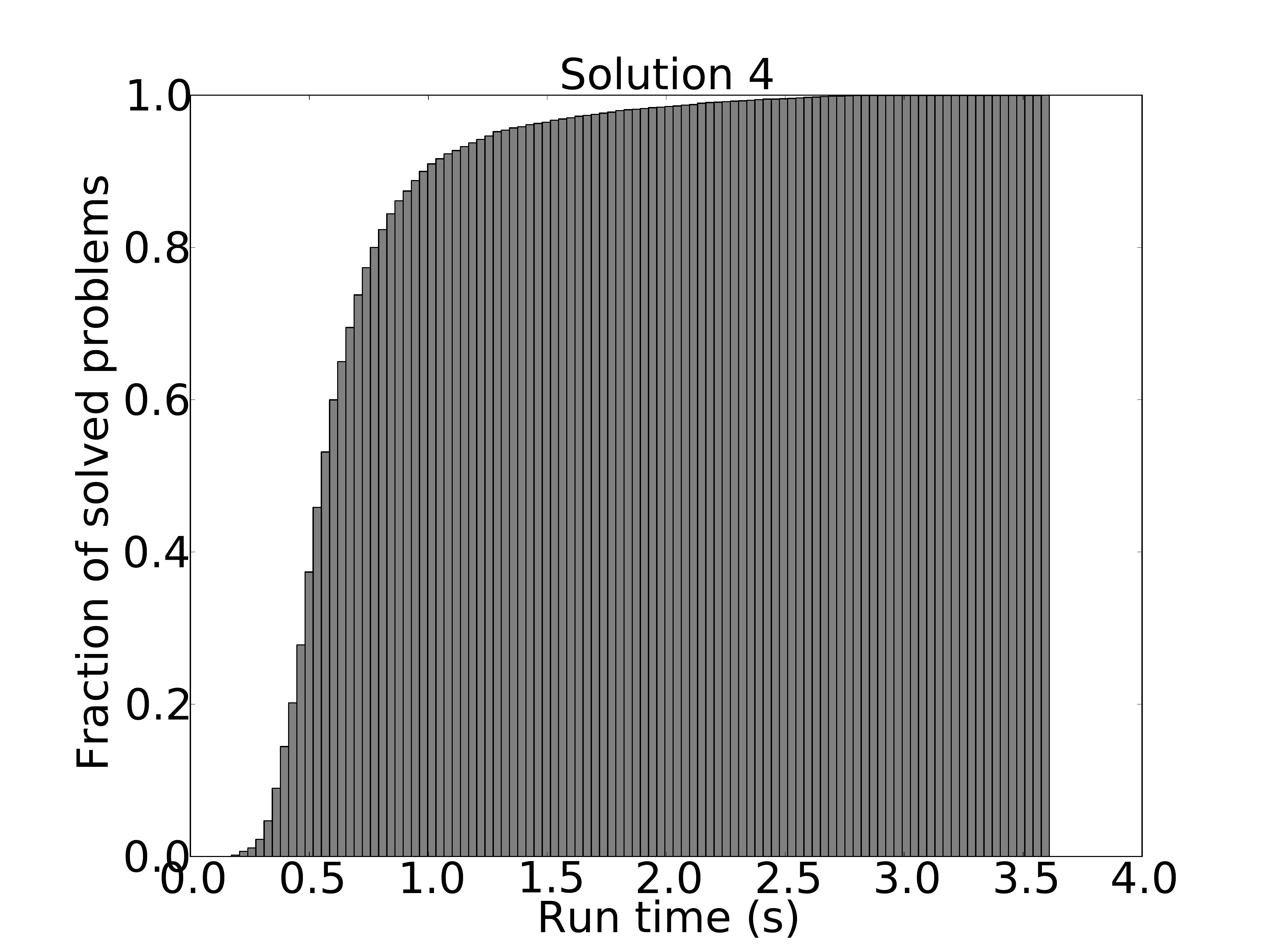}
	}
\end{center}
\caption{Normalized cumulative histogram of time taken to compute solution of discomfort minimization problem.}{\small{This includes both successful and unsuccessful cases. Total 7500 cases.}}
\label{fig:full_problem_timecum}
\end{figure}

From \Fig{fig:full_problem_time} we see that 99\% or more
of the solutions of the full problem are computed in less than
4 seconds. To get an estimate of the percentage of outliers,
we fit a Gaussian to each of the four guesses. 
The results are shown in
Table~\ref{tab:full_problem_gaussians}.
This is a simplified approximation and should
be seen just as an indicator of reliability of the method.
Time taken to compute the solution is further visualized 
in \Fig{fig:full_problem_timecum} that shows a normalized 
cumulative histogram.

\begin{figure}
\begin{center}
	\subfigure[][89\% solved in 100 iterations or less]
	{
		\includegraphics[trim = 0mm 0mm 35mm 0mm, clip, scale = 0.19] {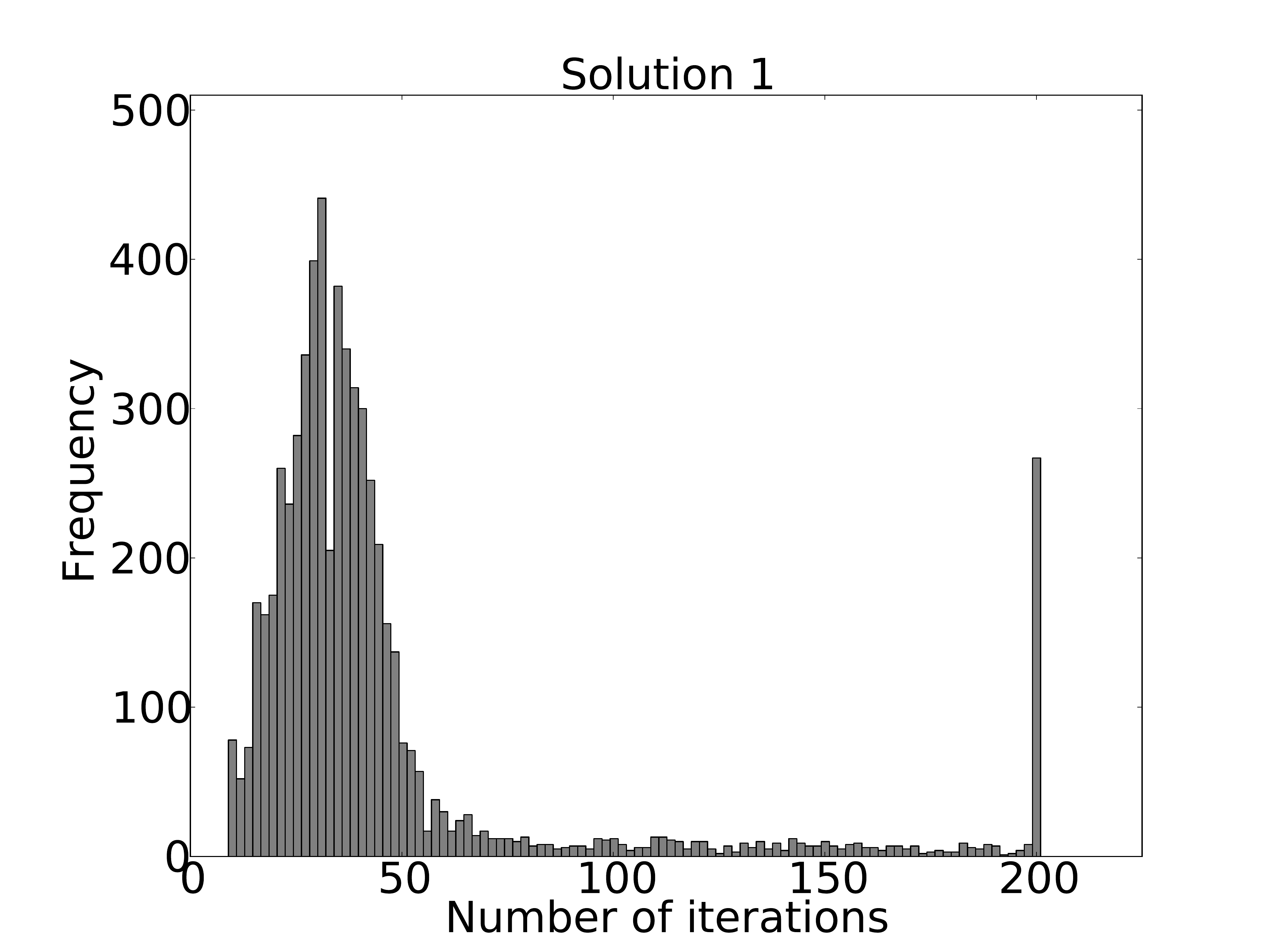}
	}
	\subfigure[][88\% solved in 100 iterations or less]
	{   
	    \includegraphics[trim = 0mm 0mm 35mm 0mm, clip, scale = 0.19] {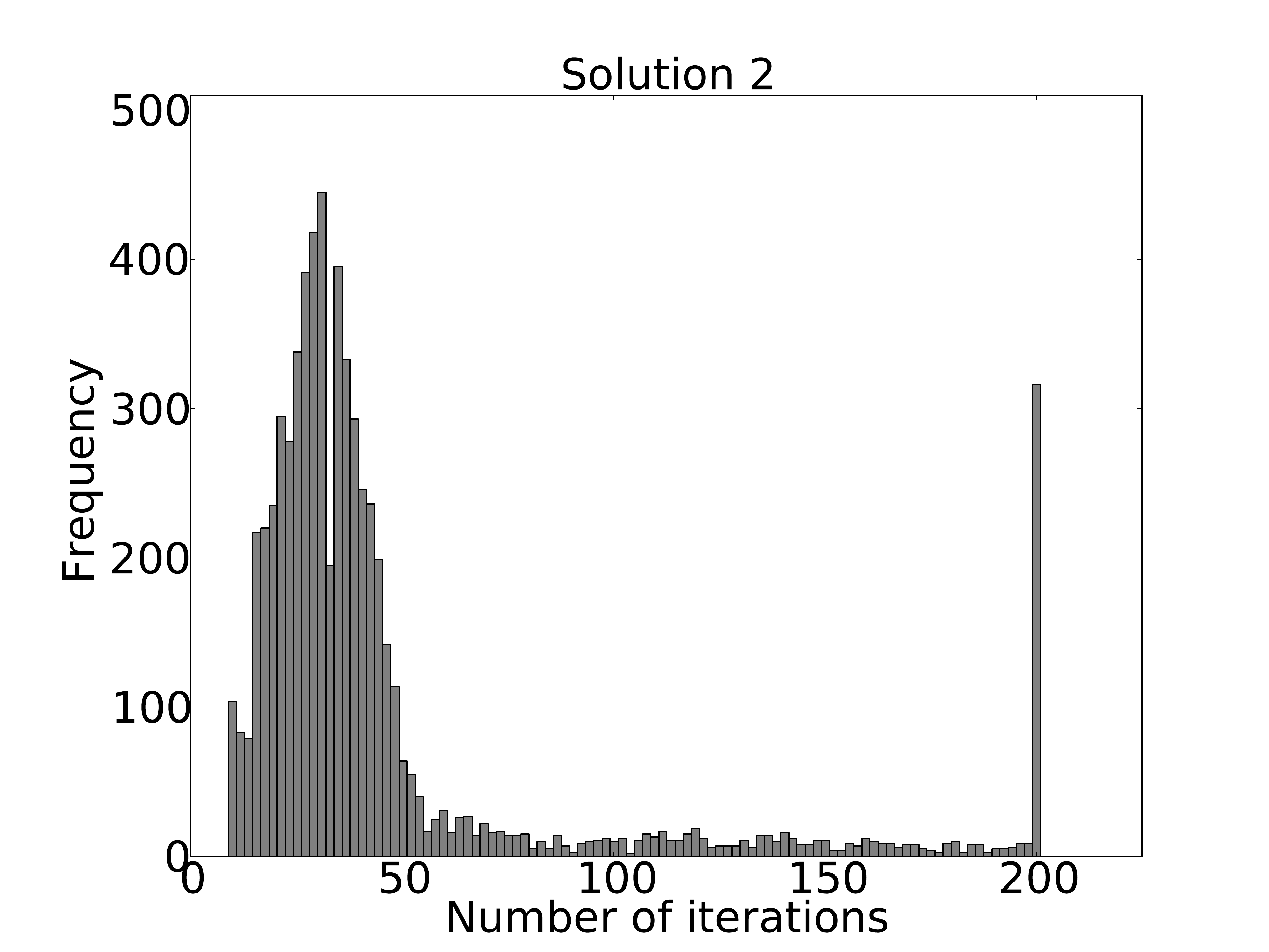}
	}\\
	\subfigure[][93\% solved in 100 iterations or less]
	{
	    \includegraphics[trim = 0mm 0mm 35mm 0mm, clip, scale = 0.19]
	    {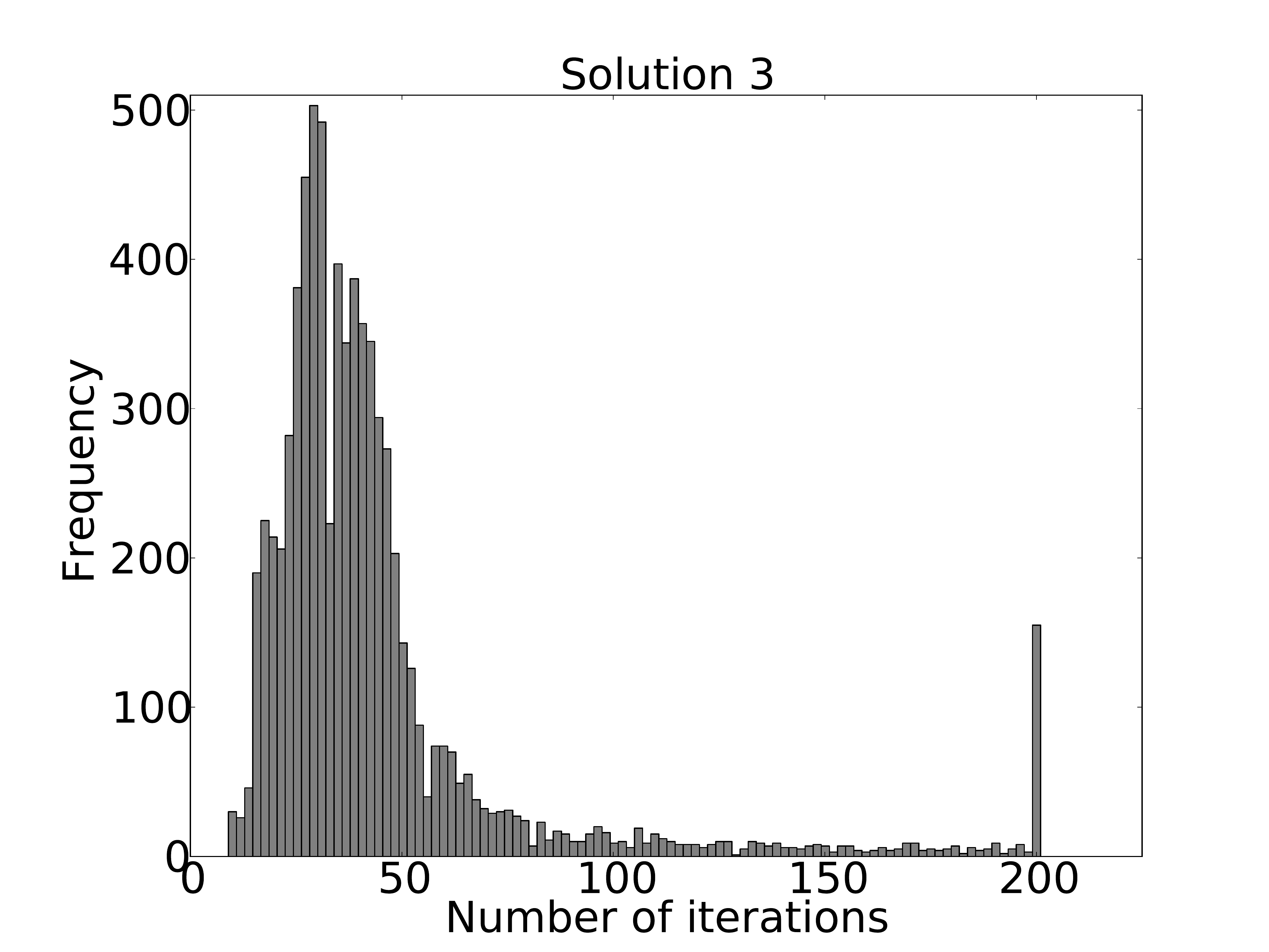}
	}
	\subfigure[][94\% solved in 100 iterations or less]
	{
	    \includegraphics[trim = 0mm 0mm 35mm 0mm, clip, scale = 0.19]
	    {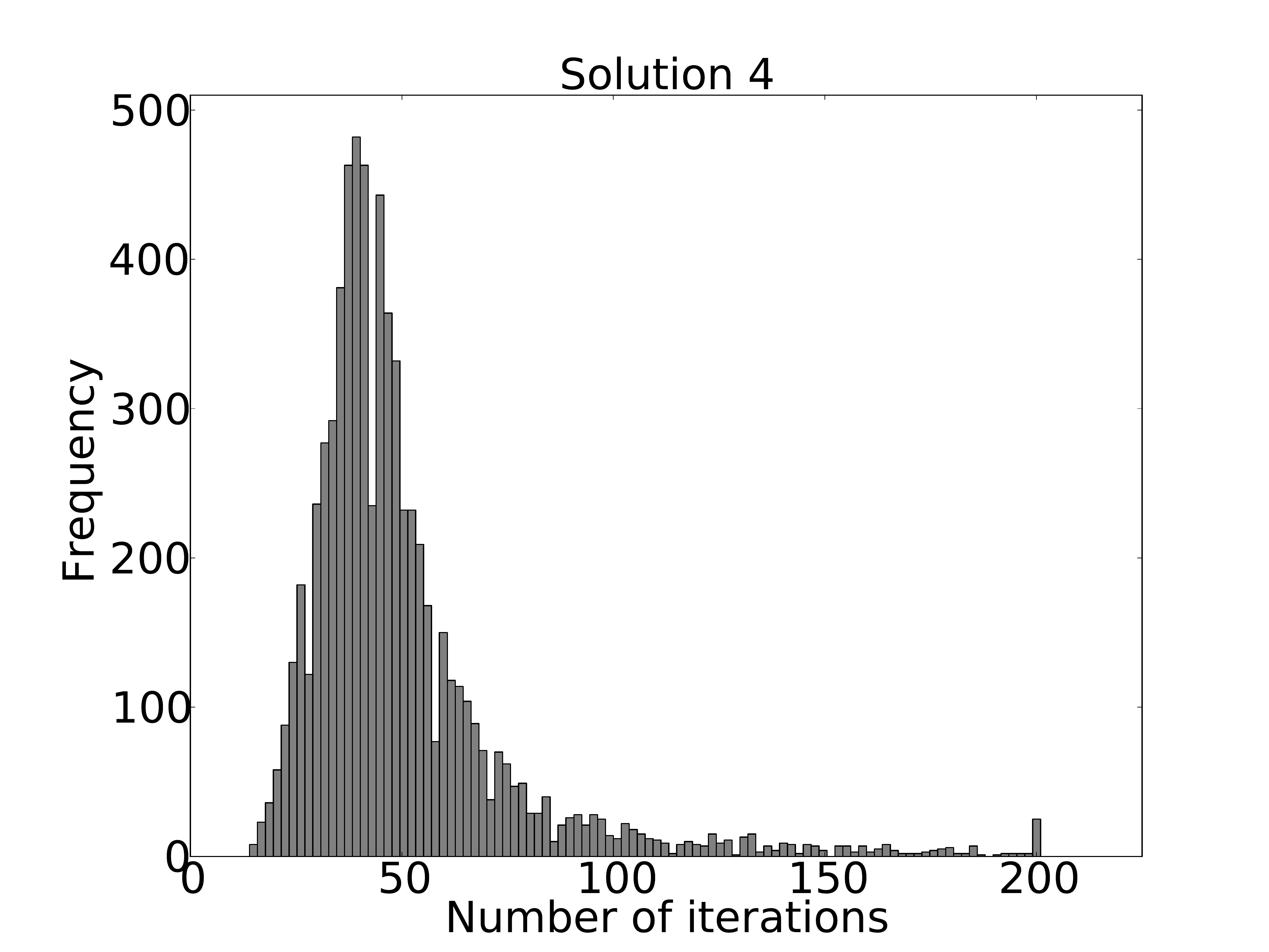}
	}
\end{center}
\caption{Histogram of number of iterations to compute solution of discomfort minimization problem.}
{\small{Total number of problems is 7500. The peak at 200 iterations is due to failed cases since maximum number of iterations was set to 200.}}
\label{fig:full_problem_iters}
\end{figure}

Histograms of number of iterations for computing 
final solution are shown in \Fig{fig:full_problem_iters}.
On average, 90\% all four solutions were computed in
100 iterations or less.

\begin{table}[p]
\begin{center}
  \begin{tabular}{  | c|c|c| }
    \hline
    {\bf Percentage of} & {\bf Number of }\\
    {\bf problems}			& {\bf successful solutions}\\
     \hline
    0.39 &    			 1\\  \hline
    6.20 &   			   2\\  \hline
    34.45 &   			3 \\ \hline
    58.96 &   			4\\  \hline
  \end{tabular}
\end{center}
\caption{Percentage of problems with one, two, three, or four successfully computed solutions.}{\small{ At least one solution
was found for all of the 7500 problems while all four solutions were found 
for almost 60\% of the problems.}}
\label{tab:successful_prob_stats}
\end{table}

\begin{table}[p]
\begin{center}
  \begin{tabular}{  | c|c|c| }
    \hline
    {\bf Solution}  & {\bf Gaussian} &{\bf Percentage}  \\
     & ($\mu$, $\sigma$) & {\bf of outliers}\\
     \hline
    1 &    			 (0.668, 0.751) & 3.56\\  \hline
    2 &   		 (0.686, 0.795) & 2.74\\  \hline
    3 &   			(0.612, 0.564) & 3.35 \\ \hline
    4 &   			(0.662, 0.344) & 2.64\\  \hline
  \end{tabular}
\end{center}
\caption{Estimating the percentage of outliers in computation time for
the discomfort minimization problem.}{\small{Mean and standard deviation of 
the Gaussian fitted to the data of
\Fig{fig:full_problem_time}. Points that lie outside $[\mu-  3\sigma, \mu
+ 3 \sigma]$ are outliers.}}
\label{tab:full_problem_gaussians}
\end{table}

\subsection{Discussion of results and limitations}
Results show that our framework is capable of reliably planning 
trajectories between a large variety of boundary conditions and for a 
range of weights. $97\%$ of $2916$ unconstrained problems
for a fixed boundary condition but varying weights were
solved successfully when weights were varied by 8 orders
of magnitude. Out of a set of $7500$ examples with
varying boundary conditions, and all dynamic constraints
imposed, 3.6 solution paths, on average, were found per example. 
At least one solution was found for all of the problems and
four solutions were found for roughly 60\% of the problems.
The time taken to compute the solution to the discomfort
minimization problem was less than 10 seconds for all
the cases, 99\% of all problems were solved in less
than 4 seconds, and roughly 90\% were solved in less
than 100 iterations.

We also saw that our framework can plan trajectories 
with a variety of boundary conditions that avoid obstacles.
We presented concrete examples for circular, elliptical,
and star-shaped obstacles. 
 
Thus, our framework, with some more speedups in run-time,
can be implemented for efficient and robust motion-planning
of nonholonomic mobile robots. We will discuss possible 
way of achieving speedups in computational time in
\Sec{sec:future_work}. 
One of the limitations of our current
implementation is that if the initial guess of path
passes through obstacles, it may take a large number of iterations
for the optimization algorithm to converge to a solution,
and sometimes a solution may not even be found. 
We have observed this on some example cases and
this will need a more careful analysis in the future.
One way to deal with this issue is to generate
initial guesses of path that are obstacle free and this is part
of future work.

There are many tasks in which an autonomous mobile robot must back
up and then move towards the goal. For example, if an assistive
robot is positioned at a user's desk, it cannot move forward. To
go anywhere it must back up first. Such tasks can be handled with
the help of a high-level planner that breaks this sequence into two
and provides a set of two boundary conditions in sequence 
to our framework -- one for backing up and one for the goal.
The intermediate waypoint can also be chosen by an
optimization process.

In our method, we impose obstacle avoidance constraints on
a discrete set of points on the path. Thus, we cannot
guarantee that segments of the path between these points
will not intersect obstacles. In practice, if the points 
are chosen to be close enough, so that distance between
these points is smaller than most obstacles, the path
would be collision-free. Even so, sharp pointed
corners of obstacles can intersect the path. This is a 
problem that will arise with any discretization. This can be
handled in two ways. First, when we incorporate robot's
body for obstacle-avoidance
an extra margin of safety can be added. Second,
obstacles can be represented with a piecewise smooth boundary
curve that encloses the obstacle shape such that sharp corners 
are smoothed out. 

\section{Concluding remarks and directions for future research}
\label{sec:concluding_remarks}
We make two main contributions in this work. First, we characterize
comfort for a user of an autonomous nonholonomic mobile robot.
Among the various contributing factors 
to comfort, we focus on dynamic
factors. For comfortable motion, a trajectory should have the
following properties -- it should satisfy boundary conditions on 
position, orientation, speed
and tangential acceleration at the start and end points, 
have continuous acceleration, the geometric path
should avoid obstacles, have curvature continuity, and should 
satisfy boundary conditions on curvature. In addition, the trajectory
should respect bounds on curvature, speed, angular speed, and 
tangential and normal accelerations.
While human user studies are required to validate this
characterization of comfort, we believe that we have taken an important first step
in formalizing motion comfort for autonomous mobile robots.

Second, we develop a nonlinear constrained optimization based motion planning framework to plan 
trajectories such that the trajectories
minimize discomfort and have all the properties
described above. To the best of our knowledge, this is
the first comprehensive formulation of kinodynamic motion
planning for a planar nonholonomic mobile robot that includes all of
the following -- a careful analysis of boundary conditions
and continuity requirements on trajectory, dynamic constraints,
obstacle avoidance constraints, and a robust numerical
method that computes solution trajectories in a few seconds.

One of the 
strengths of our framework is that it is easy to incorporate additional kinematic
and dynamic constraints, and additional terms can also be incorporated in
the discomfort functional. Of course, care has to be taken
to keep the problem mathematically meaningful. 

We believe that our work is an important step in developing
autonomous robots that are acceptable to human users.
For application to real-world robotic systems, some important 
extensions to our framework will be required. First, our current 
implementation achieves obstacle avoidance for a point robot. We have described
a method for incorporating robot shape, and this will have
to be implemented. Second, our results show that time taken
to find a solution is of the order of seconds. This will have
to be reduced for real-time planning. We discuss these, and several
other extensions, below.

\subsection{Directions for future research}
\label{sec:future_work}

{\bf Incorporating robot shape for obstacle avoidance}.
We described a general method to incorporate arbitrary shaped robot body 
in \Sec{sec:robot_shape}. This method consists of modeling the robot
as a closed curve that encloses the projection of its boundary
in the plane of motion, choosing a set of points on this curve, 
and imposing the constraints that all these points be outside all 
obstacles. If $m$ points are chosen on the boundary and
there are $n$ obstacles, this method will result in
$m \times n$ constraints. A more efficient approach may
be possible when the robot can be
modeled by a simple shape such as a circle or a convex polygon.
Since most mobile robots, in practice, have simple shapes,
it is worthwhile to explore these shapes as special cases
for obstacle avoidance. 
\bigskip

\noindent {\bf Incorporating moving obstacles}.
One way to incorporate moving obstacles is to frequently 
update a map of the world and use this updated map to re-plan
a new trajectory starting from the current state. 
For comfort of a human user, it may be useful to develop
models that estimate a moving obstacle's trajectory, and use
this trajectory during planning. This could result in paths
that have fewer changes in direction (compared to those
found by fast-re planning) and are perceived
to be more comfortable.
Such obstacle models have been previously employed for motion 
planning~\citep{Fiorini_1998}.
\bigskip

\noindent {\bf Culling obstacles intelligently}.
In our method,
we choose a set of points on the path, and impose the 
constraint that all obstacles be outside all
points on the path. In our earlier approaches, we have
experimented with culling these obstacles intelligently
so that the number of obstacle constraints is reduced.
If the trajectory is well-behaved, that is, if the
geometric path does not have too many self intersections,
and if
one iterate does not vary too wildly from the previous,
then
we may be able to achieve a reduction in the number of
constraints. 

First, we can remove, in advance, all 
obstacles that are too far from the initial guess of
path. Second, for every point, we impose the constraint
that it be outside obstacles within its ``neighborhood''
rather than being outside all obstacles.
Under the above described conditions, if a point 
is outside obstacles in its neighborhood, it can 
be expected to be outside all other obstacles that are far from it.
In our experiments with our current approach, we have 
observed that the above conditions hold if the initial 
guess of path is outside obstacles.
\bigskip

\noindent{\bf Computing initial guesses that avoid obstacles}
We have observed that the solution to the discomfort minimization problem
converges slowly if the initial guess of path passes through an obstacle.
We believe that we can achieve fast convergence if the initial guess of path lies
outside obstacles even if it does not respect continuity and kinodynamic
constraints.
Many of the existing path planning approaches can be used to compute
an initial guess of path that has the above properties.

\noindent {\bf Reducing computational time}.
For real-time implementation, it would be necessary to
achieve a reduction
in the computational time so that the problem is solved
in a few milliseconds. Many steps can be taken
to achieve this. 

First, we have observed that when
an initial guess of path is inside an obstacle, it takes
longer for the optimization algorithm to converge to a solution.
Therefore, it would be worthwhile to invest some effort in generating
an initial guess of path that is outside obstacles.
This would reduce the number of iterations required to find
a solution.

Second, intelligently culling obstacles
and efficiently implementing obstacle avoidance
constraints for special robot shapes, as discussed earlier,
could result in significant reduction in the number of constraints
and faster computations in every iteration. 

Third, a multi-step optimization procedure can be
tried. A coarser finite element mesh with fewer elements
can be used to find a solution which would serve as
an initial guess for a problem with a finer mesh.

Finally, 
parallelism inherent in the problem can be exploited
and parts of the program can be executed on a GPU.
For example, computation of constraint values, gradients
and Hessians can be parallelized. Other such parallelisms
should also be exploited. In addition, many other
code optimizations can also be implemented.
\bigskip

\noindent {\bf Evaluating the ``goodness'' of discomfort measure.}
We have formulated a measure of discomfort based on comfort studies
in ground vehicles such as automobiles and trains. To the best of our
knowledge, no such studies have been conducted for assistive robots.
Since discomfort is subjective, the best way to
assess comfort is to ask a user. Hence, to validate this discomfort
measure, human user studies should be conducted with enough users to
yield statistically significant data. 
We provide some guidelines on how such
a study may be conducted in \Sec{sec:guidelines_for_human_user_impl}
below.
\bigskip

\noindent {\bf Motion planning for non-planar surfaces.}
The motion planning framework presented in this work was developed for
planning trajectories for a nonholonomic mobile robot moving on a plane.
This assumption holds, for the most part, in indoor environments. 
For navigating in an urban outdoor environment, this framework can be extended
by parameterizing the path as a space curve rather than a 2D curve and formulating
the cost functional and constraints to take into account the 3D geometry of the
surface on which the robot moves.
\bigskip

\subsection{Implementation of the motion planning framework for human users}
\label{sec:guidelines_for_human_user_impl}
A human user study can be conducted to
either confirm
that the measure of discomfort is good by showing that multiple human
users can achieve comfort after choosing the weights, or failing that,
to provide additional insight into what might be missing. 
Below are some guidelines on implementing the framework
on an assistive robot and conducting such a study.

\begin{itemize}
\item Our motion planning framework requires a representation of
the local environment to plan trajectories. An occupancy-grid based
representation can be used. In such
a representation, obstacles are represented as occupied cells
in the grid. See~\citep{Thrun_2005} for a detailed discussion
of such a representation. For efficient motion planning, these cells should
be grouped together, where possible, into a single star-shaped polygon.
When such a grouping yields an obstacle that is not star-shaped,
it should be decomposed into a union of star-shaped polygons.
An efficient algorithm for doing so can be found in~\citep{Avis_1981}.

\item A goal state consisting of position, orientation, curvature, 
speed, and magnitude of tangential acceleration,  
is required as input to the
motion planning framework. Position and orientation may be
provided by a human user through some input device (e.g
by clicking on a map as in~\citep{Murarka_2009a}). Curvature
should be set to zero. Speed may be specified as zero if
it is desired to stop at the final position, otherwise is
should be a 
speed that is typically found comfortable by the user. 
Tangential acceleration should be set to zero. For navigating
in large-scale space, a high-level planner such as that
used in~\citep{Murarka_2009a} could be used for generating
intermediate way points. Such a planner usually
provides only position and orientation. The rest of the quantities
can be provided according to the guidelines above.

\item All necessary bounds should also be provided as input.
The bounds in Table~\ref{tab:bounds} may be used as a start
if the study is conducted for an intelligent wheelchair, while
the references cited in \Sec{sec:comfort} can be used for
the bounds if the study is conducted for an autonomous car.

\item A controller that can track the planned trajectory 
should be implemented. We have achieved good tracking accuracy, in our
previous work~\citep{Murarka_2009b}, with a feedback-linearization
based controller described in~\citep{DeLuca_1998}. 

\item Before performing human user experiments, the framework
should be comprehensively tested in the environment in which
the users will evaluate it. If the environment is likely to have moving
obstacles, fast re-planning should be implemented. This requires
trajectories to be computed in at most a tenth of second.
A relatively safe indoor environment with no drop-offs
and other hazards should be chosen and common failure cases should be 
identified via experimentation.

\item In the first step of the study, a user should be asked to
manually operate the assistive robot on a variety of tasks. A
speed that the user typically operates at should be determined
from these tasks.

\item Although a more detailed study than that described in \Sec{sec:effect_of_weights}
could yield an empirical relationship between weights and the
individual terms in our discomfort measure, such a study
is not an absolute prerequisite to performing human user 
studies. The two dimensionless factors corresponding
to the weights for integral of squared tangential jerk and
squared normal jerk are the parameters that should be varied
in the experiments.

\item First, the weight factor for tangential jerk should be determined.
To do this, the following experiment can be conducted.
Set start and end boundary conditions such that motion is
along a straight line. Set initial and final speed and
acceleration to zero. Use the motion planning framework to
plan trajectories for this task for a range of weight factors
for tangential jerk. Ask the user to compare discomfort
for every pair of weights. This comparison should include
subjective questions on overall comfort as well as questions
comparing the level of tangential jerk, and asking 
whether the time of travel was satisfactory. Vary the total 
length of the path and repeat the experiment for multiple lengths.
Based on these experiments, fix a value of this weight factor.

\item Next, the weight factor for normal jerk should be determined.
To do this, the following experiment can be conducted.
Set start at end boundary conditions such that most of the motion is
along a curved path. One way to achieve this is by choosing final position
very close to the start position such that the robot has to
travel along a curve to reach the goal. Follow a procedure
similar to the one described above (for tangential jerk) to determine
the weight factor for normal jerk.

\item Once the weight factors are determined, a set of 
motion tasks with a variety of boundary conditions should 
be performed and user should be asked to rate comfort.

\item If the motion for the above tasks is found to
be comfortable, then it can be concluded that the measure
of discomfort, in fact, captures user discomfort. If not,
a set of questions designed to learn what might be missing
should be asked.

\item In all cases, all quantitative information such
as speed, acceleration, jerk, travel time, length of path
etc., should be collected.

\end{itemize}

\section{Acknowledgements}
This work has taken place in the Intelligent Robotics Lab
at the Artificial Intelligence Laboratory, The University of
Texas at Austin. Research of the Intelligent Robotics lab
was supported in part by grants from the National Science
Foundation (IIS-0413257, IIS-0713150, and IIS-0750011), 
the National Institutes of Health (EY016089), and from the 
Texas Advanced Research Program (3658-0170-2007).

\bibliographystyle{apalike}
\bibliography{SG_CJ_BK_2012_part1}

\end{document}